\documentclass{article}

\usepackage{microtype}
\usepackage{graphicx}
\usepackage{subfigure}
\usepackage{booktabs} 
\usepackage{multirow}
\RequirePackage[table]{xcolor}
\usepackage{hyperref}

\usepackage[accepted]{icml2024}

\usepackage{amsmath}
\usepackage{amssymb}
\usepackage{mathtools}
\usepackage{amsthm}
\usepackage{subcaption}
\usepackage[capitalize,noabbrev]{cleveref}

\theoremstyle{plain}

\theoremstyle{definition}

\theoremstyle{remark}

\usepackage[textsize=tiny]{todonotes}
\usetikzlibrary{calc}

\definecolor{forestgreen4416044}{RGB}{44,160,44}
\definecolor{crimson}{RGB}{214,39,40}
\usepackage{tikz}
\newcommand{\tikzxmark}{%
\tikz[scale=0.23] {
    \draw[line width=0.7,line cap=round, draw=crimson] (0,0) to [bend left=6] (1,1);
    \draw[line width=0.7,line cap=round, draw=crimson] (0.2,0.95) to [bend right=3] (0.8,0.05);
}}
\newcommand{\tikzcmark}{%
\tikz[scale=0.23] {
    \draw[line width=0.7,line cap=round, draw=forestgreen4416044] (0.25,0) to [bend left=10] (1,1);
    \draw[line width=0.8,line cap=round, draw=forestgreen4416044] (0,0.35) to [bend right=1] (0.23,0);
}}

\usepackage{xcolor}
\usepackage{multicol}

\usepackage{amsmath}
\usepackage{amsfonts} 
\usepackage{bm}

\definecolor{mfvi}{RGB}{214,39,40} 
\definecolor{gmmvi}{RGB}{176,176,176} 
\definecolor{nfvi}{RGB}{255,127,14} 
\definecolor{smc}{RGB}{23,190,207} 
\definecolor{aft}{RGB}{44,160,44} 
\definecolor{craft}{RGB}{188,189,34} 
\definecolor{fab}{RGB}{127,127,127} 
\definecolor{dds}{RGB}{148,103,189} 
\definecolor{pis}{RGB}{227,119,194} 
\definecolor{mcd}{RGB}{140,86,75} 
\definecolor{ldvi}{RGB}{31,119,180} 
\definecolor{cmcd}{RGB}{227,119,194} 

\definecolor{darkgray176}{RGB}{176,176,176}

\newcommand{\olsi}[1]{\,\overline{\!{#1}}} 
\newcommand{\x}{\mathbf{x}}
\newcommand{\w}{\mathbf{w}}
\newcommand{\s}{\mathbf{s}}

\newcommand{\score}{\s^{\theta}}
\newcommand{\flow}{{T}^{\theta}}

\newcommand{\ex}{\mathbf{x}_{0:T}}
\newcommand{\uu}{\mathbf{z}}
\newcommand{\X}{\mathbf{X}}
\newcommand{\y}{\mathbf{y}}
\newcommand{\Y}{\mathbf{Y}}
\newcommand{\I}{\textbf{I}}
\newcommand{\dx}{\text{d}\x}

\newcommand{\extelbo}{\olsi{\text{ELBO}}}
\newcommand{\target}{\pi}
\newcommand{\KL}{D_{\text{KL}}}
\newcommand{\E}{\mathbb{E}}
\newcommand{\rd}{\mathbb{R}^d}
\newcommand{\model}{q^{\theta}}
\newcommand{\M}{\xi}
\def\gN{{\mathcal{N}}}

\newcommand\baseline[1]{\textcolor{nfvi}{#1}}

\icmltitlerunning{Beyond ELBOs: A Large-Scale Evaluation of Variational Methods for Sampling}

\begin{document}

\twocolumn[


\icmltitle{
Beyond ELBOs: A Large-Scale Evaluation of Variational Methods for Sampling
}




\begin{icmlauthorlist}
\icmlauthor{Denis Blessing}{yyy}
\icmlauthor{Xiaogang Jia}{yyy}
\icmlauthor{Johannes Esslinger}{yyy}
\icmlauthor{Francisco Vargas}{sch}
\icmlauthor{Gerhard Neumann}{yyy,comp}
\end{icmlauthorlist}

\icmlaffiliation{yyy}{Autonomous Learning Robots, Karlsruhe Institute of Technology, Karlsruhe, Germany}
\icmlaffiliation{comp}{FZI Research Center for Information Technology, Karlsruhe, Germany}
\icmlaffiliation{sch}{University of Cambridge, Cambridge, United Kingdom}

\icmlcorrespondingauthor{Denis Blessing}{denis.blessing@kit.edu}

\icmlkeywords{Machine Learning, ICML}

\vskip 0.3in
]



\printAffiliationsAndNotice{}  

\begin{abstract}
Monte Carlo methods, Variational Inference, and their combinations play a pivotal role in sampling from intractable probability distributions. However, current studies lack a unified evaluation framework, relying on disparate performance measures and limited method comparisons across diverse tasks, complicating the assessment of progress and hindering the decision-making of practitioners. In response to these challenges, our work introduces a benchmark that evaluates sampling methods using a standardized task suite and a broad range of performance criteria.
Moreover, we study existing metrics for quantifying mode collapse and introduce novel metrics for this purpose. Our findings provide insights into strengths and weaknesses of existing sampling methods, serving as a valuable reference for future developments. The code is publicly available \href{https://github.com/DenisBless/variational_sampling_methods}{here}.
\end{abstract}

\section{Introduction} \label{sec:intro}
Sampling methods are designed to address the challenge of generating approximate samples or estimating the intractable normalization constant $Z$ for a probability density $\pi$ on $\rd$ of the form
 \begin{equation}
    \target(\x) = \frac{\gamma(\x)}{Z}, \quad Z = \int_{\rd} \gamma(\x) \dx,
\end{equation}
where $\gamma: \rd \rightarrow \mathbb{R}^+$ can be pointwise evaluated. This formulation has broad applications in fields such as Bayesian statistics and the natural sciences \cite{liu2001monte, stoltz2010free, frenkel2023understanding, mittal2023exploring}.

Monte Carlo (MC) methods \cite{hammersley2013monte}, including Annealed Importance Sampling (AIS) \cite{neal2001annealed} and its Sequential Monte Carlo (SMC) extensions \cite{del2006sequential}, have traditionally been considered the gold standard for addressing the sampling problem. An alternative approach is Variational Inference (VI) \cite{blei2017variational}, where a tractable family of distributions is parameterized, and optimization tools are employed to maximize similarity to the intractable target distribution $\target$.

In recent years, there has been a surge of interest in the development of sampling methods that merge MC with VI techniques to approximate complex, potentially multimodal distributions \cite{wu2020stochastic, zhang2021path, arbel2021annealed, matthews2022continual, jankowiak2022surrogate, midgley2022flow, berner2022optimal, richter2023improved, vargas2023denoising,vargas2023bayesian, akhound2024iterated}.

However, the evaluation of these methods faces significant challenges, including the absence of a standardized set of tasks and diverse performance criteria. This diversity complicates meaningful comparisons between methods. Existing evaluation protocols, such as the evidence lower bound (ELBO), often rely on samples from the model, restricting their evaluation capabilities to the model's support. This limitation becomes especially problematic when assessing the ability to mitigate mode collapse on target densities with well-separated modes. To overcome this challenge, others propose the use of integral probability metrics (IPMs), like maximum mean discrepancy \cite{arenz2018efficient} or Wasserstein distance \cite{richter2023improved, vargas2023denoising}, leveraging samples from the target density to assess performance beyond the model's support. However, these metrics often involve subjective design choices such as kernel selection or cost function determination, potentially leading to biased evaluation protocols.

To address these challenges, our work introduces a comprehensive set of tasks for evaluating variational methods for sampling. We explore existing evaluation criteria and propose a novel metric specifically tailored to quantify mode collapse. Through this evaluation, we aim to provide valuable insights into the strengths and weaknesses of current sampling methods, contributing to the future design of more effective techniques and the establishment of standardized evaluation protocols.
\section{Preliminaries} \label{sec:relwork}
We provide an overview of Monte Carlo methods, Variational Inference, and combinations. The notation introduced in this section is used throughout the remainder of this work.

\textbf{Monte Carlo Methods.} A  variety of Monte Carlo (MC) techniques have been developed to tackle the sampling problem and estimation of $Z$. In particular sequential importance sampling methods such as Annealed Importance Sampling (AIS) \citep{neal2001annealed} and its Sequential Monte Carlo (SMC) extensions \citep{del2006sequential} are often regarded as a gold standard to compute $Z$. These approaches construct a sequence of distributions $(\target_t)_{t=1}^T$ that `anneal' smoothly from a tractable proposal distribution $\target_0$ to the target distribution $\target_T = \target$. One typically uses the geometric average, that is, $\gamma_t(\x) = \pi_0(\x)^{1- \beta_t}\gamma(\x)^{\beta_t}$, with $\target_t \propto \gamma_t$ for $0=\beta_0<\beta_1<...<\beta_T=1$. Approximate samples from $\target$ are then obtained by starting from $\x_0 \sim \target_0(\cdot)$ and running a sequence of Markov chain Monte Carlo (MCMC) transitions that target $(\target_t)_{t=1}^T$. 

\textbf{Variational Inference.} Variational inference (VI) \cite{blei2017variational} is a popular alternative to MCMC and SMC where one considers a flexible family of easy-to-sample distributions $q^{\theta}$ whose parameters $\theta$ are optimized by minimizing the reverse Kullback--Leibler (KL) divergence, i.e.,
\begin{equation}
    \KL(\model(\x)\|\target(\x)) = -\underbrace{\E_{\model(\x)} \Big[\log \frac{\gamma(\x)}{\model(\x)} \Big]}_{\text{ELBO}} + \log Z 
\end{equation}
It is well known that minimizing the reverse KL is equivalent to maximizing the evidence lower bound (ELBO) and that $\text{ELBO} \leq \log Z$ with equality if and only if $\model = \target$.
 Later, VI was extended to other variational objectives such as $\alpha$-divergences \cite{li2016renyi, midgley2022flow}, log-variance loss \cite{richter2020vargrad}, trajectory balance, 
\cite{malkin2022trajectory} or general $f$- divergences \cite{wan2020f}.
Typical choices for $q^{\theta}$ include mean-field approximations \citep{Wainwright:2008}, mixture models \cite{arenz2022unified} or normalizing flows \citep{papamakarios2019normalizing}. To construct more flexible variational distributions \citep{Agakov2004} modeled $q^{\theta}(\x)$ as the marginal of a latent variable model, i.e. $q^{\theta}(\x)=\int q^{\theta}(\x,\uu)\mathrm{d}\uu$ \footnote{\citet{Agakov2004} coined the term `augmentation' for $\uu$. We adopt the more established terminology and refer to $\uu$ as a latent variable.}. As this marginal is typically intractable, $\theta$ is then learned by minimizing a discrepancy measure between $q^{\theta}(\x,\uu)$ and an extended target $p^{\theta}(\x,\uu)=\target(\x)p^\theta(\uu|\x)$ where $p^\theta(\uu|\x)$ is an auxiliary conditional distribution. Using the chain rule for the KL-divergence \cite{cover1999elements} one obtains an extended version of the ELBO, that is,
\begin{equation}
    \KL(\model(\x)\|\target(\x)) \leq -\underbrace{\E_{\model(\x, \uu)} \Big[\log \frac{\gamma(\x)p^\theta(\uu|\x)}{\model(\x,\uu)} \Big]}_{\olsi{\text{ELBO}}} + \log Z.
\end{equation}
Although the extended ELBO is often referred to as ELBO, latent variables $\uu$ introduce additional looseness, i.e., $\extelbo \leq \text{ELBO}$ with equality if $p^\theta(\uu|\x) = \model(\x,\uu)/\model(\x)$. To compute expectations with respect to $\model(\x,\uu)$, one typically chooses tractable distributions $\model(\x|\uu)$ and $\model(\uu)$ and performs a Monte Carlo estimate using ancestral sampling.

\textbf{Variational Monte Carlo Methods.} Over recent years, the idea of using extended distributions has been further explored \cite{wunoe2020stochastic, Geffner:2021, Thin:2021, ZhangAIS2021,doucet2022annealed,geffner2022langevin}. In particular, these ideas marry Monte Carlo with variational techniques by constructing the variational distribution and extended target as Markov chains, i.e., 
$\model(\ex) =  \target_0(\x_{0}) \prod_{t=1}^{T} F^\theta_{t} (\x_{t}  | \x_{t-1})$ and $p^\theta(\ex) =  \target(\x_{T}) \prod_{t=0}^{T-1} B^\theta_{t} (\x_{t}  | \x_{t+1})$ with $\x = \x_T$, $\uu = (\x_0, \hdots, \x_{T-1})$ and tractable $\target_0$. Common choices of transition kernels $F^\theta_{t}, B^\theta_{t}$ include Gaussian distributions \cite{doucet2022annealed,geffner2022langevin} or normalizing flow maps \cite{wu2020stochastic, arbel2021annealed, matthews2022continual} and can be optimized by e.g. maximization of the extended ELBO via stochastic gradient ascent. Recently, \citet{vargas2023denoising, zhang2021path, vargas2023bayesian, vargas2023transport, richter2023improved, berner2022optimal} explored the limit of $T \rightarrow \infty$ in which case the Markov chains converge to forward and backward time stochastic differential equations (SDEs) \cite{anderson1982reverse, song2020score} inducing the path distributions $\mathbb{Q}^\theta$ and $ \mathbb{P}^\theta$ which can be thought of as continuous time analogous of $\model$ and $p^\theta$ respectively. \citet{zhang2021path, berner2022optimal, richter2023improved,vargas2023transport} leveraged the continuous-time perspective to establish connections with Schr\"odinger bridges \cite{leonard2013survey} and stochastic optimal control \cite{dai1991stochastic}, resulting in the development of novel sampling algorithms.



\begin{figure*}[t]
  \centering
  \includegraphics[width=0.7\textwidth]{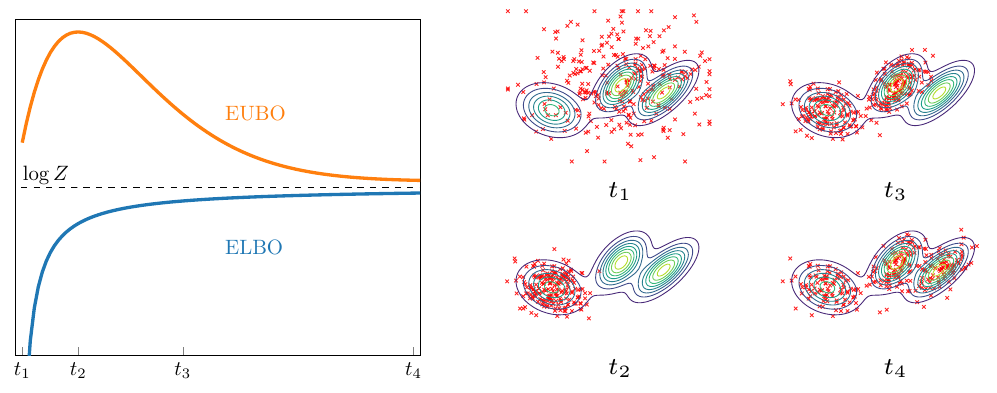}
\caption{Illustration of the evidence upper (EUBO) and lower bound (ELBO). The mode-seeking nature of reverse KL results in $\text{ELBO} \ll \log Z$ if the model density $\model$ (indicated by the samples $\color{red}{\times}$) averages over the target $\pi$ (indicated by the level plot) ($t_1$) and  $\text{ELBO} \approx \log Z$ if $\target \geq 0$ whenever $\model \geq 0$ ($t_2-t_4)$. As a result, the ELBO is not sensitive to mode collapse. In contrast, the mass-covering nature of the forward KL ensures that $\text{EUBO} \gg \log Z$ if $\model \approx 0$ whenever $\target > 0$ ($t_2)$ and $\text{EUBO} \approx \log Z$ if $\model \geq 0$ whenever $\target \geq 0$ ($t_1$). Consequently, the EUBO is well suited to quantify mode collapse.}
\label{fig:elbo_eubo}
\vspace{-0.2cm}
\end{figure*}

\textbf{Performance Criteria.}
Several performance criteria have been proposed for evaluating sampling methods, notably, those comparing the density ratio between the target and model density and integral probability metrics (IPMs).

Density ratio-based criteria make use of the ratio between the (unnormalized) target density $\gamma(\x)$ and the model $\model(\x)$. Due to the intractability of $\model(\x)$ for methods that work with latent variables, the density ratio between the joint distributions of $\x$ and $\uu$ is considered, i.e., 
\begin{equation}
    w = \frac{\gamma (\x)}{\model(\x)}, \ \text{ and } \ \olsi{w} = \frac{\gamma(\x)p^\theta(\uu|\x)}{\model(\x, \uu)},
\end{equation}
respectively. Note that $w$ and $\olsi{w}$ are also referred to as (unnormalized) importance weights. Using this notation, we can recover commonly used metrics such as the reverse effective sample size (ESS$_r$) or the ELBO, that is,
\begin{equation}
    \text{ESS}_{r} = \frac{(\E_{\model}[w])^2}{\E_{\model}[w^2]}
    \ \text{ and } \
    \text{ELBO} = \E_{\model}[\log w],
\end{equation}
respectively. Here, `reverse' is used to denote that expecations are computed with respect to $\model$. In addition, if the true normalization constant is known, an importance-weighted reverse estimate of $\log Z$ is often employed to report the esimation bias, i.e., $\Delta \log Z_{r} = |\log Z - \log \hat Z_{r}|$ with 
\begin{equation}
     \log \hat Z_{r} = \log \E_{\model}[w].
\end{equation}
Please note that extended versions of these criteria are obtainable by replacing $w$ with the extended version $\olsi{w}$ and taking expectations under the joint distribution $\model(\x, \uu)$.



\section{Quantifying Mode-Collapse}
Quantifying the ability to avoid mode collapse is difficult as identifying all modes of the target density $\target$ and determining whether a model captures them accurately is inherently challenging. In particular, methods that are optimized using the reverse KL divergence are forced to assign high probability to regions with non-negligible probability in the target distribution $\pi$. This is referred to as mode-seeking behavior and can result in an overemphasis on a limited set of modes, leading to mode collapse. 
Consequently, performance criteria that use expectations under the model $\model$, such as ELBO, (reverse) ESS, or $\Delta \log Z_r$, are influenced by the mode-seeking nature of the reverse KL divergence, making them less sensitive to mode collapse. 


Here, we aim to explore criteria that are sensitive to mode collapse such as density-ratio based `forward' criteria, that leverage expectations under $\pi$ and integral probability metrics (IPMs). Furthermore, we introduce \textit{entropic mode coverage}, a novel criterion that leverages prior knowledge about the target to heuristically quantify mode coverage.


\textbf{Forward Criteria.} We discuss the `forward' versions of the criteria discussed in Section \ref{sec:relwork}. First, evidence upper bounds (EUBOs) are based on the forward KL divergence and have already been leveraged as learning objectives in VI \cite{ji2019stochastic}. Here, we explore them as performance criteria that are sensitive to mode collapse. Formally, the EUBO is the sum of the forward KL and $\log Z$, that is, 
\begin{equation}
    \KL(\target(\x)\|\model(\x)) = \underbrace{\E_{\target(\x)} [\log w ]}_{\text{EUBO}} - \log Z ,
\end{equation}
with importance weights $w = {\gamma(\x)}/{\model(\x)}$.
Due to the non-negativeness of the KL divergence, it is easy to see that $\text{EUBO} \geq \log Z$ with equality if and only if $\model = \target$. Hence, a lower EUBO means that $\model$ is closer to $\target$ in a $\KL$ sense. The mass-covering nature of the forward KL leads to high EUBO values if the model fails to cover regions of non-negligible probability in the target distribution $\pi$ and is therefore well suited to quantify mode-collapse. This is further illustrated in Figure \ref{fig:elbo_eubo}. We can again leverage the chain rule for the KL-divergence \cite{cover1999elements} to obtain an extended version of the EUBO, i.e., $\E_{\target(\x, \uu)} [\log \olsi{w} ]$ that satisfies $\olsi{\text{EUBO}} \geq \text{EUBO}$, where the introduction of latent variables introduce additional looseness. The extended EUBO requires computing the importance weights $\olsi{w}$ and expectations under $\target(\x, \uu)$. The former depends on the specific choice of sampling algorithm and is further discussed in Section \ref{sec:methods} when introducing the methods considered for evaluation. The latter can be approximated by propagating target samples $\x$ back to $\uu$ using $\target(\uu|\x)$. Additionally, having access to samples from $\target$ allows for computing forward versions of $Z$ and ESS which have already been used to quantify mode collapse by e.g. \cite{midgley2023se}. Formally, they are defined as 
\begin{equation}
    Z_{f} =  1/\E_{\target}[ w^{-1}], \  \text{ and } \ \text{ESS}_{f} = {Z_{f}}/{\E_{\target}[w]},
\end{equation}
where expectations are taken with respect to the target $\target$. For a detailed discussion see Appendix \ref{appendix:densityratio}. 

\textbf{Integral Probability Metrics.}
 Alternatively, IPMs are often employed if samples from the target distribution $\target$ are available \cite{arenz2018efficient, richter2023improved, vargas2023denoising, vargas2023transport}. Common IPMs for assessing sample quality are 2-Wasserstein distance ($\mathcal{W}_{2}$) 
 \cite{peyre2019computational} or the maximum mean discrepancy (MMD) \cite{gretton2012kernel}. The former uses a cost function to calculate the minimum cost required to transport probability mass from one distribution to another while the latter assesses distribution dissimilarity by examining the differences in their mean embeddings within a reproducing kernel Hilbert space \cite{aronszajn1950theory}. For further details see Appendix \ref{appendix:ipms}.

\textbf{Entropic Mode Coverage.} Inspired by inception scores and distances from generative modelling \citep{salimans2016improved,heusel2017gans} we propose a heuristic approach for detecting mode collapse by introducing the entropic mode coverage (EMC). To compute EMC, we partition $\rd$ into disjoint subsets $\M_i, i\in \{1,\hdots,M\}$ that describe different modes of the target density $\target$. Moreover, we introduce an auxiliary distribution that measures the probability of a sample $\x$ being element of a mode descriptor, i.e., $p(\M_i|\x) = p(\x \in \M_i)$. We then compute the expected entropy of the auxiliary distribution, that is, 
\begin{align}
     \text{EMC} \coloneqq & \ \mathbb{E}_{\model(\x)}\mathcal{H}\left(p(\M|\x)\right) \nonumber \\  \approx &- \frac{1}{N}\sum_{\x \sim \model} \sum_{i=1}^M 
    p(\M_i|\x) \log_{M} {p(\M_i|\x)},
\end{align}
where the expectation is approximated using a Monte Carlo estimate. Here, $N$ denotes the number of samples drawn from $\model$. Please note that we employ the logarithm with a base of $M$ to ensure that $\text{EMC} \in [0,1]$. This choice of base facilitates a straightforward interpretation: A value of $0$ signifies a model that consistently produces samples that are elements of the same mode descriptor. In contrast, a value of $1$ represents a model that can produce samples from all mode descriptors with equal probability. 

Clearly, EMC is limited to targets where mode descriptors are available which is further discussed in Section \ref{sec:target}. Moreover, a suitable criterion for cases where mode descriptors are not equally probable is discussed in Appendix \ref{appendix:ejs}.

\section{Benchmarking Methods} \label{sec:methods}
\begin{table}[t!]
\centering
\resizebox{.49\textwidth}{!}{ 
\renewcommand{\arraystretch}{1.2}
\begin{tabular}{l|ll}
\toprule
\textbf{Acronym} &  \textbf{Method} & \textbf{Reference}  \\
\midrule
MFVI & Gaussian Mean-Field VI &\cite{bishop2006pattern} \\
GMMVI & Gaussian Mixture Model VI &\cite{arenz2022unified} \\
NFVI$^\dagger$ & Normalizing Flow VI &\cite{rezende2015variational} \\
SMC & Sequential Monte Carlo &\cite{del2006sequential} \\
AFT & Annealed Flow Transport &\cite{arbel2021annealed} \\
CRAFT & Continual Repeated AFT &\cite{matthews2022continual} \\
FAB & Flow Annealed IS Bootstrap &\cite{midgley2022flow} \\
ULA$^\dagger$ & Uncorrected Langevin Annealing &\cite{Thin:2021} \\
MCD & Monte Carlo Diffusion &\cite{doucet2022score} \\
UHA$^\dagger$ & Uncorrected Hamiltonian Annealing &\cite{Geffner:2021} \\
LDVI & Langevin Diffusion VI &\cite{geffner2022langevin} \\
CMCD$^\dagger$ & Controlled MCD &\cite{vargas2023transport} \\
PIS & Path Integral Sampler &\cite{zhang2021path} \\
DIS & Time-Reversed Diffusion Sampler & \cite{berner2022optimal} \\
DDS & Denoising Diffusion Sampler &\cite{vargas2023denoising}\\
GFN$^\dagger$ & Generative Flow Networks & \cite{lahlou2023theory} \\
GBS & General Bridge Sampler & \cite{richter2023improved} \\
\bottomrule
    \end{tabular}
    }
    \caption{Sampling Methods. For methods marked with `$\dagger$', implementation is available, but the results are either not included or only partially presented in this work.
    }
    \label{tab:methods}
\end{table}
\begin{table*}[t!]
    \centering
    \resizebox{.99\textwidth}{!}{ 
    \renewcommand{\arraystretch}{1.2}
    \begin{tabular}{l|cccccccccccc}
        \toprule
        & Funnel & Credit & Seeds & Cancer & Brownian  & Ionosphere & Sonar & Digits & Fashion & LGCP & MoG & MoS  \\
        \toprule
True $\log Z$ 
& $\tikzcmark$
& $\tikzxmark$
& $\tikzxmark$
& $\tikzxmark$
& $\tikzxmark$
& $\tikzxmark$
& $\tikzxmark$
& $\tikzcmark$
& $\tikzcmark$
& $\tikzxmark$
& $\tikzcmark$
& $\tikzcmark$
\\
Samples from $\target$ 
& $\tikzcmark$
& $\tikzxmark$
& $\tikzxmark$
& $\tikzxmark$
& $\tikzxmark$
& $\tikzxmark$
& $\tikzxmark$
& $\tikzcmark$
& $\tikzcmark$
& $\tikzxmark$
& $\tikzcmark$
& $\tikzcmark$
\\
Mode descriptors $\M$
& $\tikzxmark$
& $\tikzxmark$
& $\tikzxmark$
& $\tikzxmark$
& $\tikzxmark$
& $\tikzxmark$
& $\tikzxmark$
& $\tikzcmark$
& $\tikzcmark$
& $\tikzxmark$
& $\tikzcmark$
& $\tikzcmark$
\\
Dimensionality $D$
& 10
& 25
& 26
& 31
& 32
& 35
& 61
& 196
& 784
& 1600
& $\mathbb{N}_+$
& $\mathbb{N}_+$
\\
\bottomrule
    \end{tabular}
    }
    \caption{Target densities $\pi(\x)=\gamma(\x)/Z$ considered in this work.
    }
    \label{tab:targets}
\end{table*}

In this section, we elaborate on the methods included in this benchmark, categorizing them into three distinct groups based on the computation of importance weights. Please refer to Table \ref{tab:methods} for an overview of these methods and to Appendix \ref{appendix:is} for further details.

\textbf{Tractable Density Models.} Tractable density models allow for computing the model likelihood $\model(\x)$. It is therefore straightforward to compute performance criteria associated with importance weights $w = {\gamma(\x)}/{\model(\x)}$. Notable works include factorized (`mean-field') Gaussian distributions (\baseline{MFVI}), Normalizing Flows (\baseline{NFVI}) \cite{rezende2015variational} and full rank Gaussian mixture models (\baseline{GMMVI}) \cite{arenz2022unified}. 

\textbf{Sequential Importance Sampling Methods.} Sequential importance sampling (SIS) methods define $\olsi{w}$ in terms of incremental importance sampling (IS) weights, that is, $\olsi{w} =\prod_{t=1}^T G_{t}(\x_{t-1}, \x_t)$ with
\begin{equation}
    G_{t}(\x_{t-1}, \x_t) = \frac{\gamma_t(\x_t)B^{\theta}_{t-1} (\x_{t-1}  | \x_{{t}} )}{\gamma_{t-1}(\x_{t-1})F^{\theta}_{t} (\x_{{t}}  | \x_{{t-1}} )},
\end{equation}
with annealed versions $\gamma_t$ of $\gamma$. For example, choosing  $B^{\theta}_{t-1} (\x_{t-1}  | \x_{{t}} ) = \target_{t}(\x_{t-1})F^{\theta}_{t}(\x_t|\x_{t-1})/\target_{t}(\x_{t})$ recovers AIS \cite{neal2001annealed}. 
\citet{midgley2022flow} proposed to parameterize the proposal distribution $\pi_0$ with normalizing flows and, in combination with AIS, to minimize the $\alpha$-divergence, resulting in the Flow Annealed Importance Sampling Bootstrap (\baseline{FAB}) algorithm. Additionally, when AIS is coupled with resampling, it gives rise to Sequential Monte Carlo (\baseline{SMC}) as originally proposed by \citet{del2006sequential}.

Recent advancements include the development of Stochastic Normalizing Flows \cite{wu2020stochastic}, Annealed Flow Transport (\baseline{AFT}) \cite{arbel2021annealed}, and Continual Repeated AFT (\baseline{CRAFT}) \cite{matthews2022continual}. These methods extend Sequential Monte Carlo by employing sets of normalizing flows that define deterministic transport maps between neighboring distributions $\gamma_t$. For further details on $F^{\theta}_{t}, B^{\theta}_{t-1}$ and the corresponding $G_{t}$ see Table \ref{tab:sis}. For an in-depth exploration of the commonalities and distinctions among these methods, please refer to \citep{matthews2022continual}.

\textbf{Diffusion-Based Methods.} Diffusion-based methods typically build on stochastic differential equations (SDEs) with parameterized drift terms \cite{tzen2019neural}, i.e., 
\begin{align}
        \label{eq:SDE}
        \mathrm d \x_t &= f_t^{\theta}(\x_t) \mathrm d t + \sigma_t {\mathrm d} \w_t, \qquad \x_0 \sim \pi_0, \nonumber \\
        \mathrm d \x_t &= b_t^{\theta}(\x_t) \mathrm d t + \sigma_t {\mathrm d} \bar \w_t, \qquad \x_T \sim \pi_T, 
\end{align}
with diffusion coefficient $\sigma_t$ and standard Brownian motion $\w_t, \bar \w_t$. Using the Euler-Maruyama method \cite{sarkka2019applied}, their discretized counterparts can be characterized by Gaussian forward-backward transition kernels
\begin{align}
    \label{eq:fwd_bk}
    F^{\theta}_{t+1} (\x_{t+1}  | \x_{t} ) & = \mathcal{N}(\x_{t+1}| \x_{t}+ f_t^{\theta}(\x_t)\Delta_t, \sigma_t^2 \Delta_t) \text{ and} \nonumber \\
    B^{\theta}_{t-1} (\x_{t-1}  | \x_{t} ) & = \mathcal{N}(\x_{t-1}| \x_{t}+ b_t^{\theta}(\x_t)\Delta_t, \sigma_t^2 \Delta_t),
\end{align}
with discretization step size $\Delta_t$.
 The extended (unnormalized) importance weights $\olsi{w}$ can then be constructed as 
\begin{equation}
    \frac{p^\theta(\ex)}{\model(\ex)} \propto \olsi w = 
    \frac{ \gamma(\x_{T}) \prod_{t=1}^{T} \ B^\theta_{t-1} (\x_{t-1}  | \x_{t})}{\target_0(\x_{0}) \prod_{t=0}^{T-1} F^\theta_{t+1} (\x_{t+1}  | \x_{t})}.
\end{equation}
One line of work considers annealed Langevin dynamics to model Eq. \eqref{eq:SDE}. Works include Unadjusted Langevin Annealing (\baseline{ULA}) \cite{Thin:2021}, Monte Carlo Diffusions (\baseline{MCD}) \cite{doucet2022score}, Controlled Monte Carlo Diffusion (\baseline{CMCD}) \cite{vargas2023transport}, Uncorrected Hamiltonian Annealing (\baseline{UHA}) \cite{Geffner:2021} and Langevin Diffusion Variational Inference (\textcolor{black}{\baseline{LDVI}}) \cite{geffner2022langevin}. A second line of work describes diffusion-based sampling from a stochastic optimal control perspective \cite{dai1991stochastic}. Works include methods such as Path Integral Sampler (\baseline{PIS}) \cite{zhang2021path,vargas2023bayesian}, Denoising Diffusion Sampler (\baseline{DDS}) \cite{vargas2023denoising}, Time-Reversed Diffusion Sampler (\baseline{DIS}) \cite{berner2022optimal} and Generative Flow Networks (\baseline{GFN}) \cite{lahlou2023theory, malkin2022gflownets, zhang2023diffusion}. Furthermore, \citet{richter2023improved} identify several of these methods as special cases of a General Bridge Sampler (\baseline{GBS}) where both processes in Eq. \ref{eq:SDE} are freely parameterized. Specific choices of $\pi_0, F^{\theta}_{t+1}, B^{\theta}_{t-1}$ are detailed in Table \ref{tab:diffusion_methods}. Lastly, we refer the interested reader to \cite{sendera2024diffusion} which concurrently benchmarked diffusion-based sampling methods.

\begin{minipage}[h]{0.48\textwidth}
\centering
 \resizebox{\textwidth}{!}{%
\includegraphics[]{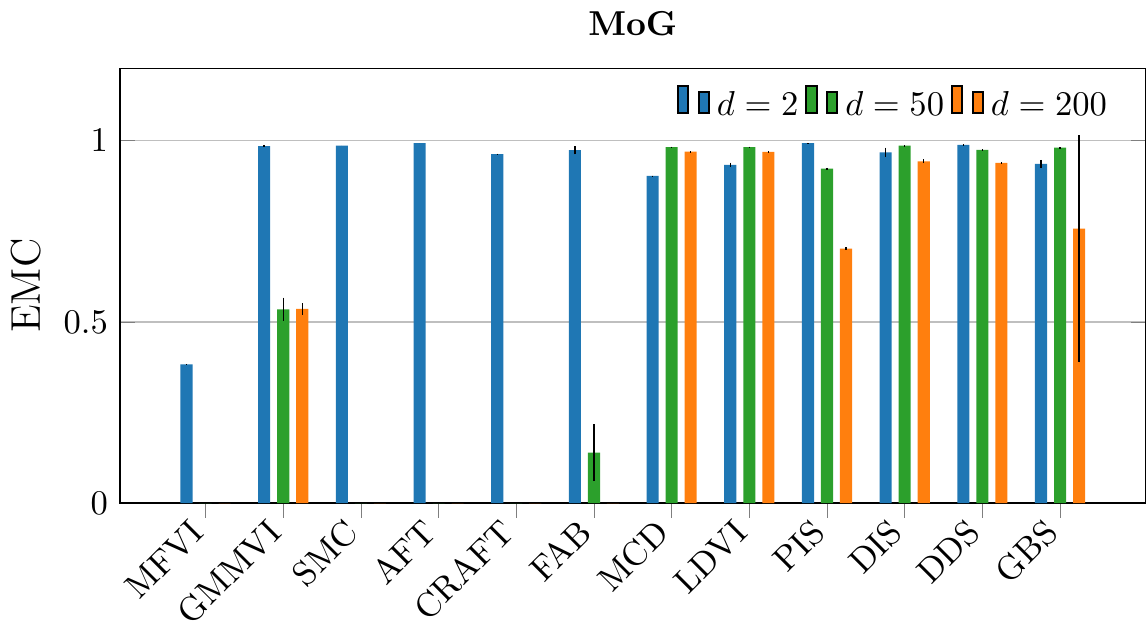}
}

\centering
 \resizebox{\textwidth}{!}{%
\includegraphics[]{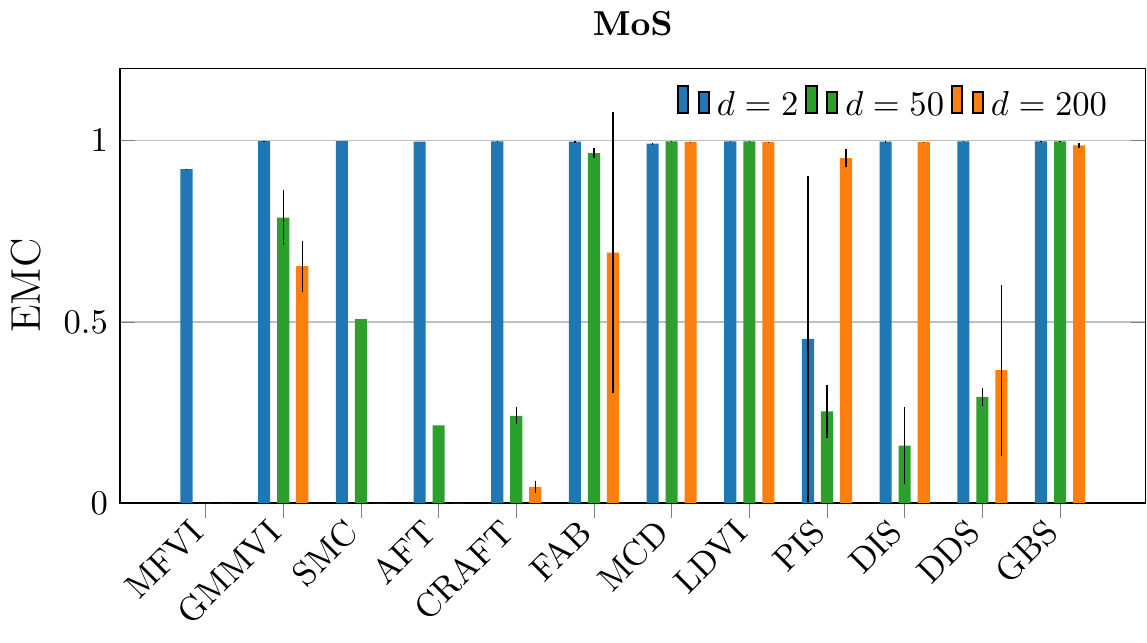}
}
\captionof{figure}{Mean and standard deviation of EMC values for MoG and MoS across varying dimensions $d$.}
\label{fig:dim}
\end{minipage}

\section{Benchmarking Target Densities} \label{sec:target}
\begin{figure}[t!]
\setlength{\tabcolsep}{1pt}
\centering
\begin{tabular}{cccc}
  Funnel &   
  MoS &   
  MoG &   
  \\
  \includegraphics[width=0.141\textwidth]{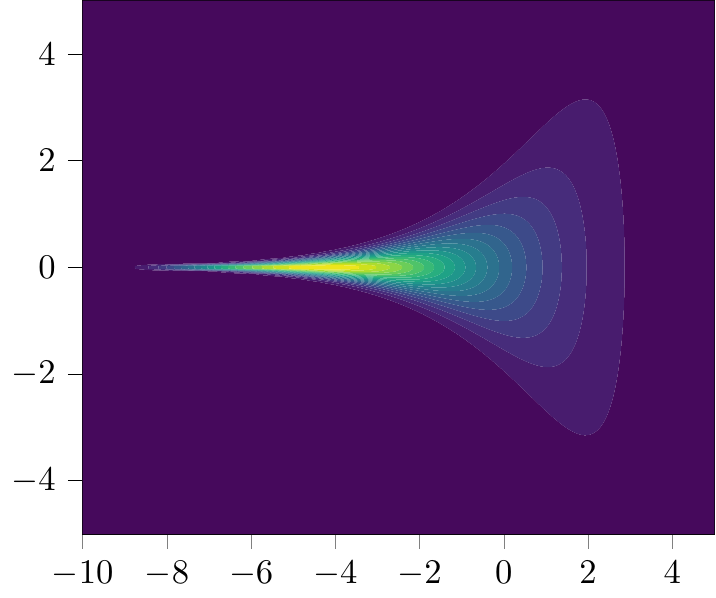} &   
  \includegraphics[width=0.15\textwidth]{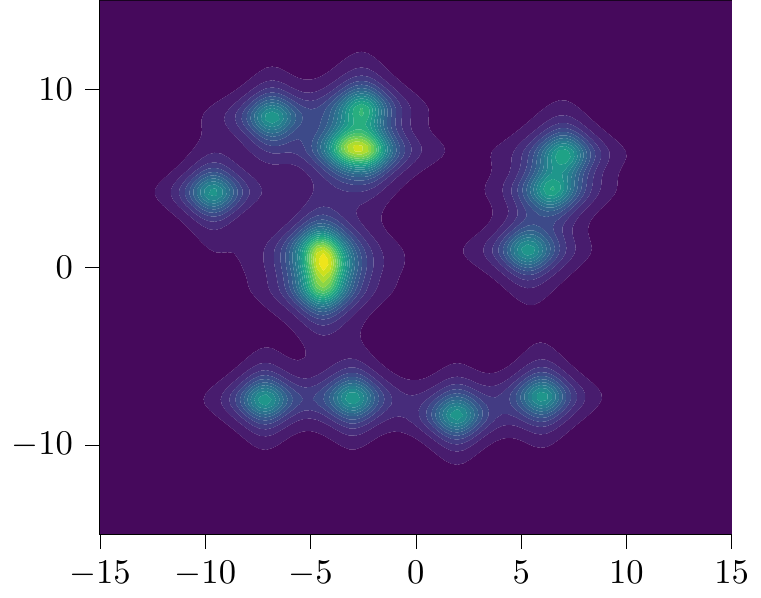} &   
  \includegraphics[width=0.15\textwidth]{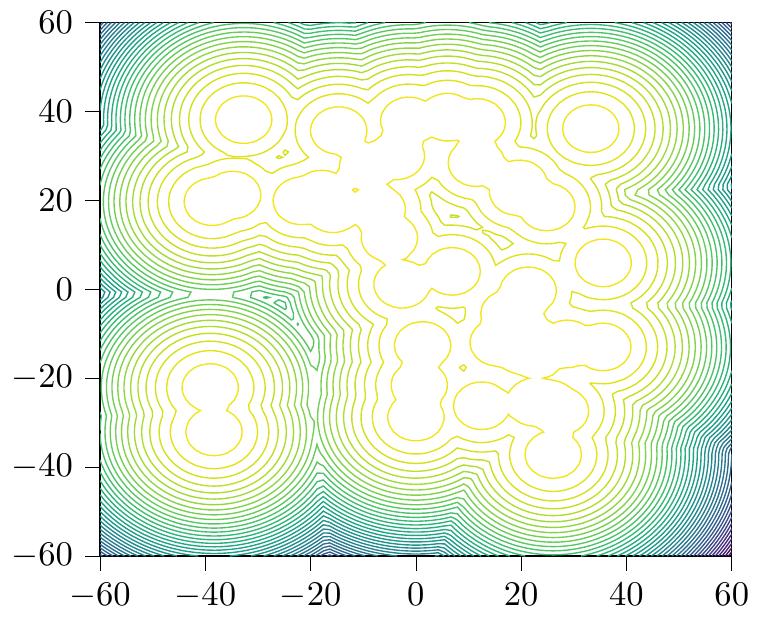} &   
  \\
\end{tabular}
\caption{Synthetic target densities. Left: First two dimensions of the funnel density. Middle: Mixture of Student-t distribution with $15$ components (MoS). Right: Mixture of $40$ isotropic Gaussian distributions (MoG).}
\label{fig:targets}
\end{figure}

Here, we briefly summarize the target densities $\pi$ considered in this work. The dimensionality of the problem, if we have access to the log normalizer $\log Z$, target samples, or mode descriptors for computing the entropic mode coverage is presented in Table \ref{tab:targets}. Further details and formal definitions of the target densities can be found in Appendix \ref{appendix:target_details}.

\textbf{Bayesian Logistic Regression.} We consider four experiments where we perform inference over the parameters of a Bayesian logistic regression model for binary classification. The datasets \textit{Credit} and \textit{Cancer} were taken from \citet{nishihara2014parallel}. The former distinguishes individuals as either good or bad credit risks, while the latter deals with the classification of recurrence events in breast cancer. The \textit{Ionosphere} dataset \cite{sigillito1989classification} involves classifying radar signals passing through the ionosphere as either good or bad. Similarly, the \textit{Sonar} dataset \cite{gorman1988analysis} tackles the classification of sonar signals bounced off a metal cylinder versus those bounced off a roughly cylindrical rock.

\textbf{Random Effect Regression.} The \textit{Seeds} data was collected by \cite{crowder1978beta}. The goal is to perform inference over the variables of a random effect regression model that models the germination proportion of seeds arranged in a factorial layout by seed and type of root. 

\textbf{Time Series Models.} We consider the \textit{Brownian} time series model obtained by discretizing a stochastic differential equation, modeling
a Brownian motion with a Gaussian observation model, developed by \cite{inferencegym2020}.

\textbf{Spatial Statistics.} The log Gaussian Cox process (\textit{LGCP}) \cite{moller1998log} is a probabilistic model commonly used in statistics to model spatial point patterns. In this work, the log Gaussian Cox process is applied to modeling the positions of pine saplings in Finland.


\textbf{Synthetic Targets.} We additionally consider synthetic target densities as they commonly give access to the true normalization constant $Z$, target samples, and mode descriptors. The \textit{Funnel} target was introduced by \cite{neal2003slice} and provides a complex `funnel'-shaped distribution. Moreover, we consider two different types of mixture models: a mixture of isotropic Gaussians (\textit{MoG}) as proposed by \citet{midgley2022flow}, and Student-t distributions (\textit{MoS}). To obtain mode descriptors for a mixture model with $K$ components, i.e., $\target(\x) = \sum_k\target_k(\x)/K$ we compute the density per component $\target_k(\x)$ and say that $\x \in \M_i$ if  $i = \text{argmax}_k \{\target_k(\x)\}_{k=1}^K$.
Lastly,  we follow \citet{doucet2022score} and train NICE \cite{dinh2014nice} on a down-sampled $14 \times 14$ variant of MNIST (\textit{Digits}) \cite{lecun1998gradient} and a $28 \times 28$ variant of Fashion MNIST (\textit{Fashion}) and use the trained model as target density. Here, we obtain the mode descriptors by training a classifier $p(\mathbf{c}|\x)$ on samples from $\pi$ where the classes $\mathbf{c}$ are represented by ten different digits. If $i = \text{argmax}_{\mathbf{c}} \ p(\mathbf{c}|\x)$ we conclude $\x \in \M_i$.


\section{Hyperparameters and Tuning} \label{section:tuning}
In this section, we provide details on hyperparameter tuning. For further information, please refer to Appendix \ref{appendix:algos_params}.

\textbf{Tractable Density Methods.} For MFVI, we used a batch size of 2000 and performed 100k gradient steps, tuning the learning rate via grid search. For targets with known support, we adjusted the initial model variance accordingly. For GMMVI, we adhered to the default settings from \cite{arenz2022unified}, utilizing 100 samples per mixture component. We initialized with 10 components and employed an adaptive scheme to add and remove components heuristically. The initial variance of the components was set based on the target support, and we conducted 3000 training iterations.

\textbf{Sequential Importance Sampling Methods.} In SIS methods, we employed 2000 particles for training. All methods except FAB used 128 annealing steps; FAB followed the original 12 steps as proposed by its authors. The choice and parameters of the MCMC transition kernel significantly impacted performance. Hamilton-Monte Carlo \cite{duane1987hybrid} generally outperformed Metropolis-Hastings \cite{chib1995understanding} (see Appendix \ref{abl:smc_choices}). Step sizes for $\beta_t \geq 0.5$ and $\beta_t < 0.5$ were tuned using grid search. For AFT and CRAFT, we used diagonal affine flows \cite{papamakarios2019normalizing}, which yielded more robust results than complex flows like inverse autoregressive flows \cite{kingma2016improved} or neural spline flows \cite{durkan2019neural} (see Appendix \ref{abl:flows}). FAB employed RealNVP \cite{dinh2016density} for the proposal distribution $\pi_0$. Learning rates for these flows were also tuned via grid search. For targets with known support, the variance of $\pi_0(\x) = \mathcal{N}({0}, \sigma^2_0{I})$ was set accordingly, otherwise, a grid search was performed. We used multinomial resampling with a threshold of 0.3 \cite{douc2005comparison}.

\textbf{Diffusion-based Methods.} Training involved a batch size of 2000 and 40k gradient steps. SDEs were discretized with 128 steps, $T=1$, and a fixed $\Delta t$. The diffusion coefficient was chosen as $\sigma_t = \sigma_{\text{max}} \cos^2({\pi (T-t)}/{2T})$, following \cite{vargas2023denoising} for better performance compared to linear or constant schedules. We used the architecture from \cite{zhang2021path} with 2 layers of 64 hidden units each. For targets with known prior support, the initial model support was set accordingly. For all methods except PIS, this involved setting the variance of the prior distribution $\pi_0(\x) = \mathcal{N}({0}, \sigma^2_0{I})$. For PIS, $\sigma_{\text{max}}$ was carefully chosen. In MCD and LDVI, we learned the annealing schedule $\beta_t$ and $\sigma_{\text{max}}$ end-to-end by maximizing the ELBO.

\begin{table*}[t!]
\begin{center}
\begin{small}
\begin{sc}
\resizebox{\textwidth}{!}{%
\renewcommand{\arraystretch}{1.2}
\begin{tabular}{l|rr|rr|rr|rr|rr}
\toprule
& \multicolumn{2}{c}{\textbf{Funnel}} 
& \multicolumn{2}{c}{\textbf{MoG} $(d=50)$}  
& \multicolumn{2}{c}{\textbf{MoS} $(d=50)$} 
& \multicolumn{2}{c}{$14 \times 14$ \textbf{Digits}}  
& \multicolumn{2}{c}{$28 \times 28$ \textbf{Fashion}} 
\\ 
\midrule
& {$\mathcal{W}_{2}$ $\downarrow$}
& {$\text{MMD}$ $\downarrow$}
& {$\mathcal{W}_{2}$ $\downarrow$}
& {$\text{MMD}$ $\downarrow$}
& {$\mathcal{W}_{2}$ $\downarrow$}
& {$\text{MMD}$ $\downarrow$}
& {$\mathcal{W}_{2}$ $\downarrow$}
& {$\text{MMD}$ $\downarrow$}
& {$\mathcal{W}_{2}$ $\downarrow$}
& {$\text{MMD}$ $\downarrow$}
\\
\midrule
MFVI
& $178.264 \scriptstyle \pm 0.271$ 
& $0.303 \scriptstyle \pm 0.002$ 
& $39360.196 \scriptstyle \pm 12.49$ 
& $0.209 \scriptstyle \pm 0.000$ 
& $2462.260 \scriptstyle \pm 1.009$ 
& $0.215 \scriptstyle \pm 0.000$ 
& $254.179 \scriptstyle \pm 0.025$ 
& $0.351 \scriptstyle \pm 0.000$ 
& $1327.517 \scriptstyle \pm 0.845$ 
& $0.285 \scriptstyle \pm 0.000$ 
\\
GMMVI
& $\mathbf{105.620 \scriptstyle \pm 3.472}$ 
& $\mathbf{0.031 \scriptstyle \pm 0.000}$ 
& $32004.968 \scriptstyle \pm 1069.$ 
& $0.203 \scriptstyle \pm 0.013$ 
& $1255.216 \scriptstyle \pm 296.9$ 
& $0.135 \scriptstyle \pm 0.017$ 
& $207.163 \scriptstyle \pm 14.60$ 
& $0.373 \scriptstyle \pm 0.042$ 
& $1343.495 \scriptstyle \pm 136.9$ 
& $0.462 \scriptstyle \pm 0.033$ 
\\
SMC
& $149.353 \scriptstyle \pm 2.973$ 
& $0.162 \scriptstyle \pm 0.015$ 
& $46351.236 \scriptstyle \pm 4.795$ 
& $0.631 \scriptstyle \pm 0.000$ 
& $3297.640 \scriptstyle \pm 1372.$ 
& $0.431 \scriptstyle \pm 0.161$ 
& $159.255 \scriptstyle \pm 1.877$ 
& $1.168 \scriptstyle \pm 0.008$ 
& $6696.287 \scriptstyle \pm 250.4$ 
& $1.556 \scriptstyle \pm 0.008$ 
\\
AFT
& $145.138 \scriptstyle \pm 6.061$ 
& $0.159 \scriptstyle \pm 0.010$ 
& $44914.194 \scriptstyle \pm 1154.$ 
& $0.622 \scriptstyle \pm 0.009$ 
& $2648.410 \scriptstyle \pm 301.3$
& $0.395 \scriptstyle \pm 0.082$ 
& $172.685 \scriptstyle \pm 3.661$ 
& $1.180 \scriptstyle \pm 0.004$ 
& $6413.147 \scriptstyle \pm 548.6$ 
& $1.538 \scriptstyle \pm 0.010$ 
\\
CRAFT
& $134.335 \scriptstyle \pm 0.663$ 
& $0.115 \scriptstyle \pm 0.003$ 
& $43412.038 \scriptstyle \pm 420.9$ 
& $0.604 \scriptstyle \pm 0.002$ 
& $1893.926 \scriptstyle \pm 117.3$ 
& $0.257 \scriptstyle \pm 0.024$ 
& $151.791 \scriptstyle \pm 11.02$ 
& $0.139 \scriptstyle \pm 0.032$ 
& $1413.303 \scriptstyle \pm 11.20$ 
& $0.562 \scriptstyle \pm 0.002$ 
\\
FAB
& $153.894 \scriptstyle \pm 3.916$ 
& $\mathbf{0.032 \scriptstyle \pm 0.000}$
& $9567.319 \scriptstyle \pm 626.1$ 
& $0.073 \scriptstyle \pm 0.005$ 
& $\mathbf{1204.160 \scriptstyle \pm 147.7}$ 
& $\mathbf{0.093 \scriptstyle \pm 0.014}$ 
& $\mathbf{126.863 \scriptstyle \pm 0.581}$ 
& $\mathbf{0.129 \scriptstyle \pm 0.003}$ 
& $1186.967 \scriptstyle \pm 263.4$ 
& $0.347 \scriptstyle \pm 0.007$ 
\\
MCD
& $163.317 \scriptstyle \pm 0.101$ 
& $0.228 \scriptstyle \pm 0.001$ 
& $5026.147 \scriptstyle \pm 40.03$
& $0.043 \scriptstyle \pm 0.000$ 
& $6418.981 \scriptstyle \pm 22.15$ 
& $0.256 \scriptstyle \pm 0.000$ 
& $220.710 \scriptstyle \pm 5.547$ 
& $0.252 \scriptstyle \pm 0.007$ 
& $1898.472 \scriptstyle \pm 3.783$ 
& $0.327 \scriptstyle \pm 0.002$
\\
LDVI
& N/A 
& N/A 
& $5038.420 \scriptstyle \pm 73.77$ 
& $0.043 \scriptstyle \pm 0.000$
& $2919.688 \scriptstyle \pm 103.4$ 
& $0.182 \scriptstyle \pm 0.003$ 
& $154.167 \scriptstyle \pm 0.816$ 
& $0.133 \scriptstyle \pm 0.000$ 
& $3432.724 \scriptstyle \pm 406.2$ 
& $0.284 \scriptstyle \pm 0.016$ 
\\
PIS
& N/A 
& N/A 
& $10495.164 \scriptstyle \pm 83.20$ 
& $0.083 \scriptstyle \pm 0.000$ 
& $2113.172 \scriptstyle \pm 31.17$ 
& $0.218 \scriptstyle \pm 0.007$ 
& $186.007 \scriptstyle \pm 0.466$ 
& $0.193 \scriptstyle \pm 0.001$ 
& $1484.598 \scriptstyle \pm 5.125$ 
& $0.240 \scriptstyle \pm 0.000$ 
\\
DIS
& $118.947 \scriptstyle \pm 12.81$ 
& $0.159 \scriptstyle \pm 0.036$ 
& $\mathbf{3044.733 \scriptstyle \pm 464.7}$ 
& $\mathbf{0.034 \scriptstyle \pm 0.003}$ 
& $2200.590 \scriptstyle \pm 18.73$ 
& $0.155 \scriptstyle \pm 0.001$ 
& $220.392 \scriptstyle \pm 11.69$ 
& $0.194 \scriptstyle \pm 0.011$ 
& $3927.754 \scriptstyle \pm 858.9$ 
& $0.282 \scriptstyle \pm 0.019$ 
\\
DDS
& $142.890 \scriptstyle \pm 9.552$ 
& $0.172 \scriptstyle \pm 0.031$ 
& $5551.107 \scriptstyle \pm 116.4$ 
& $0.046 \scriptstyle \pm 0.001$ 
& $2154.884 \scriptstyle \pm 3.861$ 
& $0.131 \scriptstyle \pm 0.001$ 
& $188.789 \scriptstyle \pm 2.297$ 
& $0.173 \scriptstyle \pm 0.003$ 
& $1811.685 \scriptstyle \pm 24.47$ 
& $\mathbf{0.208 \scriptstyle \pm 0.006}$ 
\\
GBS
& $178.075 \scriptstyle \pm 0.103$ 
& $0.305 \scriptstyle \pm 0.002$ 
& $5080.413 \scriptstyle \pm 125.8$ 
& $0.043 \scriptstyle \pm 0.001$ 
& $5722.074 \scriptstyle \pm 22.71$ 
& $0.232 \scriptstyle \pm 0.000$ 
& $186.436 \scriptstyle \pm 1.834$ 
& $0.176 \scriptstyle \pm 0.005$ 
& $\mathbf{1137.399 \scriptstyle \pm 1.819}$ 
& $0.246 \scriptstyle \pm 0.003$ 
\\
\midrule
& {${\Delta \log Z}_{r} \downarrow$}
& {${\Delta \log Z}_{f} \downarrow$}
& {${\Delta \log Z}_{r} \downarrow$}
& {${\Delta \log Z}_{f} \downarrow$}
& {${\Delta \log Z}_{r} \downarrow$}
& {${\Delta \log Z}_{f} \downarrow$}
& {${\Delta \log Z}_{r} \downarrow$}
& {${\Delta \log Z}_{f} \downarrow$}
& {${\Delta \log Z}_{r} \downarrow$}
& {${\Delta \log Z}_{f} \downarrow$}
\\
\midrule
MFVI
& $0.612 \scriptstyle \pm 0.101$
& $0.036 \scriptstyle \pm 0.001$
& $3.658 \scriptstyle \pm 0.040$
& $0.185 \scriptstyle \pm 0.002$
& $3.009 \scriptstyle \pm 0.291$
& $\mathbf{0.048 \scriptstyle \pm 0.002}$
& $7.388 \scriptstyle \pm 0.107$
& $5.866 \scriptstyle \pm 0.016$
& $34.389 \scriptstyle \pm 0.757$
& $108.379 \scriptstyle \pm 0.438$
\\
GMMVI
& $\mathbf{0.001 \scriptstyle \pm 0.000}$
& $\mathbf{0.001 \scriptstyle \pm 0.000}$
& $\mathbf{1.715 \scriptstyle \pm 0.119}$
& $\mathbf{0.048 \scriptstyle \pm 0.007}$
& $1.282 \scriptstyle \pm 0.221$
& $0.084 \scriptstyle \pm 0.055$
& $3.098 \scriptstyle \pm 0.140$
& $\mathbf{0.124 \scriptstyle \pm 0.079}$
& $\mathbf{8.099 \scriptstyle \pm 1.919}$
& $\mathbf{11.676 \scriptstyle \pm 4.041}$
\\
SMC
& $0.187 \scriptstyle \pm 0.054$
& $2.676 \scriptstyle \pm 0.000$
& $690.721 \scriptstyle \pm 11.21$
& $161.796 \scriptstyle \pm 0.000$
& $3.880 \scriptstyle \pm 1.105$
& $80.992 \scriptstyle \pm 0.000$
& $80.184 \scriptstyle \pm 0.162$
& $375.676 \scriptstyle \pm 0.000$
& $11742.014 \scriptstyle \pm 139.2$
& $1530.824 \scriptstyle \pm 0.000$
\\
AFT
& $0.181 \scriptstyle \pm 0.106$
& $414.619 \scriptstyle \pm 141.5$
& $765.624 \scriptstyle \pm 108.0$
& $110.955 \scriptstyle \pm 18.37$
& $4.081 \scriptstyle \pm 1.579$
& $205.297 \scriptstyle \pm 23.91$
& $16.726 \scriptstyle \pm 2.511$
& $163.871 \scriptstyle \pm 6.557$
& $11653.343 \scriptstyle \pm 1628.$
& $1071.777 \scriptstyle \pm 9.475$
\\
CRAFT
& $0.091 \scriptstyle \pm 0.018$
& $255.046 \scriptstyle \pm 7.478$
& $337.094 \scriptstyle \pm 9.296$
& $100.987 \scriptstyle \pm 0.065$
& $\mathbf{0.822 \scriptstyle \pm 0.087}$
& $210.245 \scriptstyle \pm 6.098$
& $1.458 \scriptstyle \pm 0.406$
& $63.792 \scriptstyle \pm 3.329$
& $445.101 \scriptstyle \pm 8.273$
& $1156.718 \scriptstyle \pm 7.810$
\\
FAB
& $\mathbf{0.001 \scriptstyle \pm 0.000}$
& $0.019 \scriptstyle \pm 0.003$
& $2.952 \scriptstyle \pm 0.247$
& $126.363 \scriptstyle \pm 1.789$
& $3.358 \scriptstyle \pm 1.062$
& $84.592 \scriptstyle \pm 22.64$
& $\mathbf{0.847 \scriptstyle \pm 0.076}$
& $63.910 \scriptstyle \pm 1.565$
& $350.544 \scriptstyle \pm 599.0$
& $3721.720 \scriptstyle \pm 4646.$
\\
MCD
& $0.207 \scriptstyle \pm 0.039$
& N/A
& $31.319 \scriptstyle \pm 1.793$
& $21.148 \scriptstyle \pm 1.478$
& $28.607 \scriptstyle \pm 1.275$
& $24.757 \scriptstyle \pm 0.841$
& $884.610 \scriptstyle \pm 9.674$
& $258.840 \scriptstyle \pm 3.047$
& $15122.090 \scriptstyle \pm 996.7$
& $1125.475 \scriptstyle \pm 5.198$
\\
LDVI
& N/A
& N/A
& $8.159 \scriptstyle \pm 0.775$
& $15.477 \scriptstyle \pm 0.815$
& $4.360 \scriptstyle \pm 0.741$
& $5.472 \scriptstyle \pm 0.938$
& $537.763 \scriptstyle \pm 25.07$
& $265.674 \scriptstyle \pm 1.181$
& $12237.989 \scriptstyle \pm 381.8$
& $1087.592 \scriptstyle \pm 4.844$
\\
PIS
& $0.918 \scriptstyle \pm 0.598$
& $0.436 \scriptstyle \pm 0.002$
& $7.122 \scriptstyle \pm 0.630$
& $3113.492 \scriptstyle \pm 1.978$
& $12.248 \scriptstyle \pm 0.326$
& $54.090 \scriptstyle \pm 0.151$
& $104.002 \scriptstyle \pm 0.847$
& $2149.224 \scriptstyle \pm 19.39$
& $1884.013 \scriptstyle \pm 10.20$
& $8785.873 \scriptstyle \pm 9.880$
\\
DIS
& $0.113 \scriptstyle \pm 0.083$
& $25.544 \scriptstyle \pm 8.267$
& $87.709 \scriptstyle \pm 8.942$
& $369.352 \scriptstyle \pm 16.29$
& $10.448 \scriptstyle \pm 0.607$
& $87.897 \scriptstyle \pm 5.255$
& $569.837 \scriptstyle \pm 35.40$
& $1354.472 \scriptstyle \pm 181.1$
& $8807.430 \scriptstyle \pm 337.6$
& $17566.520 \scriptstyle \pm 256.6$
\\
DDS
& $0.190 \scriptstyle \pm 0.077$
& $0.321 \scriptstyle \pm 0.052$
& $1.739 \scriptstyle \pm 0.442$
& $207.545 \scriptstyle \pm 1.163$
& $7.952 \scriptstyle \pm 0.299$
& $53.411 \scriptstyle \pm 0.024$
& $82.460 \scriptstyle \pm 5.480$
& $659.497 \scriptstyle \pm 9.786$
& $1579.602 \scriptstyle \pm 41.65$
& $2910.345 \scriptstyle \pm 71.25$
\\
GBS
& $0.553 \scriptstyle \pm 0.273$
& $0.127 \scriptstyle \pm 0.008$
& $8.103 \scriptstyle \pm 1.696$
& $9.321 \scriptstyle \pm 0.776$
& $53.767 \scriptstyle \pm 0.732$
& $47.441 \scriptstyle \pm 0.098$
& $75.160 \scriptstyle \pm 2.321$
& $62.733 \scriptstyle \pm 1.168$
& $1495.194 \scriptstyle \pm 42.03$
& $527.580 \scriptstyle \pm 9.426$
\\
\midrule
& {{ELBO} $\uparrow$}
& {{EUBO} $\downarrow$}
& {{ELBO} $\uparrow$}
& {{EUBO} $\downarrow$}
& {{ELBO} $\uparrow$}
& {{EUBO} $\downarrow$}
& {{ELBO} $\uparrow$}
& {{EUBO} $\downarrow$}
& {{ELBO} $\uparrow$}
& {{EUBO} $\downarrow$}
\\
\midrule
MFVI
& $-1.834 \scriptstyle \pm 0.009$
& $105.694 \scriptstyle \pm 0.002$
& $-3.690 \scriptstyle \pm 0.000$
& $164.114 \scriptstyle \pm 0.000$
& $-5.957 \scriptstyle \pm 0.007$
& $72.663 \scriptstyle \pm 0.005$
& $-14.004 \scriptstyle \pm 0.005$
& $210.713 \scriptstyle \pm 0.024$
& $-58.082 \scriptstyle \pm 0.009$
& $938.632 \scriptstyle \pm 0.055$
\\
GMMVI
& $\mathbf{-0.011 \scriptstyle \pm 0.001}$
& $\mathbf{0.012 \scriptstyle \pm 0.001}$
& $\mathbf{-1.715 \scriptstyle \pm 0.119}$
& $240.459 \scriptstyle \pm 51.13$
& $-3.890 \scriptstyle \pm 0.122$
& $57.746 \scriptstyle \pm 1.928$
& $\mathbf{-7.135 \scriptstyle \pm 0.148}$
& $142.636 \scriptstyle \pm 9.701$
& $\mathbf{-18.478 \scriptstyle \pm 4.104}$
& $595.239 \scriptstyle \pm 120.4$
\\
SMC
& $-0.242 \scriptstyle \pm 0.047$
& $4.690 \scriptstyle \pm 0.000$
& $-877.034 \scriptstyle \pm 10.23$
& $161.921 \scriptstyle \pm 0.000$
& $-4.634 \scriptstyle \pm 1.088$
& $81.325 \scriptstyle \pm 0.000$
& $-185.057 \scriptstyle \pm 0.257$
& $376.093 \scriptstyle \pm 0.000$
& $-12187.873 \scriptstyle \pm 134.6$
& $1532.904 \scriptstyle \pm 0.000$
\\
AFT
& $-0.293 \scriptstyle \pm 0.088$
& $431.329 \scriptstyle \pm 143.1$
& $-927.160 \scriptstyle \pm 103.8$
& $117.630 \scriptstyle \pm 22.16$
& $-4.923 \scriptstyle \pm 1.546$
& $207.625 \scriptstyle \pm 24.14$
& $-64.442 \scriptstyle \pm 4.464$
& $214.486 \scriptstyle \pm 4.870$
& $-11828.529 \scriptstyle \pm 1608.$
& $1448.335 \scriptstyle \pm 11.08$
\\
CRAFT
& $-0.027 \scriptstyle \pm 0.060$
& $263.474 \scriptstyle \pm 7.864$
& $-451.399 \scriptstyle \pm 7.561$
& $103.674 \scriptstyle \pm 0.069$
& $\mathbf{-0.295 \scriptstyle \pm 0.256}$
& $212.210 \scriptstyle \pm 6.160$
& $-11.154 \scriptstyle \pm 0.307$
& $89.518 \scriptstyle \pm 1.904$
& $-520.475 \scriptstyle \pm 5.531$
& $1578.114 \scriptstyle \pm 2.360$
\\
FAB
& $\mathbf{-0.014 \scriptstyle \pm 0.003}$
& $\mathbf{0.012 \scriptstyle \pm 0.002}$
& $-299.916 \scriptstyle \pm 253.4$
& $93.560 \scriptstyle \pm 5.086$
& $-26.496 \scriptstyle \pm 1.875$
& $\mathbf{18.088 \scriptstyle \pm 2.503}$
& $-11.396 \scriptstyle \pm 0.153$
& $\mathbf{12.084 \scriptstyle \pm 0.171}$
& $-892.971 \scriptstyle \pm 1518.$
& $\mathbf{394.346 \scriptstyle \pm 263.6}$
\\
MCD
& $-0.611 \scriptstyle \pm 0.005$
& N/A
& $-185.021 \scriptstyle \pm 0.743$
& $\mathbf{43.670 \scriptstyle \pm 0.457}$
& $-69.358 \scriptstyle \pm 0.633$
& $47.834 \scriptstyle \pm 0.820$
& $-1457.646 \scriptstyle \pm 13.80$
& $293.191 \scriptstyle \pm 0.208$
& $-21196.583 \scriptstyle \pm 472.8$
& $1276.456 \scriptstyle \pm 1.033$
\\
LDVI
& N/A
& N/A
& $-29.034 \scriptstyle \pm 0.591$
& $51.137 \scriptstyle \pm 0.177$
& $-28.471 \scriptstyle \pm 1.018$
& $20.887 \scriptstyle \pm 1.042$
& $-875.104 \scriptstyle \pm 43.59$
& $323.158 \scriptstyle \pm 0.142$
& $-16227.975 \scriptstyle \pm 738.0$
& $1185.331 \scriptstyle \pm 6.660$
\\
PIS
& $-3.198 \scriptstyle \pm 0.042$
& $104.975 \scriptstyle \pm 0.002$
& $-16.881 \scriptstyle \pm 0.026$
& $3626.120 \scriptstyle \pm 1.360$
& $-29.261 \scriptstyle \pm 1.743$
& $88.192 \scriptstyle \pm 0.005$
& $-172.988 \scriptstyle \pm 0.630$
& $2748.938 \scriptstyle \pm 19.19$
& $-2988.210 \scriptstyle \pm 14.13$
& $11179.374 \scriptstyle \pm 11.72$
\\
DIS
& $-1.021 \scriptstyle \pm 0.436$
& $40.892 \scriptstyle \pm 38.48$
& $-181.348 \scriptstyle \pm 15.47$
& $546.335 \scriptstyle \pm 30.86$
& $-36.704 \scriptstyle \pm 0.629$
& $193.270 \scriptstyle \pm 3.293$
& $-840.122 \scriptstyle \pm 18.66$
& $1745.719 \scriptstyle \pm 205.7$
& $-15337.229 \scriptstyle \pm 154.0$
& $20347.781 \scriptstyle \pm 318.7$
\\
DDS
& $-0.597 \scriptstyle \pm 0.142$
& $148.841 \scriptstyle \pm 7.347$
& $-13.284 \scriptstyle \pm 0.460$
& $291.867 \scriptstyle \pm 0.047$
& $-31.681 \scriptstyle \pm 0.363$
& $86.014 \scriptstyle \pm 0.001$
& $-156.145 \scriptstyle \pm 6.063$
& $881.476 \scriptstyle \pm 22.83$
& $-2617.761 \scriptstyle \pm 46.78$
& $3925.231 \scriptstyle \pm 106.8$
\\
GBS
& $-2.600 \scriptstyle \pm 0.078$
& $110.167 \scriptstyle \pm 0.000$
& $-35.771 \scriptstyle \pm 1.105$
& $67.819 \scriptstyle \pm 2.157$
& $-99.369 \scriptstyle \pm 0.158$
& $73.545 \scriptstyle \pm 0.107$
& $-154.186 \scriptstyle \pm 1.387$
& $106.777 \scriptstyle \pm 0.113$
& $-2198.997 \scriptstyle \pm 36.56$
& $705.996 \scriptstyle \pm 11.66$
\\
\bottomrule
\end{tabular}
}
\end{sc}
\end{small}
\end{center}
    \caption{Results for various sampling methods. Evaluation criteria include 2-Wasserstein distance ($\mathcal{W}_2$), maximum mean discrepancy (MMD), reverse and forward partition function error ($\Delta \log Z_{r}$, $\Delta \log Z_{f}$), and lower and upper evidence bounds (ELBO, EUBO). The best results are highlighted in bold. Arrows ($\uparrow$, $\downarrow$) indicate whether higher or lower values are preferable, respectively. N/A denotes cases where reasonable results could not be obtained due to numerical issues.
    }
\label{table:main_results}
\end{table*}
\section{Experiments}
\label{section:exp}
Here, we offer an overview of the evaluation protocol. Next, we present the results obtained for synthetic target densities, followed by those for real targets. We provide further results in Appendix \ref{appendix:exp2} and ablation studies in Appendix \ref{appendix:ablations}.

\textbf{Evaluation Protocol.} 
We compute all performance criteria 100 times during training, applying a running average with a length of 5 over these evaluations to obtain robust results within a single run. To ensure robustness across runs, we use four different random seeds and average the best results from each run. We use 2000 samples to compute the performance criteria and tune key hyperparameters such as the learning rate and variance of the initial proposal distribution $\target_0$. We report the EMC values corresponding to the method's highest ELBO value to avoid high EMC values caused by a large initial support of the model.

\begin{table*}
\begin{minipage}{\textwidth}

\begin{minipage}[t]{0.72\textwidth}
\centering
 \resizebox{\textwidth}{!}{%
\begin{tabular}{ccc|ccc|c}
\includegraphics[width=0.32\textwidth]{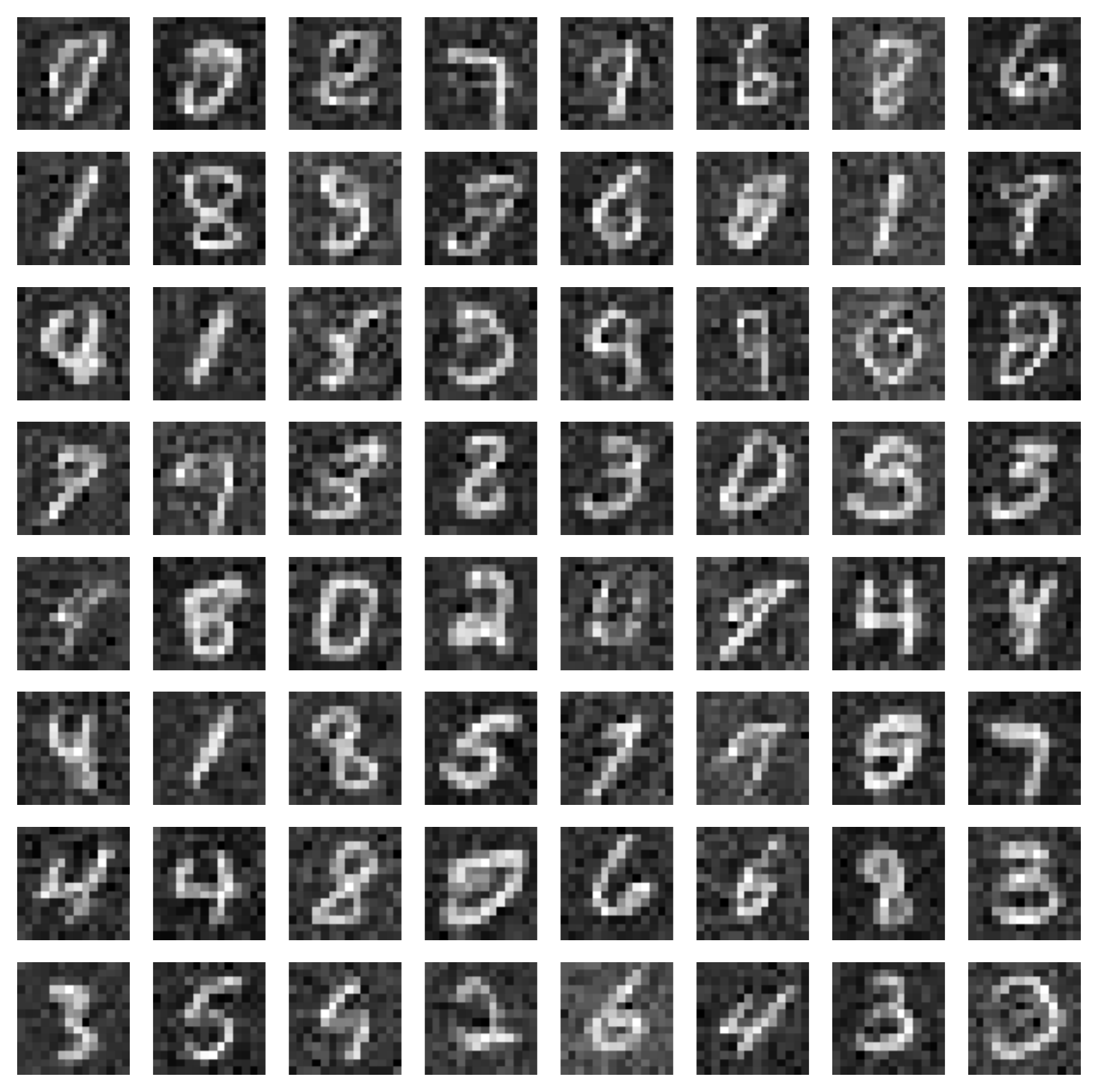} &   
\includegraphics[width=0.32\textwidth]{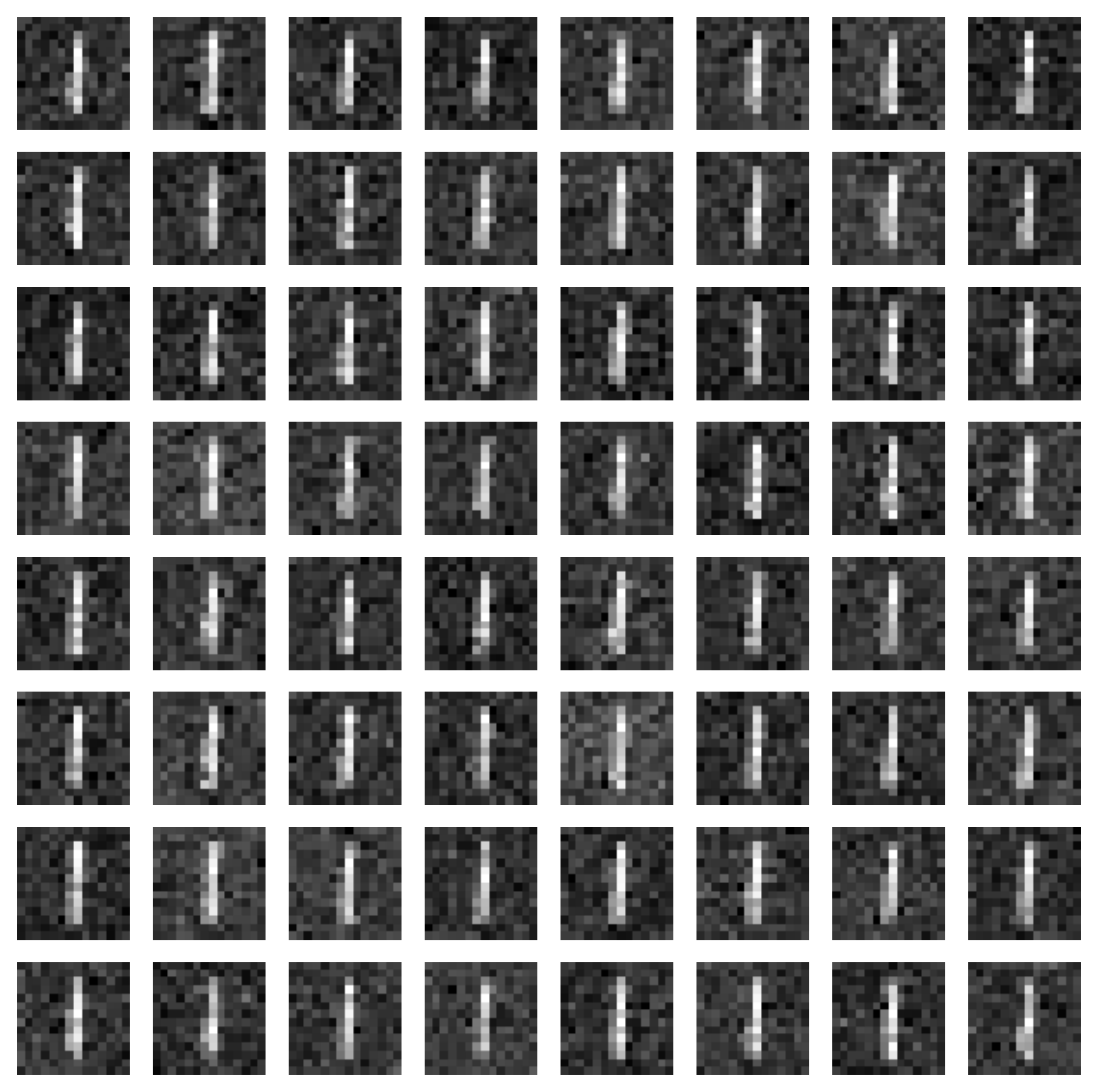} &   
\includegraphics[width=0.32\textwidth]{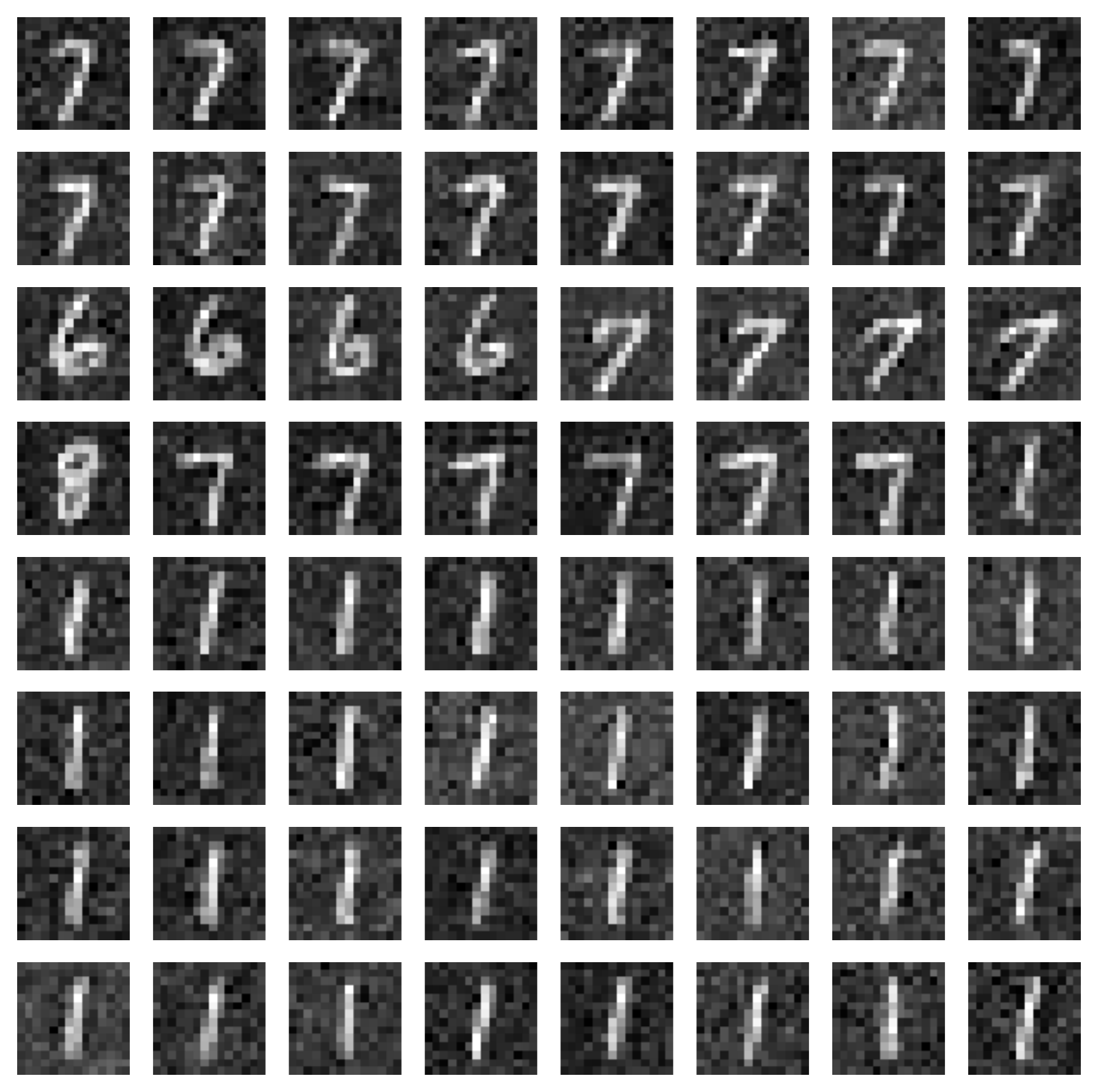} &   
\includegraphics[width=0.32\textwidth]{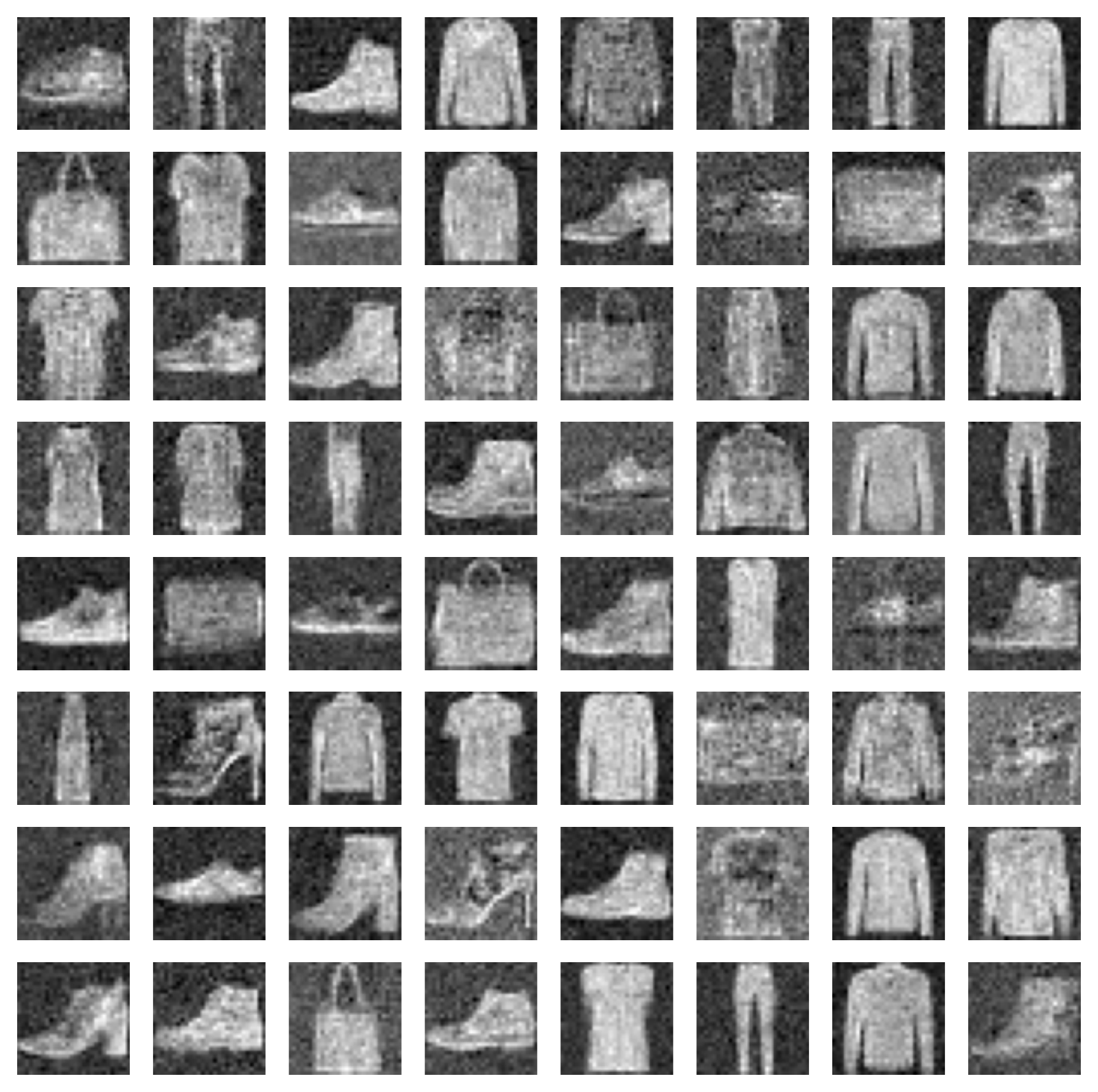} &   
\includegraphics[width=0.32\textwidth]{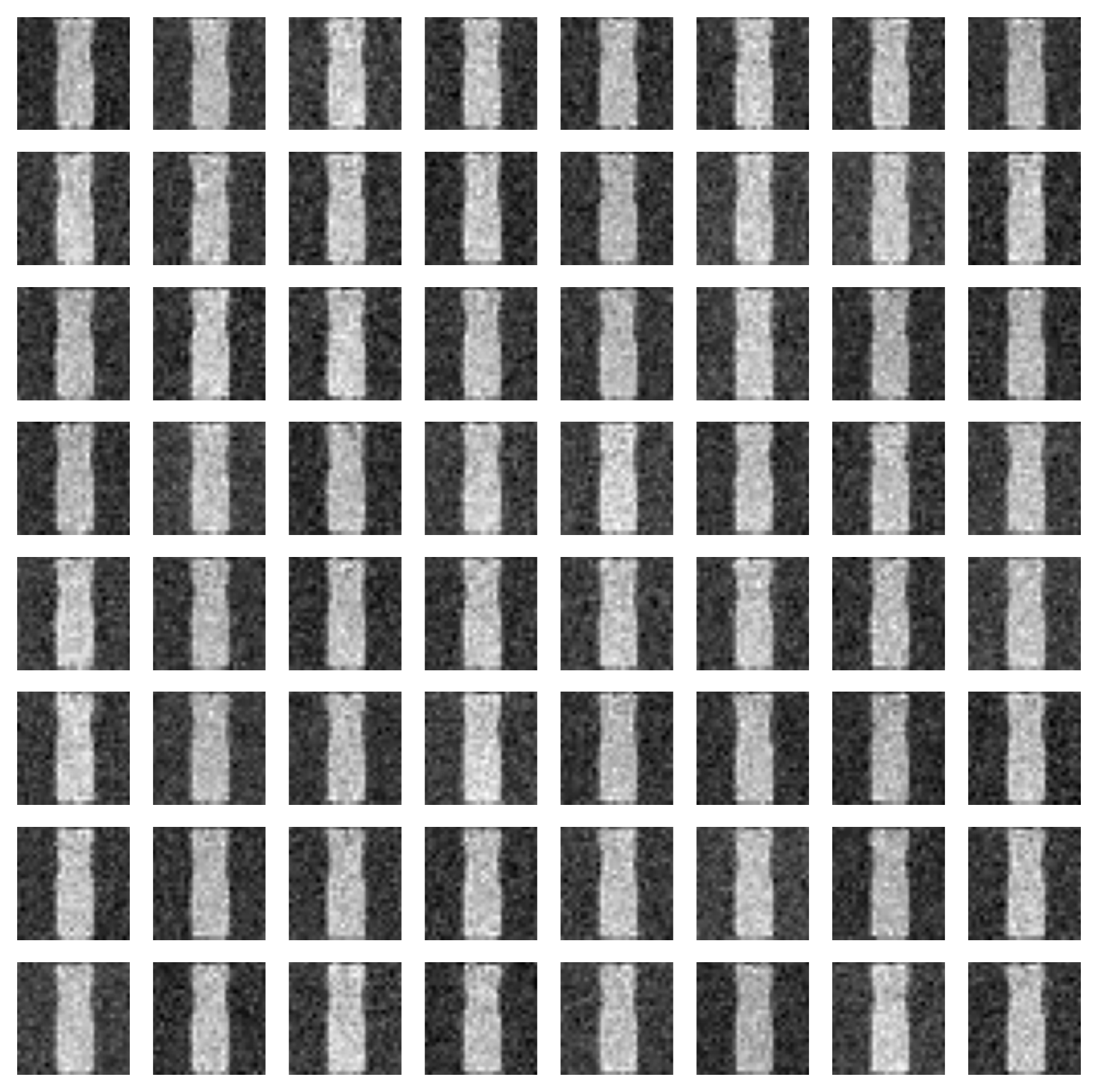} &   
\includegraphics[width=0.32\textwidth]{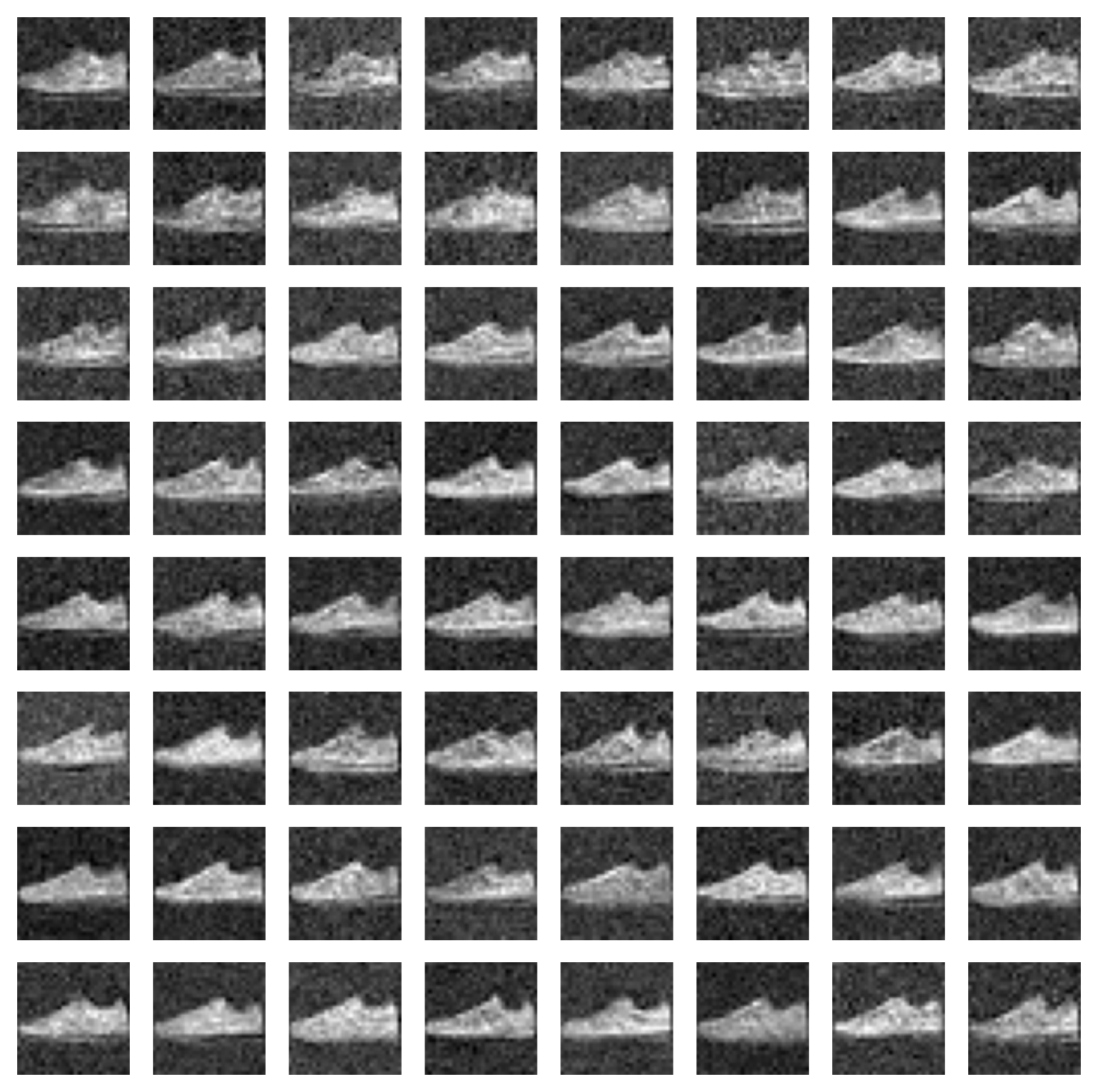} &   
\\
\small $\x_i \sim \target$ &   
\small MFVI &   
\small GMMVI &   
\small $\x_i \sim \target$ &   
\small MFVI &   
\small GMMVI &   
\\
\includegraphics[width=0.32\textwidth]{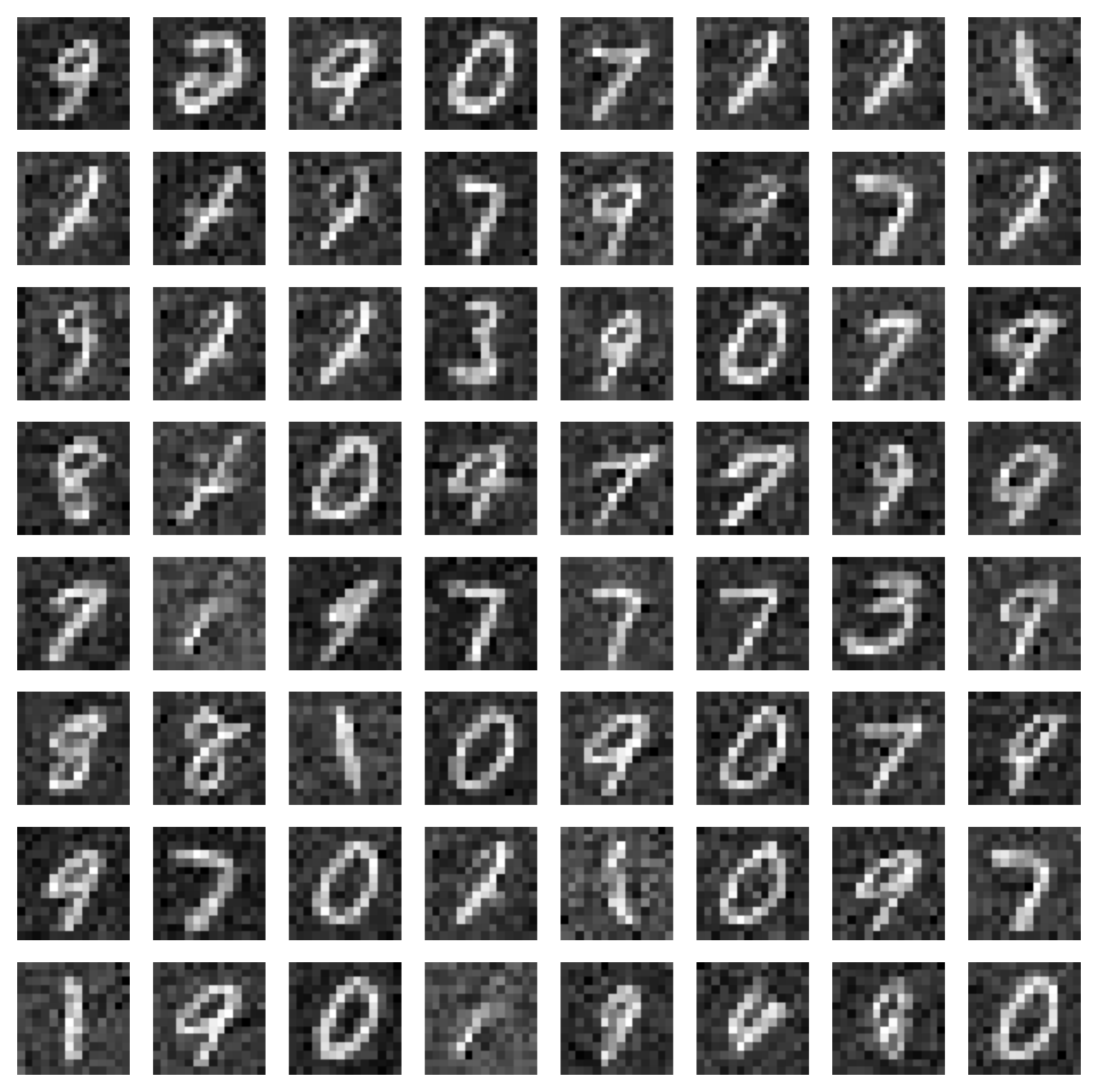} &   
\includegraphics[width=0.32\textwidth]{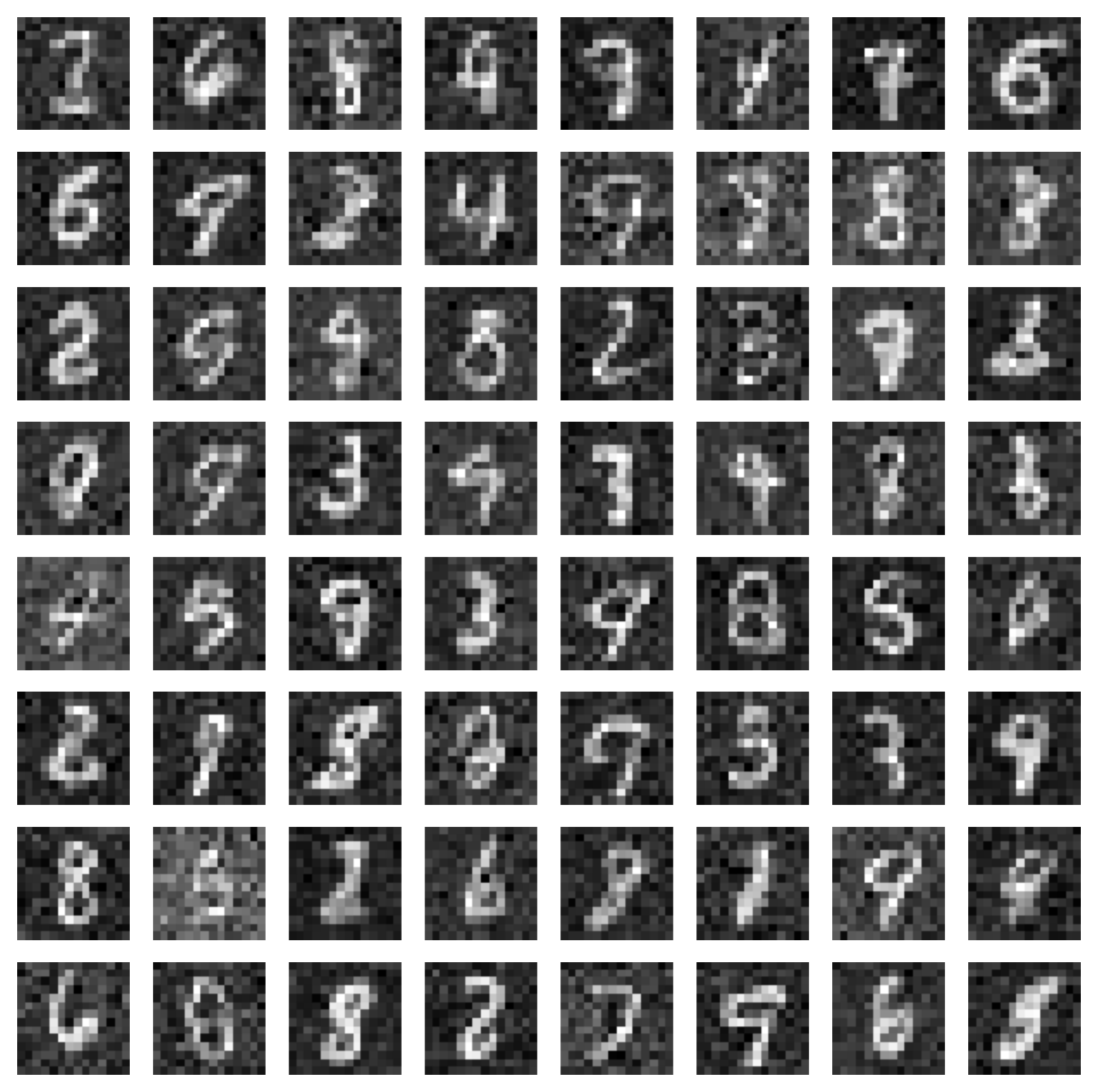} &   
\includegraphics[width=0.32\textwidth]{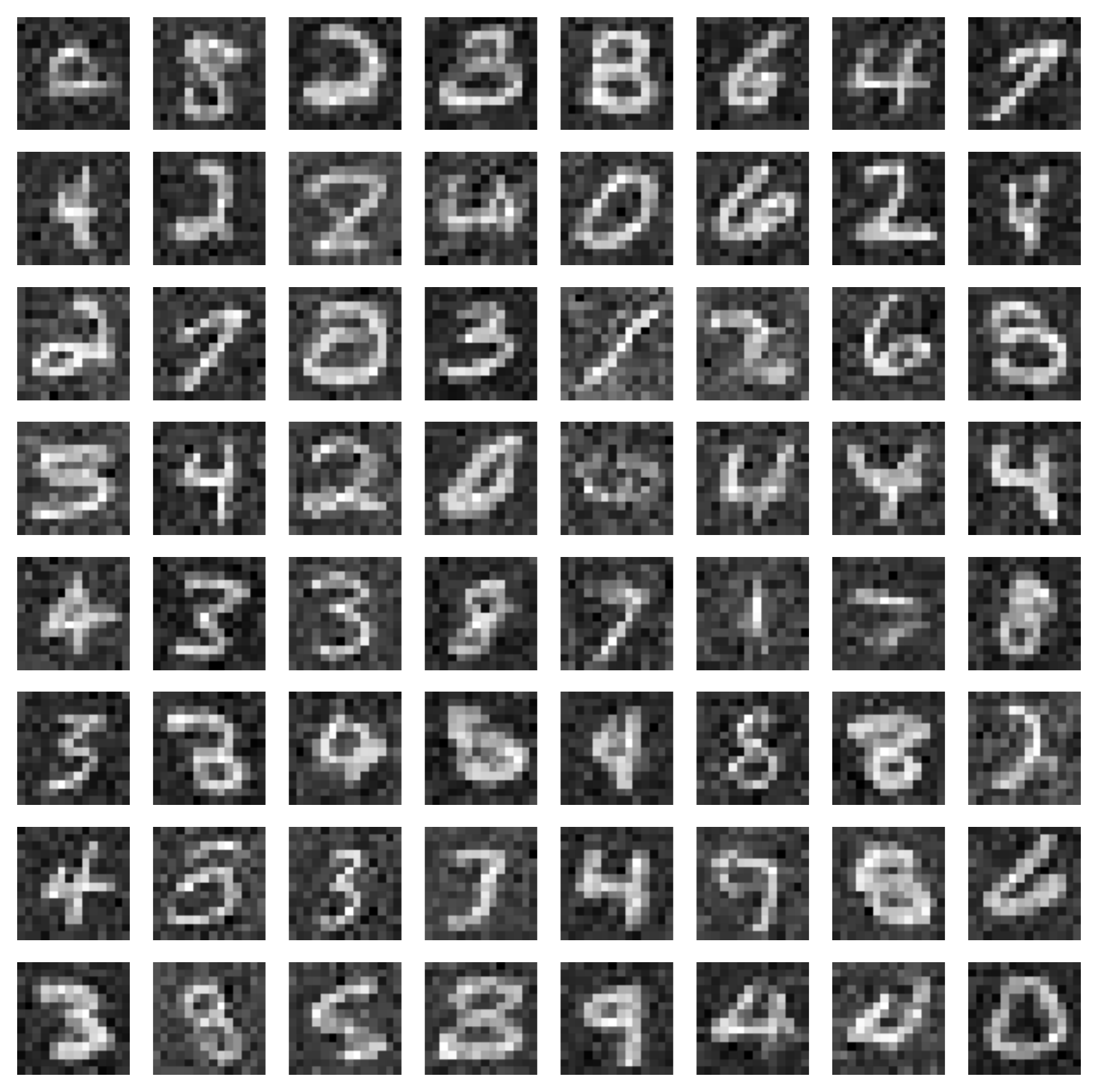} &   
\includegraphics[width=0.32\textwidth]{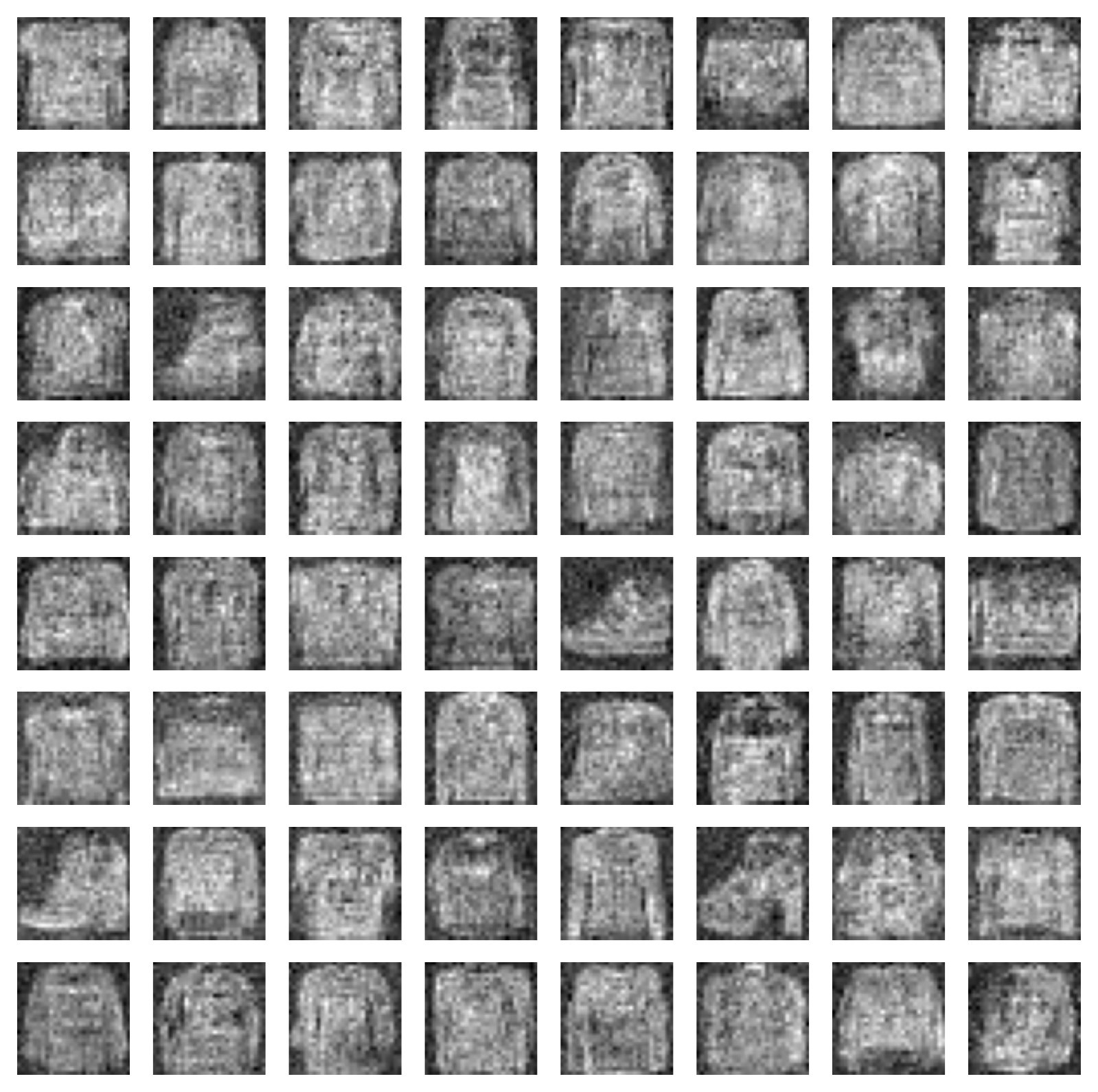} &   
\includegraphics[width=0.32\textwidth]{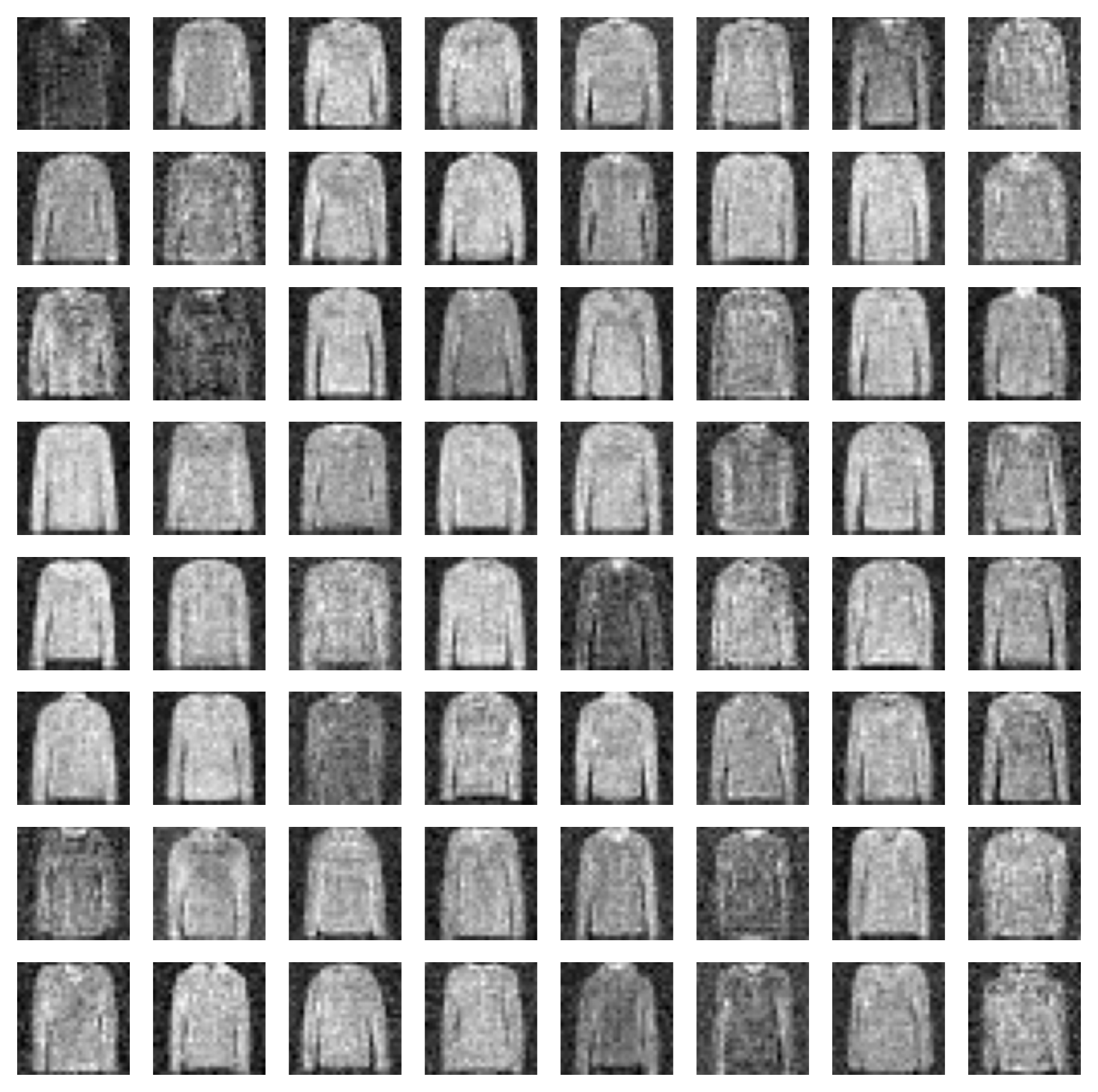} &   
\includegraphics[width=0.32\textwidth]{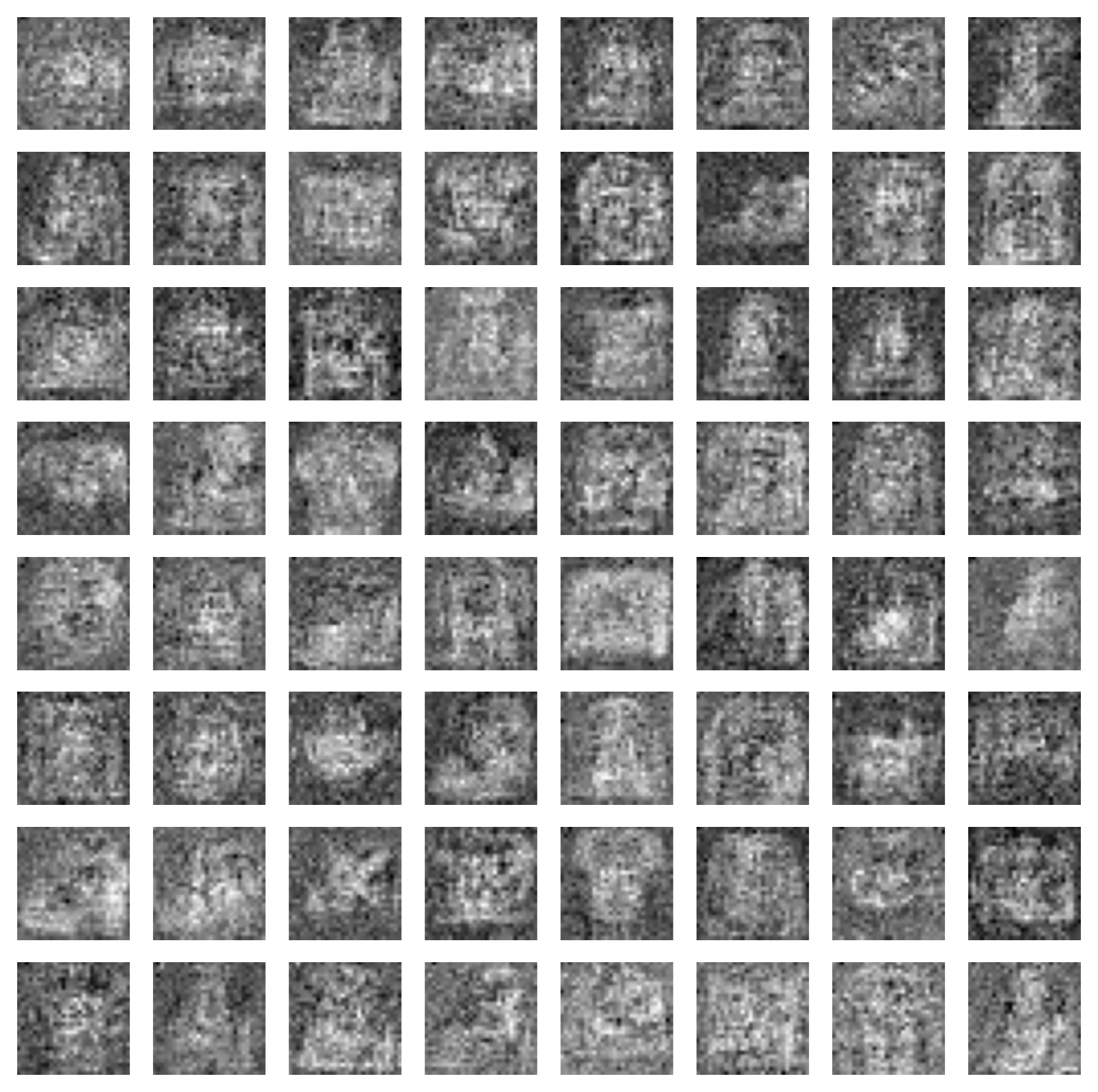} &   
\\
\small CRAFT &   
\small FAB &   
\small LDVI &   
\small GBS &   
\small FAB &   
\small LDVI &   
\\
\end{tabular}
}
\end{minipage}
\hfill
\begin{minipage}[t]{0.27\textwidth}
\centering
\resizebox{\textwidth}{!}{%
\renewcommand{\arraystretch}{1.2}
\begin{tabular}{l|rr}
\toprule
EMC $\uparrow$
& $14 \times 14$ \textbf{Digits}
& $28 \times 28$ \textbf{Fashion}
\\ 
\midrule
MFVI
& $0.000 \scriptstyle \pm 0.000$
& $0.000 \scriptstyle \pm 0.000$
\\
GMMVI
& $0.164 \scriptstyle \pm 0.081$
& $0.217 \scriptstyle \pm 0.167$
\\
SMC
& $0.873 \scriptstyle \pm 0.000$
& $0.000 \scriptstyle \pm 0.000$
\\
AFT
& $0.727 \scriptstyle \pm 0.000$
& $0.011 \scriptstyle \pm 0.000$
\\
CRAFT
& $0.772 \scriptstyle \pm 0.070$
& $0.016 \scriptstyle \pm 0.027$
\\
FAB
& $0.915 \scriptstyle \pm 0.007$
& $0.349 \scriptstyle \pm 0.137$
\\
MCD
& $0.851 \scriptstyle \pm 0.010$
& $0.619 \scriptstyle \pm 0.001$
\\
LDVI
& $\mathbf{0.951 \scriptstyle \pm 0.002}$
& $0.608 \scriptstyle \pm 0.005$
\\
PIS
& $0.816 \scriptstyle \pm 0.011$
& $0.620 \scriptstyle \pm 0.004$
\\
DIS
& $0.818 \scriptstyle \pm 0.009$
& $0.612 \scriptstyle \pm 0.008$
\\
DDS
& $0.816 \scriptstyle \pm 0.012$
& $0.621 \scriptstyle \pm 0.008$
\\
GBS
& $0.796 \scriptstyle \pm 0.005$
& $\mathbf{0.621 \scriptstyle \pm 0.006}$
\\
\bottomrule
\end{tabular}
}
\end{minipage}
\captionof{table}{
Sample visualizations for \textit{Digits} (left) and \textit{Fashion} (middle) using various methods, as indicated by the subcaptions. `$\x_i \sim \target$' refers to samples from the target density. Visualizations for the remaining methods are provided in Figure \ref{fig:nice}. Corresponding EMC values are reported on the right.
}
\label{tab:nice}
\end{minipage}

\vspace{-0.5cm}
\end{table*}

\subsection{Evaluation on Synthetic Target Densities}
\textbf{Funnel.} We utilize the funnel distribution as a testing ground to assess whether sampling methods capture high curvatures in the target density. Our findings indicate that while most methods successfully capture the funnel-like structure, they struggle to generate samples at the neck and opening of the funnel, except for FAB and GMMVI (cf. Figure \ref{fig:samples_mog_funnel}). This observation is further supported by quantitative analysis, revealing that both FAB and GMMVI achieve the best performance in terms of reverse and forward estimation of $\log Z$ and evidence bounds as shown in Table \ref{table:main_results}.


\textbf{Digits and Fashion.} 
For a comprehensive assessment of sampling methods, we conduct both qualitative and quantitative analyses on two high-dimensional target densities. For the qualitative analysis, model samples are interpreted as images and shown in Table \ref{tab:nice}. For the quantitative analysis, we report various performance criteria values, with results presented in Table \ref{table:main_results}. Additionally, we report EMC values in Table \ref{tab:nice} to quantify mode collapse.

For \textit{Digits}, most methods are able to find the majority of modes and produce high-quality samples, as visually evident from the sample visualizations and EMC values in Table \ref{tab:nice}. However, many methods, particularly diffusion-based ones, struggle to obtain reasonable estimations of $\log Z$. They also perform poorly in terms of lower and upper evidence bounds, as shown in Table \ref{table:main_results}.
For \textit{Fashion}, we observe that methods either suffer from mode collapse or produce low-quality samples. Interestingly, the methods experiencing mode collapse achieve the lowest estimation error of $\log Z$ in both reverse and forward estimations.

\textbf{Mixture Models.} We employ \textit{MoG} and \textit{MoS} to investigate mode collapse across different dimensions, specifically considering $d \in {2, 50, 200}$. For $d=2$, all methods except MFVI demonstrate the capability to generate samples from all modes, as indicated by $\text{EMC} \approx 1$. This is further supported by visualizations in Figure \ref{fig:dim}. According to EMC, all methods except diffusion-based ones exhibit mode collapse for $d=50$ and $d=200$.

We also report additional evaluation criteria for \textit{MoG} and \textit{MoS}, including 2-Wasserstein distance, maximum mean discrepancy, reverse and forward partition function error, lower and upper evidence bounds, and reverse and forward effective sample size in Appendix \ref{appendix:exp2} Table \ref{tab:dim}.

\subsection{Evaluation on Real Target Densities}
For real-world target densities, we do not have access to the ground truth normalizer $Z$ or samples from $\pi$. Consequently, we present the ELBO values in Table \ref{table:elbos}. Surprisingly, we find that GMMVI performs well across all tasks, often outperforming more complex variational Monte Carlo methods. However, it is noteworthy that GMMVI encounters scalability challenges in very high-dimensional problems, such as LGCP. Another method, FAB, consistently performs well across a majority of tasks.

\begin{table*}[t!]
\begin{center}
\begin{small}
\begin{sc}
\resizebox{\textwidth}{!}{%
\renewcommand{\arraystretch}{1.2}
\begin{tabular}{l|rrrrrrr}
\toprule
  {{ELBO} $\uparrow$}   & \textbf{Credit} & \textbf{Seeds} & \textbf{Cancer}  &  \textbf{Brownian}  & \textbf{Ionosphere} & \textbf{Sonar}  & \textbf{LGCP} \\ 
\midrule
MFVI
& $-524.859 \scriptstyle \pm 0.035$ 
& $-76.733 \scriptstyle \pm 0.012$ 
& $-29.407 \scriptstyle{\pm 0.557}$ 
& $-3.872 \scriptstyle \pm 0.012$ 
& $-123.419 \scriptstyle \pm 0.040$ 
& $-137.672 \scriptstyle \pm 0.043$ 
& $383.18 \scriptstyle \pm 0.059$ 
\\
GMMVI
& $\mathbf{-504.487 \scriptstyle \pm 0.001}$ 
& $\mathbf{-73.415 \scriptstyle \pm 0.002}$ 
& $\mathbf{121.442 \scriptstyle \pm 5.591}$ 
& $\mathbf{1.092 \scriptstyle \pm 0.006}$ 
& $-111.832 \scriptstyle \pm 0.007$ 
& $-108.726 \scriptstyle \pm 0.007$ 
& OOM 
\\
SMC
& $-580.936 \scriptstyle \pm 15.915$ 
& $-74.699 \scriptstyle \pm 0.100$ 
& $-67.959 \scriptstyle \pm 4.345$ 
& $-1.874 \scriptstyle \pm 0.622$ 
& $-114.751 \scriptstyle \pm 0.238$ 
& $-111.355 \scriptstyle \pm 1.177$ 
& $393.907 \scriptstyle \pm 5.727$ 
\\
AFT
& $-584.766 \scriptstyle \pm 13.979$ 
& $-74.269 \scriptstyle \pm 0.090$ 
& $-15.515 \scriptstyle \pm 5.100$ 
& N/A 
& $-113.272 \scriptstyle \pm 0.647$ 
& $-110.671 \scriptstyle \pm 1.240$ 
& $394.271 \scriptstyle \pm 6.432$ 
\\
CRAFT
& $-573.387 \scriptstyle \pm 17.59$ 
& $-73.793 \scriptstyle \pm 0.015$ 
& $19.283 \scriptstyle \pm 0.523$ 
& $0.886 \scriptstyle \pm 0.053$ 
& $-112.386 \scriptstyle \pm 0.182$ 
& $-115.618 \scriptstyle \pm 1.316$ 
& $\mathbf{495.291 \scriptstyle \pm 0.384}$ 
\\
FAB
& $\mathbf{-504.496 \scriptstyle \pm 0.001}$ 
& $\mathbf{-73.418 \scriptstyle \pm 0.002}$ 
& $39.922 \scriptstyle \pm 8.200$ 
& $1.031 \scriptstyle \pm 0.010$ 
& $\mathbf{-111.678 \scriptstyle \pm 0.003}$ 
& $\mathbf{-108.593 \scriptstyle \pm 0.008}$ 
& $402.212 \scriptstyle \pm 0.941$ 
\\
MCD
& N/A 
& $-73.652 \scriptstyle \pm 0.003$ 
& N/A
& $0.643 \scriptstyle \pm 0.012$ 
& $-111.942 \scriptstyle \pm 0.006$ 
& $-109.534 \scriptstyle \pm 0.055$ 
& $444.313 \scriptstyle \pm 0.452$ 
\\
LDVI
& N/A
& $-73.530 \scriptstyle \pm 0.003$ 
& N/A
& $0.772 \scriptstyle \pm 0.016$
& $-111.788 \scriptstyle \pm 0.003$ 
& $-108.841 \scriptstyle \pm 0.006$ 
& $161.839 \scriptstyle \pm 1.436$ 
\\
PIS
& $-846.568 \scriptstyle \pm 2.417$ 
& $-88.919 \scriptstyle \pm 2.051$ 
& $39.542 \scriptstyle \pm 5.302$ 
& N/A 
& $-125.030 \scriptstyle \pm 0.688$ 
& $-142.868 \scriptstyle \pm 3.289$ 
& $479.542 \scriptstyle \pm 0.403$ 
\\
DDS
& $-514.736 \scriptstyle \pm 1.223$ 
& $-75.206 \scriptstyle \pm 0.209$ 
& $19.997 \scriptstyle \pm 0.690$ 
& $0.561 \scriptstyle \pm 0.228$ 
& $-114.191 \scriptstyle \pm 0.105$ 
& $-121.222 \scriptstyle \pm 5.985$ 
& N/A 
\\
GBS
& $-508.108 \scriptstyle \pm 0.145$
& $-88.778 \scriptstyle \pm 0.109$
& $-23.495 \scriptstyle \pm 0.737$
& N/A 
& $-133.777 \scriptstyle \pm 0.152$
& $-153.094 \scriptstyle \pm 0.500$
& N/A 
\\
\bottomrule
\end{tabular}
}
\end{sc}
\end{small}
\end{center}
    \caption{ELBO values for various target densities. The best results are highlighted in bold. N/A denotes cases where reasonable results could not be obtained due to numerical issues. OOM refers to problems caused by memory constraints.
    }
\label{table:elbos}
\end{table*}

\section{Discussion and Conclusion}
\label{sec:conclusion}
Here, we list several general observations \textbf{O1)-O6)} and observations tied to specific methods \textbf{M1)-M6)} that are based on the experiments from Section \ref{section:exp} and Appendix \ref{appendix:exp2} and the Ablation studies in Appendix \ref{appendix:ablations}.

\textbf{O1)} Mode collapse gets worse in high dimensions. We observe that several methods, that do not suffer from mode collapse in low-dimensional problems encounter significant mode collapse when applied to higher-dimensional ones (cf. Fig \ref{fig:dim}).

\textbf{O2)} ELBO and reverse $\log Z$ estimates are not well suited for evaluating a model's capability to avoid mode collapse. This observation is evident, for instance, in Table \ref{tab:nice}, where MFVI achieves relatively good ELBO and $\log Z$ estimates despite suffering from mode collapse. 

\textbf{O3)} While the EUBO helps to quantify mode collapse, comparing different method categories is challenging due to the additional looseness introduced by latent variables in the extended EUBO. This is evident on the \textit{Fashion} dataset, where MFVI and GMMVI achieve a lower EUBO compared to most other methods, despite suffering from mode collapse.

\textbf{O4)} Despite being influenced by subjective design choices like the kernel or cost function, the 2-Wasserstein distance and Maximum Mean Discrepancy (MMD) generally show consistent performance across different sampling methods, as demonstrated in Table \ref{table:main_results}. Additionally, the quantitative results frequently align with the qualitative outcomes. For instance, this alignment is evident from GMMVI samples on \textit{Funnel} or the GBS samples on the \textit{Fashion}.

\textbf{O5)} For multimodal target distributions, both forward and reverse ESS tend to exhibit a 'binary' pattern, frequently taking values of 0 or 1. Forward ESS, in particular, often tends to be predominantly zero for higher dimensional problems, further complicating the evaluation of mode collapse severity. In contrast, EUBO and ELBO offer a more continuous and informative perspective in assessing model performance (cf. Appendix \ref{appendix:exp2}, Table \ref{tab:dim}).

\textbf{O6)} No single method exhibits superiority across all situations. Generally, GMMVI and FAB demonstrate good ELBO values across a diverse set of tasks, although both tend to suffer from mode collapse in high dimensions. In contrast, diffusion-based methods such as MCD and LDVI exhibit resilience against mode collapse but frequently fall short of achieving satisfactory ELBO values.

\textbf{M1)} Resampling causes mode collapse in high dimensions (cf. Ablation \ref{abl:smc_choices}). SIS methods, in particular, experience severe mode collapse in high dimensions, as illustrated in Figure \ref{fig:dim}. Notably, eliminating the resampling step in Sequential Monte Carlo (SMC) proves effective in mitigating this issue, but results in worse ELBO values.

\textbf{M2)} There exists an exploration-exploitation trade-off when setting the support of the proposal distribution $\pi_0$ in Variational Monte Carlo (cf. Ablation \ref{abl:init_support}). 
Opting for a small initial support of $\pi_0$ results in tight ELBO values but can limit coverage to only a few modes. Conversely, employing a sufficiently large initial support helps prevent mode collapse but introduces additional looseness in the ELBO. 

\textbf{M3)} Learning the proposal distribution $\pi_0$ in Variational Monte Carlo methods often leads to mode collapse, especially in high dimensions. Training the base distribution end-to-end by maximizing the extended ELBO or pre-training the base distribution, for example, using methods like MFVI, results in mode collapse, as indicated in Ablation \ref{abl:langevin_choices} and Ablation \ref{abl:pretrain_base}. Despite the occurrence of mode collapse, these strategies yield higher ELBOs, emphasizing the inherent exploration-exploitation trade-off discussed in \textbf{M2)}.

\textbf{M4)} Variational Monte Carlo methods heavily benefit from using a large number of steps $T$. This is shown in Ablation \ref{abl:num_steps}, where increasing the annealing steps for SIS methods and discretization steps for diffusion-based methods leads to tighter evidence bounds. However, increasing $T$ results in prolonged computational runtimes and demands substantial memory resources.

\textbf{M5)} GMMVI exhibits high sample efficiency (cf. Table \ref{tab:efficiency}). \citet{arenz2022unified} employ a replay buffer to enhance the sample efficiency of GMMVI, leading to orders of magnitude fewer target evaluations required for convergence. Consequently, GMMVI may be the preferable choice when target evaluations are time-consuming.

\textbf{M6)} 
Langevin diffusion methods demonstrate low sample efficiency, as highlighted in Table \ref{tab:efficiency}. These methods require evaluating the target at each intermediate discretization step due to the score function being part of the SDE, and they typically need many iterations to converge. Other diffusion-based methods that do not require target evaluations at every step, such as DDS, often perform poorly and suffer from mode collapse (cf. Ablation \ref{abl:grad_network}). To address this, \citet{zhang2021path} proposed incorporating the score function into the network architecture, resulting again in poor sample efficiency.


\section{Conclusion}
In this work, we assessed the latest sampling methods using a standardized set of tasks. Our exploration encompassed various performance criteria, with a specific focus on quantifying mode collapse. Through a comprehensive evaluation, we illuminated the strengths and weaknesses of state-of-the-art sampling methods, thereby offering a valuable reference for future developments.

\section*{Impact Statement}
This paper presents work whose goal is to advance the field of Machine Learning. There are many potential societal consequences of our work, none of which we feel must be specifically highlighted here.

\section*{Acknowledgments and Disclosure of Funding}
We thank Julius Berner, Lorenz Richter, and Vincent Stimper for many useful discussions.
D.B. acknowledges support by funding from the pilot program Core Informatics of the Helmholtz Association (HGF) and the state of Baden-Württemberg through bwHPC, as well as the HoreKa supercomputer funded by the Ministry of Science, Research and the Arts Baden-Württemberg and by the German Federal Ministry of Education and Research.

\bibliography{bibliography}
\bibliographystyle{icml2024}

\newpage
\appendix

\onecolumn
\section{Performance Criteria Details}
\label{appendix:eval_details}
Here, we provide further details on the computation of the various performance criteria introduced in the main manuscript.

\subsection{Density-Ratio-Based Criteria} \label{appendix:densityratio}


\textbf{Forward and Reverse Importance-Weighted Estimation of ${Z}$.}
Using the definition of the normalization constant, the importance-weighted reverse estimate of $Z$ is given by
\begin{equation}
\label{eq:rev_z}
    {Z}_{r} \coloneqq \int \gamma(\x) \dx = \int \frac{\model(\x)}{\model(\x)} \gamma(\x) = \E_{\model}\Big[\frac{\gamma(\x)}{\model(\x)}\Big] \approx \frac{1}{N_{\model}}\sum_{\x_i \sim \model} \frac{\gamma(\x_i)}{\model(\x_i)}
\end{equation}
where $N_{\model}$ denotes the number of samples from $\model$ used for the Monte Carlo estimate of the expectation. Using the identity $Z^{-1} = \target(\x) / \gamma(\x)$, we obtain the forward estimation of $Z$ as 
\begin{equation}
\label{eq:fwd_z}
    Z^{-1} = \int Z^{-1} \model(\x) \dx = \E_{\target}\Big[\frac{\model(\x)}{\gamma(\x)}\Big], \ \text{and thus, } \ Z_{f} \coloneqq  {1}/{\E_{\target}\Big[\frac{\model(\x)}{\gamma(\x)}\Big]} \approx 1/ \Big( \frac{1}{N_{\pi}}\sum_{\x_i \sim \target} \frac{\model(\x_i)}{\gamma(\x_i)} \Big), 
\end{equation}
where $N_{\target}$ denotes the number of samples from $\target$ used for the Monte Carlo estimate of the expectation.

\textbf{Forward and Reverse Effective Sample Size.}
The (reverse) effective sample size (ESS), or equivalently, reverse ESS \cite{shapiro2003monte} is defined as 
\begin{equation}
    \text{ESS}_{r} \coloneqq  {1}/{\E_{\model}\Big[\Big(\frac{\pi(\x)}{\model(\x)}\Big)^2\Big]} = {Z_r^2}/{\E_{\model}\Big[\Big(\frac{\gamma(\x)}{\model(\x)}\Big)^2\Big]} = {\Big(\E_{\model}\Big[\frac{\gamma(\x)}{\model(\x)}\Big]\Big)^2}/{\E_{\model}\Big[\Big(\frac{\gamma(\x)}{\model(\x)}\Big)^2\Big]},
\end{equation}
where $Z$ is approximated using the reverse estimate as defined in Eq. \ref{eq:rev_z}. 
Using the definition of the ESS, it is straightforward to see that
\begin{equation}
    \text{ESS}_{f} \coloneqq   {1}/{\E_{\model}\Big[\Big(\frac{\pi(\x)}{\model(\x)}\Big)^2\Big]} = {1}/{\E_{\target}\Big[\frac{\pi(\x)}{\model(\x)}\Big]} = {Z_f}/{\E_{\target}\Big[\frac{\gamma(\x)}{\model(\x)}\Big]} = {\E_{\target}\Big[\frac{\model(\x)}{\gamma(\x)}\Big]^{-1}}/ \ {\E_{\target}\Big[\frac{\gamma(\x)}{\model(\x)}\Big]},
\end{equation}
where $Z$ is approximated using the forward estimate as defined in Eq. \ref{eq:fwd_z}. 

\subsection{Integral Probability Metrics} \label{appendix:ipms}

\textbf{Maximum Mean Discrepancy.}
The Maximum Mean Discrepancy (MMD) \cite{gretton2012kernel} is a kernel-based measure of distance between two distributions. The MMD quantifies the dissimilarity between these distributions by comparing their mean embeddings in a
reproducing kernel Hilbert space (RKHS) \cite{aronszajn1950theory} with kernel $k$. In our setting, we are interested in computing the MMD between a model $\model$ and target distribution $\target$. Formally, if $\mathcal{H}_k$ is
the RKHS associated with kernel function $k$, the MMD between $\model$ and $\target$ is the integral probability metric defined by:
\begin{equation}
    \text{MMD}_k(\model,\target) = \sup_{f\in \mathcal{H}_k: \|f\|_{\mathcal{H}_k} \leq 1} \big( \mathbb{E}_{\x \sim \model}[f(\x)] - \mathbb{E}_{\y \sim \target}[f(\y)]
    \big),
\end{equation}
with $\text{MMD}_k(\model,\target) \geq 0$ and $\text{MMD}_k(\model,\target) = 0$ if and only if $\model = \target$.
The minimum variance unbiased estimate of $\text{MMD}_k$ between two sample sets $\X \sim \model$ and $\Y \sim \target$ with sizes $n$ and $m$ respectively is given by
\begin{equation}
    \text{MMD}_k(\model,\target) \approx \sqrt{\frac{1}{n(n-1)} \sum_{i,j}^{n} k(\x_i,\x_j) + \frac{1}{m(m-1)} \sum_{i,j}^{m} k(\y_i,\y_j) - \frac{2}{nm} \sum_i^n \sum_j^m k(\x_i,\y_j)},
\end{equation}
In our experiments, we took a squared exponential kernel given by $k(\x, \y) = \exp\big( - \|\x-\y\|_2^2/\alpha\big)$, where the bandwidth $\alpha$ is determined using the median heuristic \cite{gretton2012kernel}. 
The code for computing the MMD was built upon \url{https://github.com/antoninschrab/mmdfuse-paper}.

\textbf{Entropic Optimal Transport Distance.}
The 2-Wasserstein distance is given by
\begin{equation}
    W_2(\model, \target) = \inf \left\{ \ \int_{\rd \times \rd} c(\x,\y) \xi(\x,\y) \text{d}x\text{d}y : \ \int_{\rd} \xi (\x,\y) \text{d}\mathbf{y} = \model(\x), \ \int_{\rd} \xi (\x,\y) \dx = \target(\y) \right\}^{1/2},
\end{equation}
with cost $c$, chosen as $c(\x,\y)=\|\x -\y\|^2$ in our experiments.
To obtain a tractable objective, an entropy regularized version has been proposed \cite{peyre2019computational}, that is,  
\begin{equation}
    W_{2, \varepsilon}(\model, \target) = \inf \left\{ \ \int_{\rd \times \rd} c(\x,\y) \xi(\x,\y) \text{d}x\text{d}y - \varepsilon\mathcal{H}(\xi): \ \int_{\rd} \xi (\x,\y) \text{d}\mathbf{y} = \model(\x), \ \int_{\rd} \xi (\x,\y) \dx = \target(\y) \right\}^{1/2}.
\end{equation}
with entropy $\mathcal{H}(\xi) = -\int_{\rd \times \rd} \xi (\x,\y) \log \xi (\x,\y)\text{d}x\text{d}y$. We chose $\varepsilon = 10^{-3}$ for all experiments. The code was taken from \url{https://github.com/ott-jax/ott}.

\subsection{Extending the Entropic Mode Coverage} \label{appendix:ejs}
If the true mode probabilities $p^*(\M|\x)$ are not uniformly distributed, EMC=1 does not correspond to the optimal value. In that case, we propose the expected Jensen-Shannon divergence, that is,
\begin{equation}
    \text{EJS} \coloneqq\mathbb{E}_{q^{\theta}(\x)}D_{\text{JS}}(p(\M|\x)\|p^*(\M|\x)),
\end{equation}
with 
\begin{equation}
    D_{\text{JS}}(p(\M|\x)\|p^*(\M|\x)) = \frac{1}{2}\KL\left(p(\M|\x)\|\frac{p^*(\M|\x) + p(\M|\x)}{2}\right) + \frac{1}{2}\KL\left(p^*(\M|\x)\|\frac{p^*(\M|\x) + p(\M|\x)}{2}\right),
\end{equation}
as an alternative heuristic to quantify mode collapse. Similar to EMC, EJS is bounded and is straightforward to interpret: When employing the base 2 logarithm, EJS remains bounded, i.e., $0 \leq \text{EJS}. \leq 1$. Moreover $\text{EJS} = 0$ implies that the model matches the potentially unbalanced true mode probabilities, while $\text{EJS} = 1$ indicates that $p$ and $p^*$ possess no overlapping probability mass.

\section{Details on Unnormalized Importance Weights / Density Ratios} \label{appendix:is}
Here, we provide further details on how the unnormalized importance weights / density ratios are computed for different methods.

\textbf{Tractable Density Methods.}
For models with tractable density $\model(\x)$ the marginal (unnormalized) importance weights can trivially computed using 
\begin{equation*}
    w = \frac{\gamma(\x)}{\model(\x)}.
\end{equation*}

\textbf{Diffusion-based Methods.}
For diffusion-based methods, the extended importance weights can then be constructed as 
\begin{equation}
\label{eq:mc_rnd}
    \frac{p^\theta(\ex)}{\model(\ex)} = 
    \frac{ \pi(\x_{T}) \prod_{t=1}^{T} \ B^\theta_{t-1} (\x_{t-1}  | \x_{t})}{\target_0(\x_{0}) \prod_{t=0}^{T-1} F^\theta_{t+1} (\x_{t+1}  | \x_{t})}.
\end{equation}
The different choices of forward and backward transition kernels  $F^\theta_{t+1}, B^\theta_{t-1}$ are listed in Table \ref{tab:diffusion_methods}. Some methods such as DDS \cite{vargas2023denoising}, PIS \cite{zhang2021path} and GFN \cite{zhang2023diffusion} introduce a reference process $p^{\text{ref}}$ with 
\begin{equation}
    p^{\text{ref}}(\x_{0:T}) 
    = p^{\text{ref}}_0(\x_0)\prod_{t=0}^{T-1} F^{\text{ref}}_{t+1} (\x_{t+1}  | \x_{t})
    = p^{\text{ref}}_T(\x_T)\prod_{t=1}^{T} B^\theta_{t-1} (\x_{t-1}  | \x_{t}).
\end{equation}
This allows for rewriting Eq. \ref{eq:mc_rnd} as 
\begin{equation}
    \frac{p^\theta(\ex)}{\model(\ex)} = \frac{p^\theta(\ex)}{p^{\text{ref}}(\x_{0:T})} \cdot \frac{p^{\text{ref}}(\x_{0:T})}{\model(\ex)} = \frac{\pi(\x_T)}{p^{\text{ref}}_T(\x_T)} \cdot \frac{p^{\text{ref}}(\x_{0:T})}{\model(\ex)},
\end{equation}
potentially resulting in more tractable density ratios compared to Eq. \ref{eq:mc_rnd}. For concrete examples see e.g. \cite{zhang2023diffusion}. A continuous-time analogous of the reference process is detailed in \cite{vargas2023transport}. Moreover, in continuous-time, the importance weights correspond to a Radon–Nikodym derivative. For the sake of simplicity, we only consider the discrete-time setting in this work. We refer the reader to \cite{vargas2023transport, richter2023improved} for further details.

\begin{table}[t!]
\centering
\resizebox{0.98\textwidth}{!}{ 
\def\arraystretch{1.5}
\begin{tabular}{l|lll}
\toprule
Method  & $ \pi_0(\x_0)$ & $F^{\theta}_{t+1} (\x_{t+1}  | \x_{t} )$ & $B^{\theta}_{t-1} (\x_{t-1}  | \x_{t} )$ \\
\toprule
DDS
& $\mathcal{N}(\x_{0}| 0, \sigma_0^2\text{I})$
& $\mathcal{N}(\x_{t+1}| (\sqrt{1-\beta_t} \x_{t} + \score(\x_t, t))\Delta_t, \beta_t \sigma_0^2 \Delta_t)$
& $\mathcal{N}(\x_{t-1}| \sqrt{1-\beta_t} \x_{t}\Delta_t, \beta_t \sigma_0^2 \Delta_t)$
\\
DIS
& $\mathcal{N}(\x_{0}| 0, \sigma_0^2 \text{I})$
& $\mathcal{N}(\x_{t+1}|\x_{t} + (\beta_t \x_{t}+  \score(\x_t, t)) \Delta_t, 2 \beta_t \sigma_0^2 \Delta_t)$
& $\mathcal{N}(\x_{t-1}|  (\x_{t} - \beta_t \x_{t}) \Delta_t, 2 \beta_t \sigma_0^2 \Delta_t)$
\\
PIS/GFN
& $\delta_0$
& $\mathcal{N}(\x_{t+1}| \x_{t} + \score(\x_t, t)\Delta_t, \sigma_t^2 \Delta_t)$
& $\mathcal{N}(\x_{t-1}| \frac{t - \Delta_t}{t}\x_{t}, \frac{t - \Delta_t}{t}\sigma_t^2 \Delta_t)$
\\
ULA
& arbitrary$^*$ 
& $\mathcal{N}(\x_{t+1}|\x_t + \nabla_{\x_t}\sigma_t^2 \log \target_t(\x_t)\Delta_t, \sigma_t^2\Delta_t)$
& $\mathcal{N}(\x_{t-1}|\x_t + \nabla_{\x_t}\sigma_t^2 \log \target_t(\x_t)\Delta_t, \sigma_t^2\Delta_t)$
\\
{MCD}
& arbitrary$^*$ 
& $\mathcal{N}(\x_{t+1}|\x_t + \nabla_{\x_t}\sigma_t^2 \log \target_t(\x_t)\Delta_t, \sigma_t^2\Delta_t)$
& $\mathcal{N}(\x_{t-1}|\x_t + (\nabla_{\x_t}\sigma_t^2 \log \target_t(\x_t) + \score(\x_t, t))\Delta_t, \sigma_t^2\Delta_t)$
\\
{CMCD}
& arbitrary$^*$ 
& $\mathcal{N}(\x_{t+1}|\x_t + (\nabla_{\x_t}\sigma_t^2 \log \target_t(\x_t) + \score(\x_t, t))\Delta_t, \sigma_t^2\Delta_t)$
& $\mathcal{N}(\x_{t-1}|\x_t + (\nabla_{\x_t}\sigma_t^2 \log \target_t(\x_t) - \score(\x_t, t))\Delta_t, \sigma_t^2\Delta_t)$
\\
{GBS}
& arbitrary$^*$ 
& $\mathcal{N}(\x_{t+1}|\x_t + (\nabla_{\x_t}\sigma_t^2 \mathbf{f}^{\theta}(\x_t, t))\Delta_t, \sigma_t^2\Delta_t)$
& $\mathcal{N}(\x_{t-1}|\x_t + (\nabla_{\x_t}\sigma_t^2 \mathbf{b}^{\theta}(\x_t, t))\Delta_t, \sigma_t^2\Delta_t)$
\\
\bottomrule
    \end{tabular}
    }
    \vskip -0.1in
    \caption{Characterization of diffusion-based sampling methods. Here, $\score, \mathbf{f}^{\theta}, \mathbf{b}^{\theta}: \rd  \times [0, T]\rightarrow \rd$ denotes a parameterized function approximator. $^*$ In our experiments, we choose $\pi_0(\x_0) = \mathcal{N}(\x_{0}| 0, \sigma_0^2 \Delta_t)$. 
    }
    \label{tab:diffusion_methods}
\end{table}

\textbf{Sequential Importance Sampling Methods.}
Sequential importance sampling methods express the importance weights in terms of incremental importance sampling weights, i.e.,
\begin{equation*}
   \olsi w =\prod_{t=1}^T G_{t}(x_{t-1}, x_t) \qquad \text{with} \qquad G_{t}(\x_{t-1}, \x_t) = \frac{\gamma_t(\x_t)B^{\theta}_{t-1} (\x_{t-1}  | \x_{{t}} )}{\gamma_{t-1}(\x_{t-1})F^{\theta}_{t} (\x_{{t}}  | \x_{{t-1}} )}.
\end{equation*}
 For given forward transitions $F_t^{\theta}$, the optimal backward transitions $B^{\theta}_{t-1}$ ensure that $\olsi w = w$. As the optimal transitions are typically not available, SMC uses the AIS approximation \cite{neal2001annealed}. Moreover flow transport methods \cite{wu2020stochastic, arbel2021annealed, matthews2022continual} use a flow as a deterministic map $\flow$ to approximate the incremental IS weights. In Table \ref{tab:sis}, we list different $F^{\theta}_{t}, B^{\theta}_{t-1}$ and their corresponding incremental importance sampling weights.

\begin{table}[t!]
\centering
\resizebox{.98\textwidth}{!}{ 
\def\arraystretch{1.2}
\begin{tabular}{l|ll|ll}
\toprule
Method  & $B^{\theta}_{t-1} (\x_{t-1}  | \x_{{t}} )$ & $G_{t}(\x_{t-1}, \x_t)$ 
& $F^{\theta}_{t} (\x_{{t}}  | \x_{{t-1}} )$ & $G_{t}(\x_{t-1}, \x_t)$ \\
\toprule
Optimal
& ${\target_{t-1}(\x_{t-1})}{F^{\theta}_{t}(\x_t|\x_{t-1})}/\target_{t}(\x_{t})$
& $Z_t/Z_{t-1}$
& $\target_{t}(\x_{t}){B^{\theta}_{t-1} (\x_{t-1}  | \x_{{t}} )}/{\target_{t-1}(\x_{t-1})}$
& $Z_t/Z_{t-1}$
\\
AIS/SMC/FAB
& ${\target_{t}(\x_{t-1})}{F^{\theta}_{t}(\x_t|\x_{t-1})}/\target_{t}(\x_{t})$
& $\gamma_t(\x_{t-1})/\gamma_{t-1}(\x_{t-1})$
& $\target_{t-1}(\x_{t}){B^{\theta}_{t-1} (\x_{t-1}  | \x_{{t}} )}/{\target_{t-1}(\x_{t-1})}$
& $\gamma_t(\x_{t})/\gamma_{t-1}(\x_{t})$
\\
AFT/CRAFT
& $\delta_{\flow_{t}(x_{t})}(x_{t-1})$
& $ {\gamma_t(\flow_t(\x_{t-1}))}|\det \nabla \flow_t(\x_{t-1})|/\gamma_{t-1}(\x_{t-1})$
& $\delta_{(\flow_{t-1})^{-1}(x_{t-1})}(x_{t})$
& $ {\gamma_t(\x_{t})}/\gamma_{t-1}((\flow_{t-1})^{-1}(\x_{t}))|\det \nabla (\flow_{t-1})^{-1}(\x_{t})|$
\\
\bottomrule
    \end{tabular}
    }
    \vskip -0.1in
    \caption{Characterization of Sequential Importance Methods methods: The middle column shows the backward kernels  $B^{\theta}_{t-1}$ and the corresponding $G_{t}$ when transporting samples from the prior $\pi_0$ to 
    the target $\pi_T$ to compute reverse criteria. The right-most column shows the forward kernels $F^{\theta}_{t}$ and the corresponding $G_{t}$ when transporting samples from the target $\pi_T$ back to the prior $\pi_T$ to compute forward criteria. 
    }
    \label{tab:sis}
\end{table}

\section{Benchmark Target Details} \label{appendix:target_details}
Here, we introduce the target densities considered in this benchmark more formally.
\subsection{Bayesian Logistic Regression}
We used four binary classification problems in our benchmark, which have also been used in various other work to compare different state-of-the-art methods in variational inference and Markov chain Monte Carlo. 
We assess the performance of a Bayesian logistic model with:
$$
\begin{aligned}
& \x \sim \mathcal{N}\left(0, \sigma^2_w I\right), \\
& y_i \sim \operatorname{Bernoulli}(\operatorname{sigmoid}(\x^\top u_i))
\end{aligned}
$$
on two standardized datasets $\{(u_i, y_i)\}_i$, namely \textbf{Ionosphere} ($d=35$) with 351 data points and \textbf{Sonar} ($d=61$) with 208 data points.

The \textbf{German Credit} dataset consists of ($d=25$) features and 1000 data points, while the \textbf{Breast Cancer} dataset has ($d=31$) dimensions with 569 data points, which we standardize and apply linear logistic regression.

\subsection{Random Effect Regression}
The \textbf{Seeds} ($d=26$) target is a random effect regression model trained on the \textit{seeds} dataset:
$$
\begin{aligned}
& \tau \sim \operatorname{Gamma}(0.01,0.01) \\
& a_0, a_1, a_2, a_{12} \sim \mathcal{N}(0,10) \\
& b_i \sim \mathcal{N}\left(0, \frac{1}{\sqrt{\tau}}\right), \quad{i=1, \ldots, 21} \\
& \operatorname{logits}_i=a_0+a_1 x_i+a_2 y_i+a_{12} x_i y_i+b_1, \quad{i=1, \ldots, 21} \\
& r_i \sim \operatorname{Binomial}\left(\operatorname{logits}_i, N_i\right), \quad{i=1, \ldots, 21}. \\
&
\end{aligned}
$$
The goal is to do inference over the variables $\tau, a_0, a_1, a_2, a_{12}$ and $b_i$ for $i=1, \ldots, 21$, given observed values for $x_i, y_i$ and $N_i$.

\subsection{Time Series Models}
The \textbf{Brownian} ($d=32$) model corresponds to the time discretization of a Brownian motion:
\begin{align*}
    \alpha_{\mathrm{inn}} &\sim \mathrm{LogNormal}(0,2),\\
    \alpha_{\mathrm{obs}} &\sim \mathrm{LogNormal}(0,2) ,\\
    x_1 &\sim \gN(0, \alpha_{\mathrm{inn}}),\\
    x_i &\sim \gN(x_{i-1}, \alpha_{\mathrm{inn}}), \quad i=2,\hdots 20,\\
    y_i &\sim \gN(x_{i}, \alpha_{\mathrm{obs}}), \quad i=1,\hdots 30.
\end{align*}
inference is performed over the variables $\alpha_{\mathrm{inn}}, \alpha_{\mathrm{obs}}$ and $\{x_i\}_{i=1}^{30}$ given the observations $\{y_i\}_{i=1}^{10} \cup \{y_i\}_{i=20}^{30}$.

\subsection{Spatial Statistics}
The \textbf{Log Gaussian Cox process} ($d=1600$) is a popular high-dimensional task in spatial statistics \citep{moller1998log} which models the position of pine saplings. Using a $d = M \times M = 1600$ grid, we obtain the unnormalized target density by
\begin{equation*}
    \mathcal{N}(\x ; \mu, K) \prod_{i \in[1: M]^2} \exp \left(x_i y_i-a \exp \left(x_i\right)\right).
\end{equation*}
\subsection{Synthetic Targets}

We evaluate on three different mixture models which all follow the structure, that is,
$$
\begin{aligned}
 \target(\x) = &\sum_{k=1}^K w_k \target_k(\x), \\
 &\sum_{k=1}^K w_k = 1, \\
\end{aligned}
$$
where $K$ denotes the number of components. 

The \textbf{MoG} ($d=\mathbb{N}$) distribution, taken from \citep{midgley2022flow}, consists of $K =40$ mixture components with
$$
\begin{aligned}
 &\target_k(\x) = \mathcal{N}(\mu_k, I) \\
 & \mu_k \sim \mathcal{U}(-40, 40) \\
 & w_k = 1 / K, \\
\end{aligned}
$$
where $\mathcal{U}(l,u)$ refers to a uniform distribution on $[l,u]$.

The \textbf{MoS} ($d=\mathbb{N}$) comprises 10 Student's t-distributions $t_2$, where the $2$ refers to the degree of freedom. Generally, Student's t-distributions have heavier tails compared to Gaussian distributions, making them sharper and more challenging to approximate. 
$$
\begin{aligned}
 &\target_k(\x) = t_2 + \mu_k,\\
 & \mu_k \sim \mathcal{U}(-10, 10), \\
 & {w}_k =1 / K, \\
\end{aligned}
$$
where $\mu_k$ refers to the translation of the individual components.

The \textbf{Funnel} ($d=10$) target introduced in \citep{neal2003slice} is a challenging funnel-shaped distribution given by
\begin{equation*}
     \pi(\x) = \mathcal{N}(x_1 ; 0, \sigma_f^2) \mathcal{N}(x_{2: 10} ; 0, \exp 
 (x_1) I),
 \end{equation*}
 with $\sigma_f^2 = 9$. 


Lastly, we follow \citet{doucet2022score} and use NICE \cite{dinh2014nice} to train a normalizing flow on a $14 \times 14$ and $28 \times 28$ variant of MNIST (\textbf{DIGITS}) and on the $28 \times 28$ Fashion MNIST dataset (\textbf{Fashion}). 

\section{Algorithms and Parameter Choices}
\label{appendix:algos_params}
Here, we discuss the parameter choices of all methods. Most of these choices are based on recommendations of the authors. For some choices, we run ablation studies to find suitable values. 

\textbf{Gaussian Mean Field Variational Inference (MFVI).} We updated the mean and the diagonal covariance matrix using the Adam optimizer \cite{kingma2014adam} for $100k$ iterations with a batch size of $2000$. We ensured non-negativeness of the variance by using a $\log$ transformation. The mean is initialized at ${0}$ for all experiments. The initial covariance/scale and the learning rate are set according to Table \ref{tab:hparams_scale}. 

\textbf{Gaussian Mixture Model Variational Inference (GMMVI).} For GMMVI, we ported the tensorflow implementation of \url{https://github.com/OlegArenz/gmmvi} to Jax and integrated it into our framework. We use the specifications \cite{arenz2022unified} described as SAMTRUX. We make use of their adaptive component initializer and start using ten components. The initial variance of the components is set according to Table \ref{tab:hparams_scale}. 


\textbf{Sequential Monte Carlo (SMC).} For the Sequential Monte Carlo (SMC) approach, we leveraged the codebase available at \url{https://github.com/google-deepmind/annealed_flow_transport}. We used $2000$ particles and $128$ annealing steps (temperatures) $T$. We used resampling with a threshold of $0.3$. We used one Hamiltonian Monte Carlo (HMC) step per temperature with $10$ leapfrog steps. We tuned the stepsize of HMC according to Table \ref{tab:hparams_scale} where we used different stepsizes depending on the annealing parameter $\beta_t$. We additionally tune the scale of the initial proposal distribution $\pi_0$ as shown in Table \ref{tab:hparams_scale}.

\textbf{Continual Repeated Annealed Flow Transport (CRAFT/AFT).}
As AFT and CRAFT build on Sequential Monte Carlo (SMC), we employed the same SMC specifications detailed above. Notably, we found that employing simpler flows in conjunction with a greater number of temperatures yielded superior and more robust performance compared to the use of more sophisticated flows such as RealNVP or Neural Spline Flows. Consequently, we opted for 128 temperatures, utilizing diagonal affine flows as the transport map. Specifically for AFT, we determined that 400 iterations per temperature were sufficient to achieve converged training results. For CRAFT and SNF, we found that a total of 3000 iterations provided satisfactory convergence during training. For all methods, we use $2000$ particles for training and testing and tune the learning rate and the scale of the initial proposal distribution $\pi_0$ as shown in Table \ref{tab:hparams_scale}.

\textbf{Flow Annealed Importance Sampling Bootstrap (FAB).}
We built our implementation off of \url{https://github.com/lollcat/fab-jax}. We adjusted the parameters of FAB in accordance with the author's most important hyperparameter suggestions and to ensure that SMC performs reasonably well. To achieve this, we set the number of temperatures to $128$ and used HMC as MCMC kernel. For the flow architecture we use RealNVP \cite{dinh2016density} where the conditioner is given by an 8-layer MLP. 
Furthermore, we utilized FAB's replay buffer to speed up computations. The learning rate and base distribution scale are adjusted for target specificity, following the specifications outlined in Table \ref{tab:hparams_scale}.
We used a batch size of 2048 and trained FAB for 3000 iterations which proved sufficient for achieving a satisfactory convergence. 

\textbf{Denoising Diffusion Sampler (DDS) and Path Integral Sampler (PIS).} We use the implementation of \url{https://github.com/franciscovargas/denoising_diffusion_samplers} to integrate the Diffusion and Path Integral Sampler into our framework. We set the parameters of the SDEs according to the authors, i.e., \cite{zhang2021path} and \cite{vargas2023denoising} and use 128 timesteps and a batch size of 2000 if not otherwise stated. Both methods use the network proposed in \cite{zhang2021path} which uses a sinusoidal position embedding for the timestep and uses the gradient of the log target density as an additional term.  As proposed, we use a two-layer neural network with $64$ hidden units. For DDS we use a cosine scheduler \cite{vargas2023denoising} and for PIS a uniform time scheduler \cite{zhang2021path}. Both methods were trained using $40k$ iterations.

\textbf{Monte Carlo Diffusions (MCD) and Langevin Diffusion Variational Inference (LDVI).}
We build our implementation of Langevin Diffusion methods on \url{https://github.com/tomsons22/LDVI}. For experiments where performance is solely measured in terms of ELBO, due to the lack of samples from $\pi$ or access to $Z$, we train all parameters of the SDE by maximizing the EUBO as suggested by \cite{Geffner:2021}. For multimodal target densities, we fix the proposal distribution and the magnitude of the timestep. We found that this stabilizes training and yields better results (cf. Ablation \ref{abl_langevin}). We use the network architecture proposed by \cite{zhang2021path} with two hidden layers with 64 hidden units each. We discretize the SDEs using 128 timesteps and a batchsize of 2000 if not otherwise stated. All methods were trained using $40k$ iterations.

\textbf{Time-Reversed Diffusion Sampler (DIS) and General Bridge Samples (GBS).} We base the implementation of  DIS and GBS on \url{https://github.com/juliusberner/sde_sampler} and implemented them in Jax. The remaining parameters follow the description of DDS and PIS above. 

\begin{table}[hbtp]
\centering

\resizebox{\textwidth}{!}{  

\begin{tabular}{l|c|ccccccccccc}
	\toprule 
         \textbf{Methods} / Parameters   & Grid 
        & MoG & MoS & Funnel & Digits/Fashion & Credit & Cancer  & Brownian  & Sonar  & Seeds  & Ionosphere & LGCP\\
        \midrule
        \textbf{MFVI}  & \\
        Initial Scale   & \{0.1, 1, 10\} & - & - & 1 & 0.1  & 0.1 & 10 & 1 & 0.1 & 1 & 1 & 0.1\\
        Learning Rate   & $\{ 10^{-2},10^{-3},10^{-4},10^{-5} \}$ & $10^{-2}$ & $10^{-3}$ & $10^{-2}$ & $10^{-4}$ & $10^{-3}$ & $10^{-2}$ & $10^{-3}$ & $10^{-3}$ & $10^{-4}$ & $10^{-4}$ & $10^{-3}$\\
        \midrule
        \textbf{GMMVI}  & \\
        Initial Scale  & \{0.1, 1, 10\} & - & - & 0.1 & 10 & 0.1 & 1 & 0.1 & 1 & 0.1 & 10 & NA\\
        \midrule
        \textbf{SMC}  & \\
        Initial Scale   & \{0.1, 1, 10\} & - & - & 1 & 1 & 0.1 & 1 & 1 & 1 & 1 & 1 & 1 \\
        HMC stepsize ($\beta \leq 0.5$) & \{0.001, 0.01, 0.05, 0.1, 0.2\}  & 0.2 & 0.2 & 0.001 & 0.2 & 0.1 & 0.05 & 0.001 & 0.05 & 0.2 & 0.2 & 0.01 \\
        HMC stepsize ($\beta > 0.5$) & \{0.001, 0.01, 0.05, 0.1, 0.2\}  & 0.001 & 0.2 & 0.1 & 0.2 & 0.1 & 0.01 & 0.05 & 0.001 & 0.05 & 0.2 & 0.2\\
        \midrule
        \textbf{AFT}  & \\
        Initial Scale   & \{0.1, 1, 10\} & - & - & 1 & - & 0.1 & 0.1 & NA & 1 & 1 & 1 & 1\\
        Learning Rate   & $\{ 10^{-3},10^{-4},10^{-5} \}$ & $10^{-3}$ & $10^{-4}$ & $10^{-3}$ & $10^{-3}$ & $10^{-3}$ & $10^{-3}$ & NA & $10^{-3}$ & $10^{-4}$ & $10^{-3}$ & $10^{-3}$\\
        \midrule
        \textbf{CRAFT}  & \\
        Initial Scale   & \{0.1, 1, 10\} & - & - & 1 & - & 0.1 & 1 & 1 & 1 &  0.1 & 0.1 & 1\\
        Learning Rate   & $\{ 10^{-3},10^{-4},10^{-5} \}$ & $10^{-3}$ & $10^{-4}$ & $10^{-3}$ & $10^{-4}$ & $10^{-4}$ & $10^{-3}$ & $10^{-3}$ & $10^{-3}$ & $10^{-3}$ & $10^{-3}$ & $10^{-3}$\\
        \midrule
        \textbf{FAB}  & \\
        Initial Scale   & \{0.1, 1, 10\} & - & - & 1 & - & 0.1 & 0.1 & 1 & 0.1 & 0.1 & 1 & 0.1\\
        Learning Rate   & $\{ 10^{-3},10^{-4},10^{-5} \}$ & $10^{-5}$ & - & $10^{-4}$ & $10^{-3}$ & $10^{-4}$ & $10^{-3}$ & $10^{-3}$ & $10^{-3}$ & $10^{-4}$ & $10^{-3}$ & $10^{-3}$\\ 
        \midrule
        \textbf{DDS}/\textbf{DIS}  & \\
        Initial Scale   & \{0.1, 1, 10\} & - & - & 1 & - & 1 & 1 & 0.1 & 1 & 0.1 & 1 & 0.1\\
        Learning Rate   & $\{ 10^{-3},10^{-4},10^{-5} \}$ & $10^{-4}$ & $10^{-4}$ & $10^{-3}$ & $10^{-3}$ & $10^{-3}$ & $10^{-4}$ & $10^{-4}$ & $10^{-3}$ & $10^{-4}$ & $10^{-4}$ & $10^{-4}$\\ 
        \midrule
        \textbf{PIS}  & \\
        Learning Rate   & $\{ 10^{-3},10^{-4},10^{-5} \}$ & $10^{-3}$ & $10^{-3}$ & $10^{-3}$ & $10^{-3}$ & $10^{-4}$ & $10^{-4}$ & $10^{-4}$ & $10^{-3}$ & $10^{-3}$ & $10^{-4}$ & NA\\        
        \midrule
        \textbf{LDVI}  & \\
        Initial Scale   & \{0.1, 1, 10\} & - & - & 1 & - & 0.1 & 0.1 & 1 & 1 & 0.1 & 0.1 & 0.1\\
        Learning Rate & $\{ 10^{-3},10^{-4},10^{-5} \}$ & $10^{-3}$ & $10^{-3}$ & $10^{-4}$ & $10^{-3}$ & $10^{-3}$ & $10^{-3}$ & $10^{-3}$ & $10^{-3}$ & $10^{-3}$ & $10^{-3}$ & $10^{-3}$\\
        \midrule
        \textbf{MCD}  & \\
        Initial Scale   & \{0.1, 1, 10\} & - & - & 1 & - & 0.1 & 0.1 & 1 & 0.1 & 0.1 & 1 & 1\\
        Learning Rate & $\{ 10^{-3},10^{-4},10^{-5} \}$ & $10^{-3}$ & $10^{-3}$ & $10^{-3}$ & $10^{-3}$ & $10^{-4}$ & $10^{-3}$ & $10^{-3}$ & $10^{-3}$ & $10^{-3}$ & $10^{-3}$ & $10^{-3}$\\
	\bottomrule
\end{tabular}
}
 \caption{Hyperparameter selection for all different sampling algorithms. The `Grid' column indicates the values over which we performed a grid search. The values in the column which are marked with experiment names indicate which values were chosen for the reported results.}\label{tab:hparams_scale}
\vskip -0.15in
\end{table}

\section{Further Experimental results} \label{appendix:exp2}
\begin{figure*}
\centering
\setlength{\tabcolsep}{1pt}
\begin{tabular}{ccccc}
\toprule
  \includegraphics[width=0.18\textwidth]{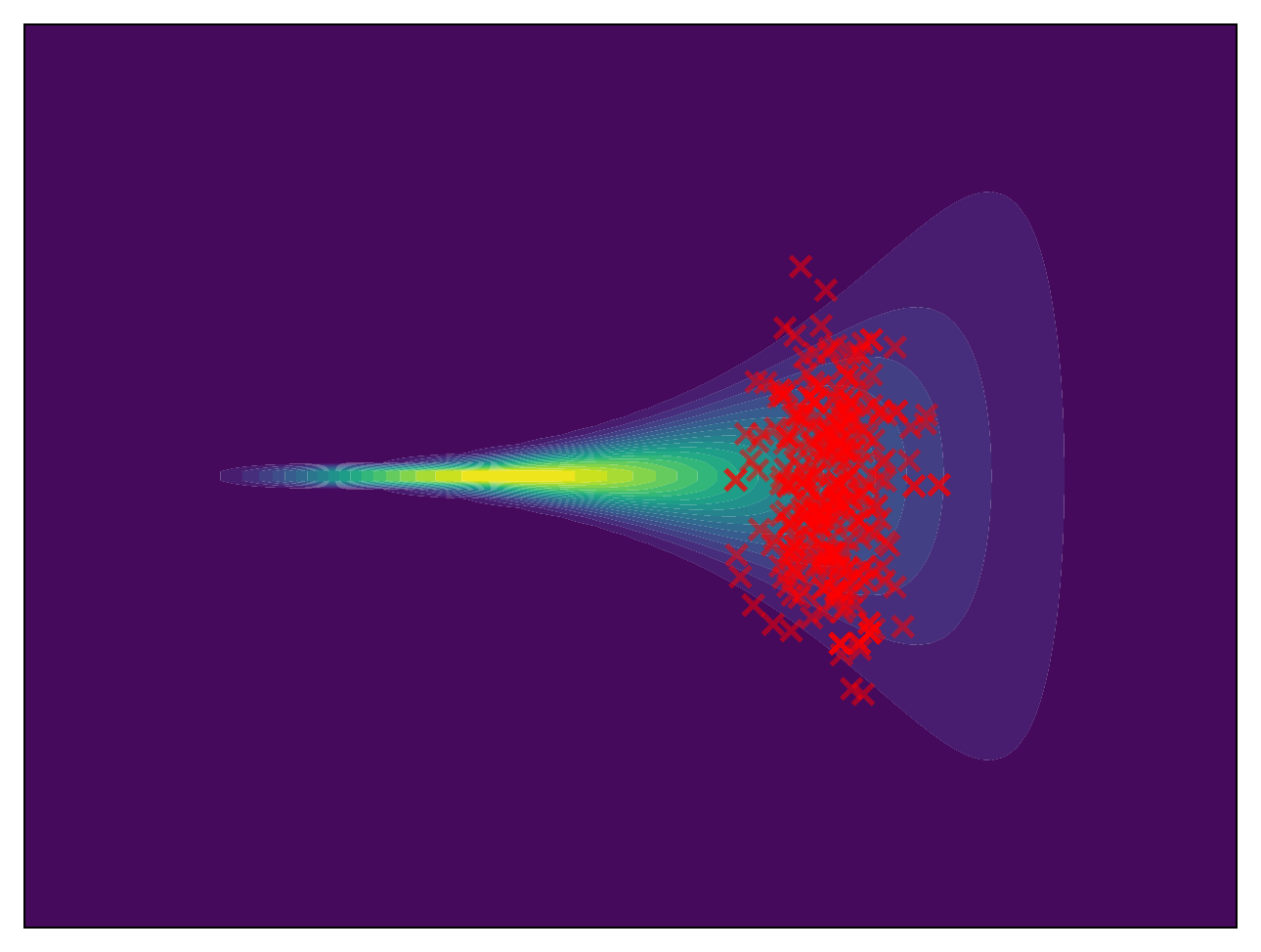} &   
  \includegraphics[width=0.18\textwidth]{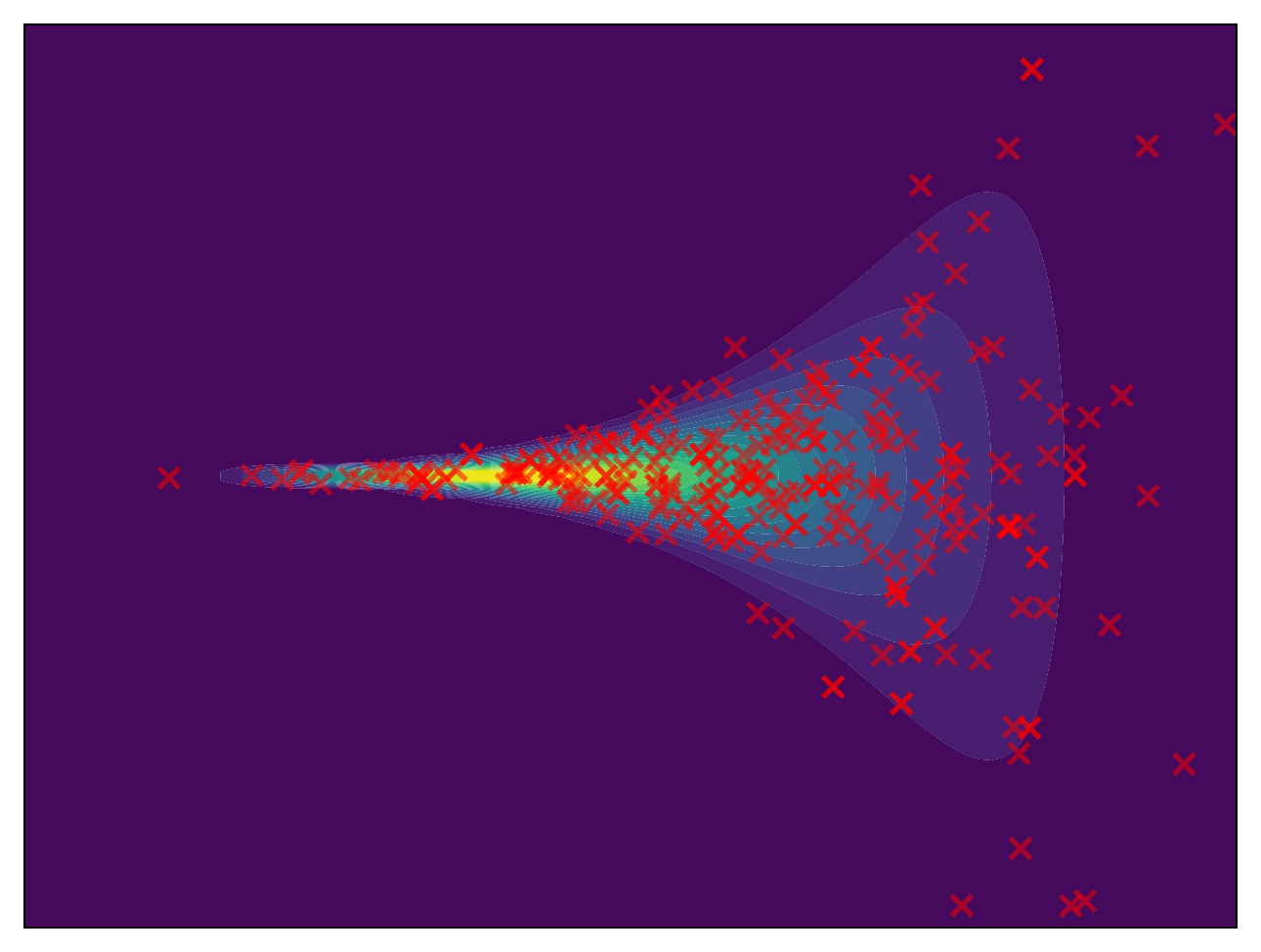} &   
  \includegraphics[width=0.18\textwidth]{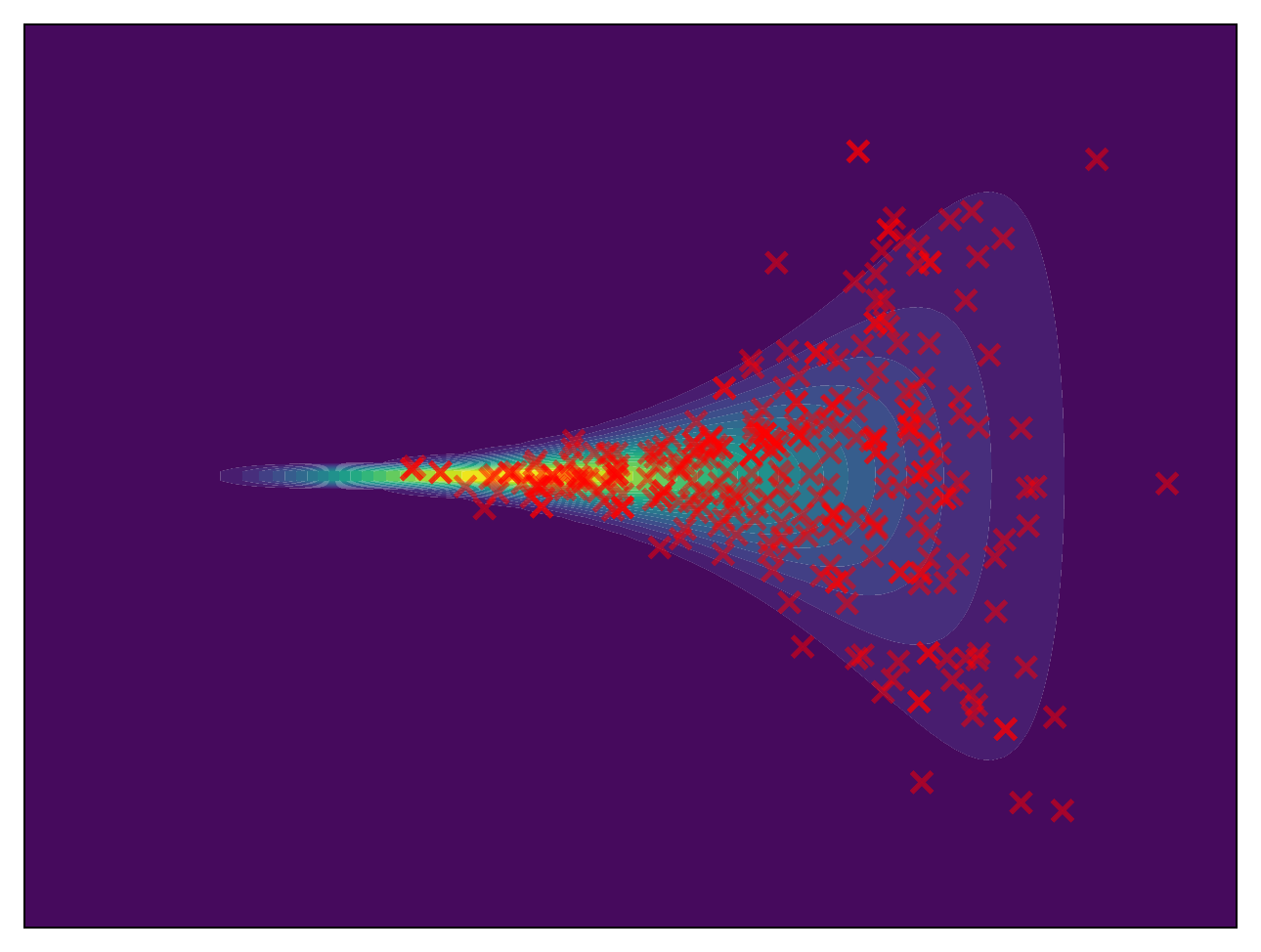} &   
  \includegraphics[width=0.18\textwidth]{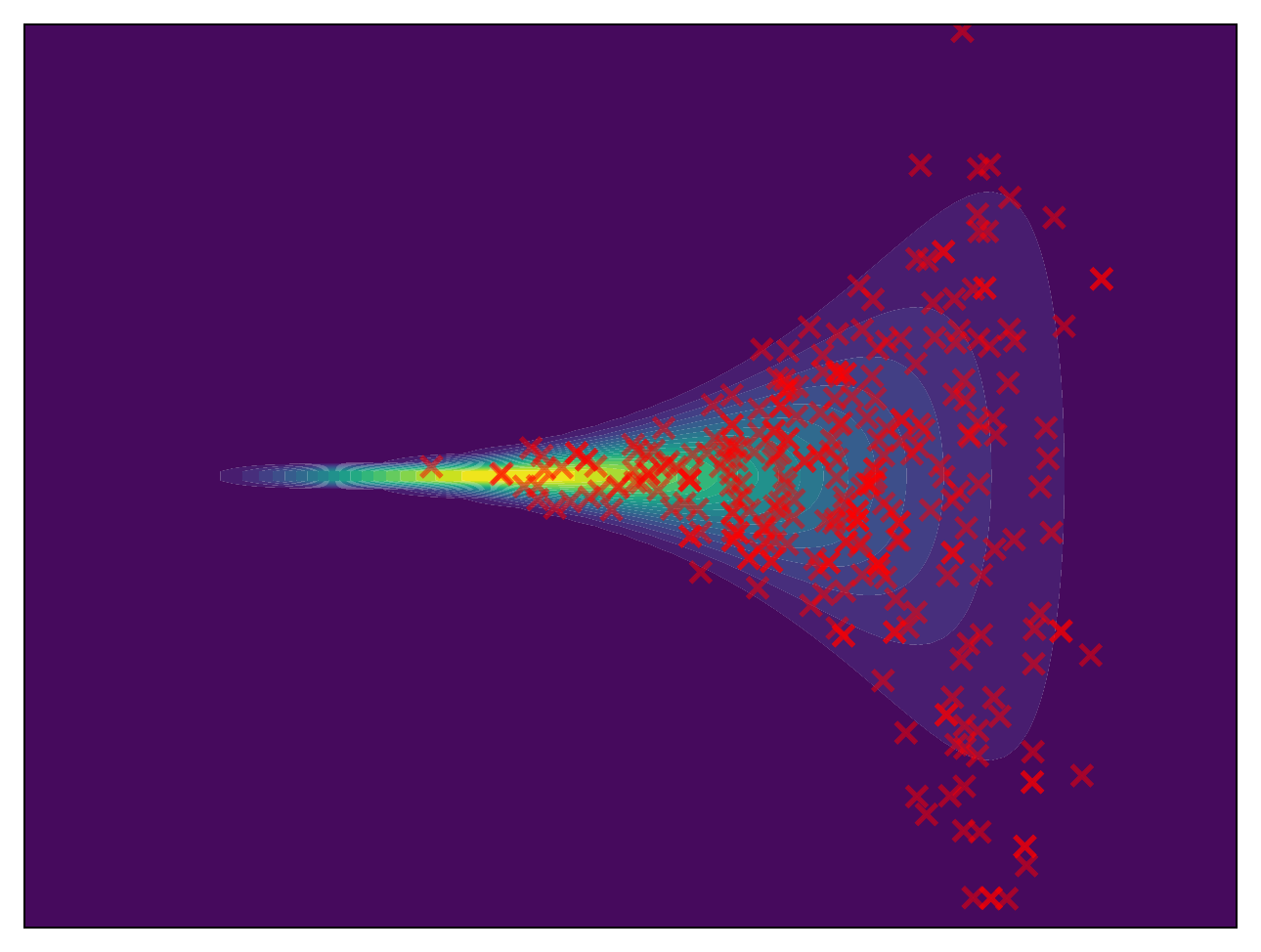} &   
  \\
  \small MFVI &
  \small GMMVI &
  \small SMC &
  \small AFT &
  \\
  \includegraphics[width=0.18\textwidth]{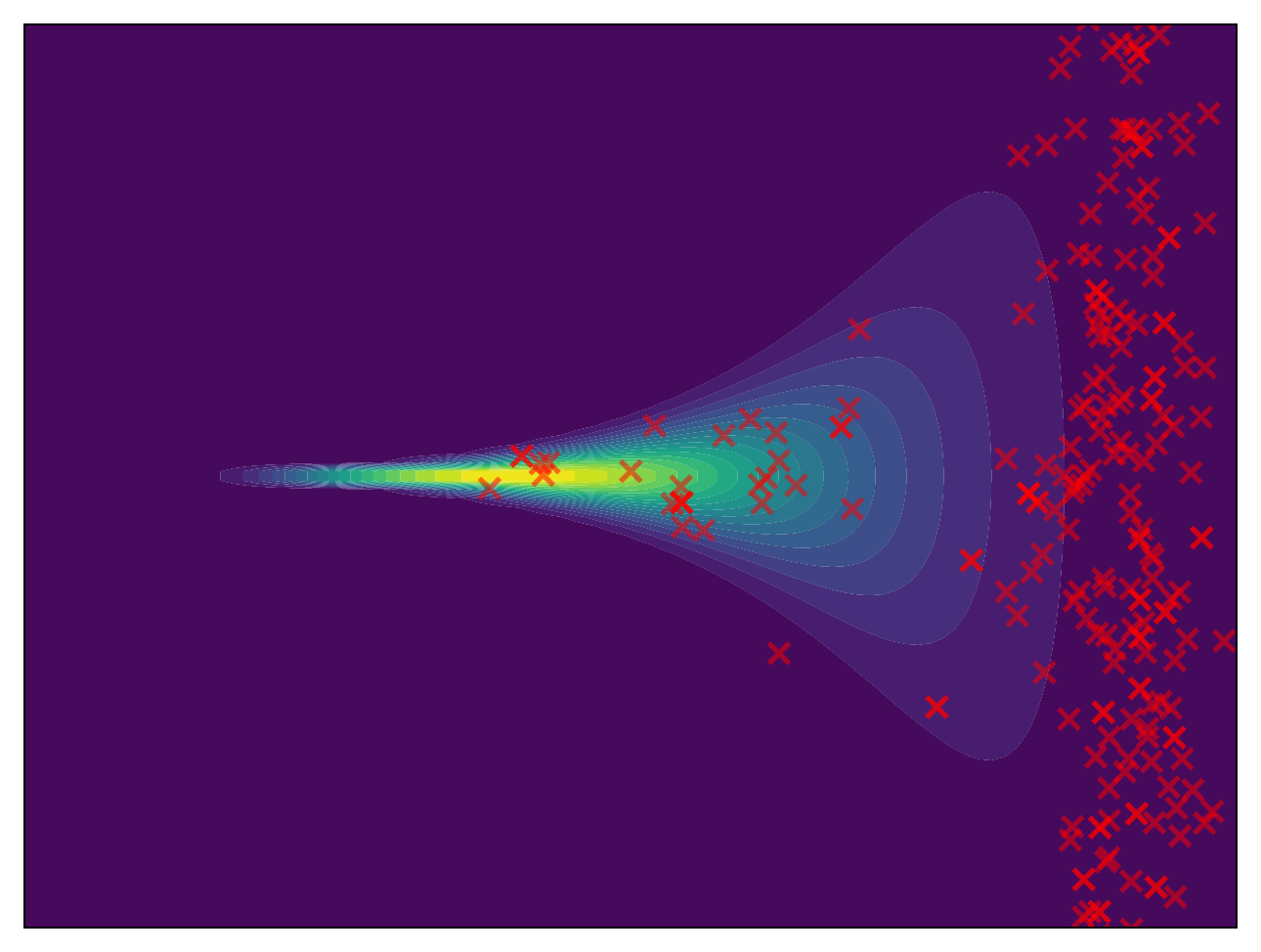} &   
  \includegraphics[width=0.18\textwidth]{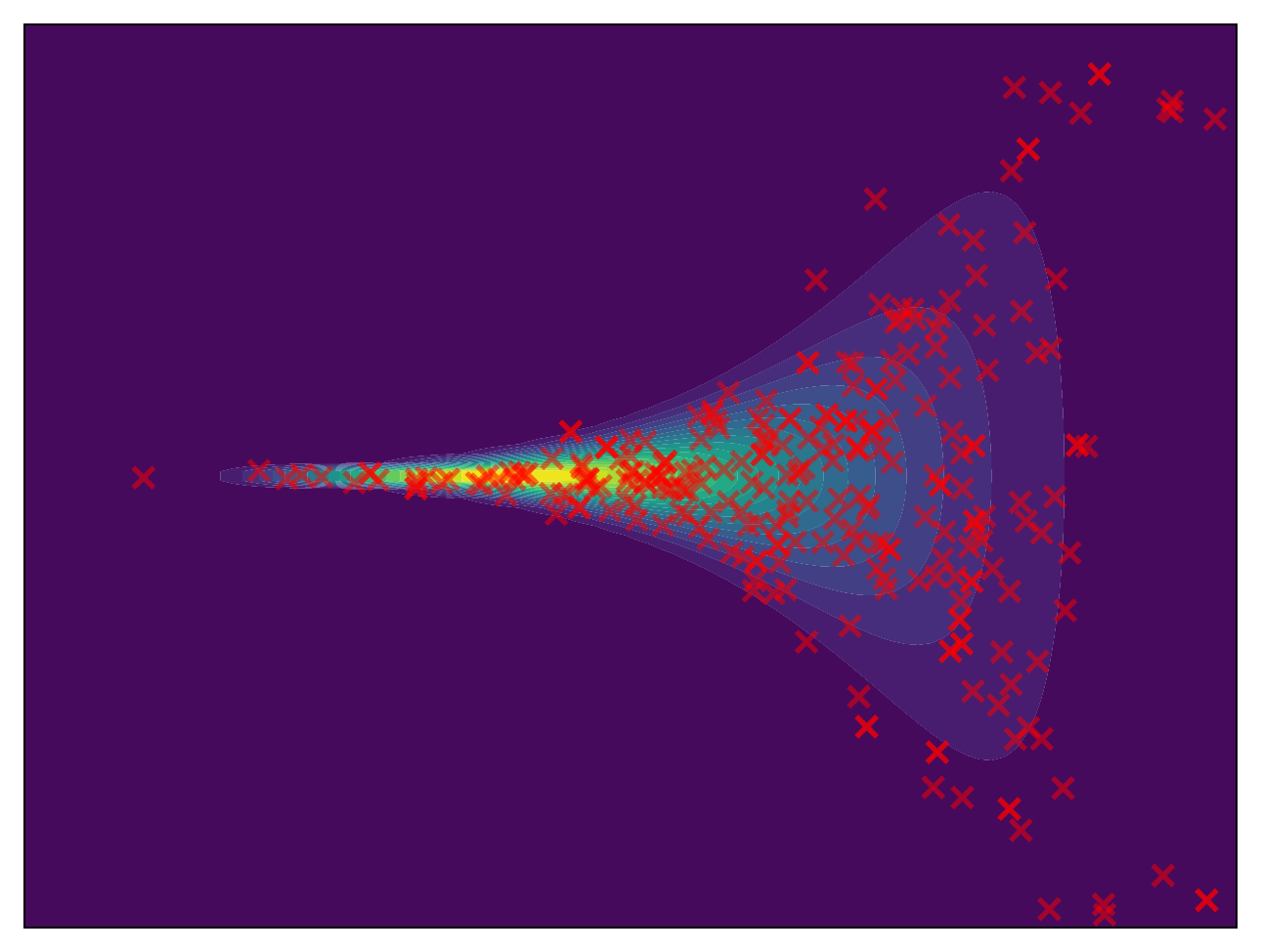} &   
  \includegraphics[width=0.18\textwidth]{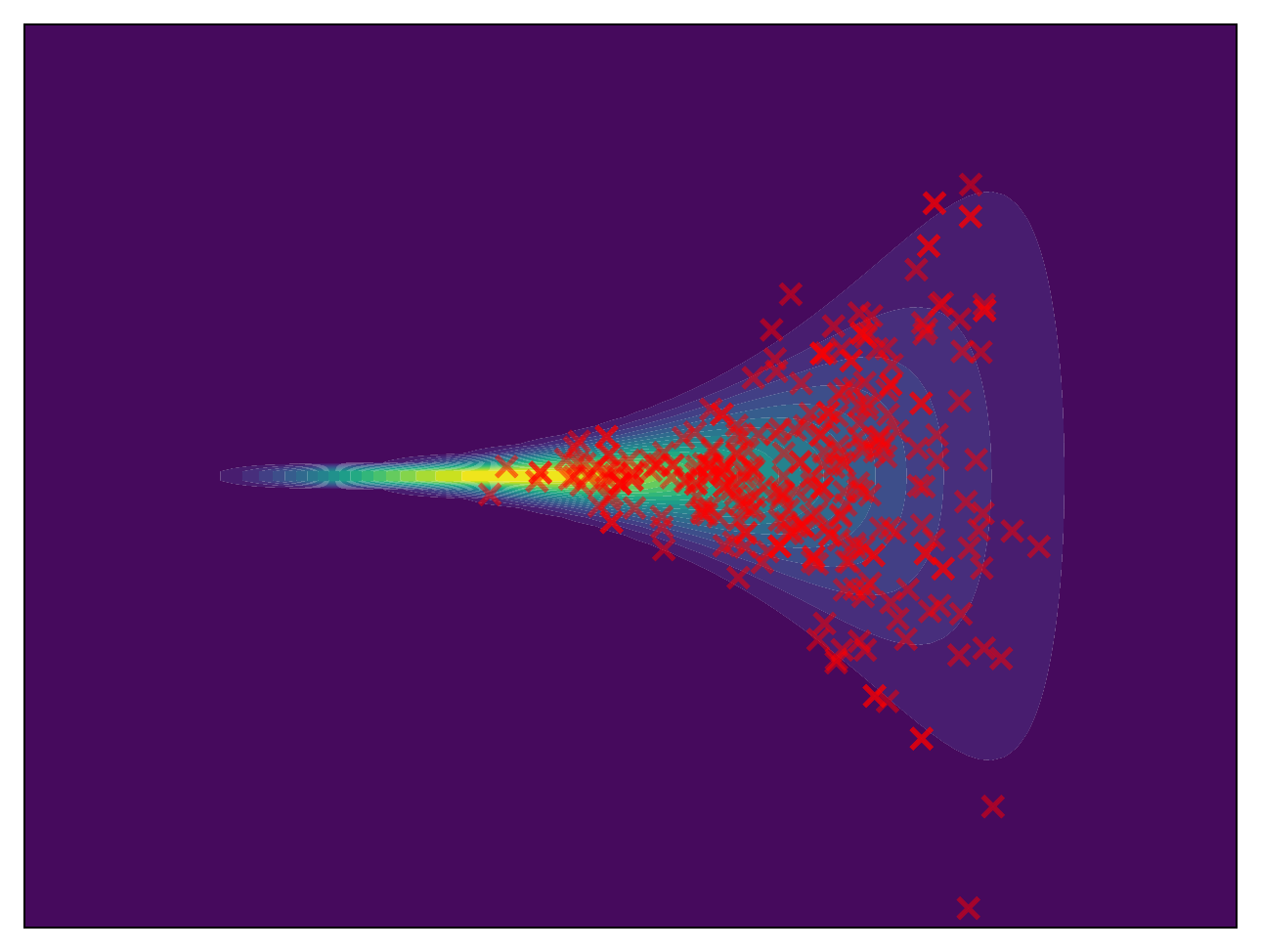} &   
  \includegraphics[width=0.18\textwidth]{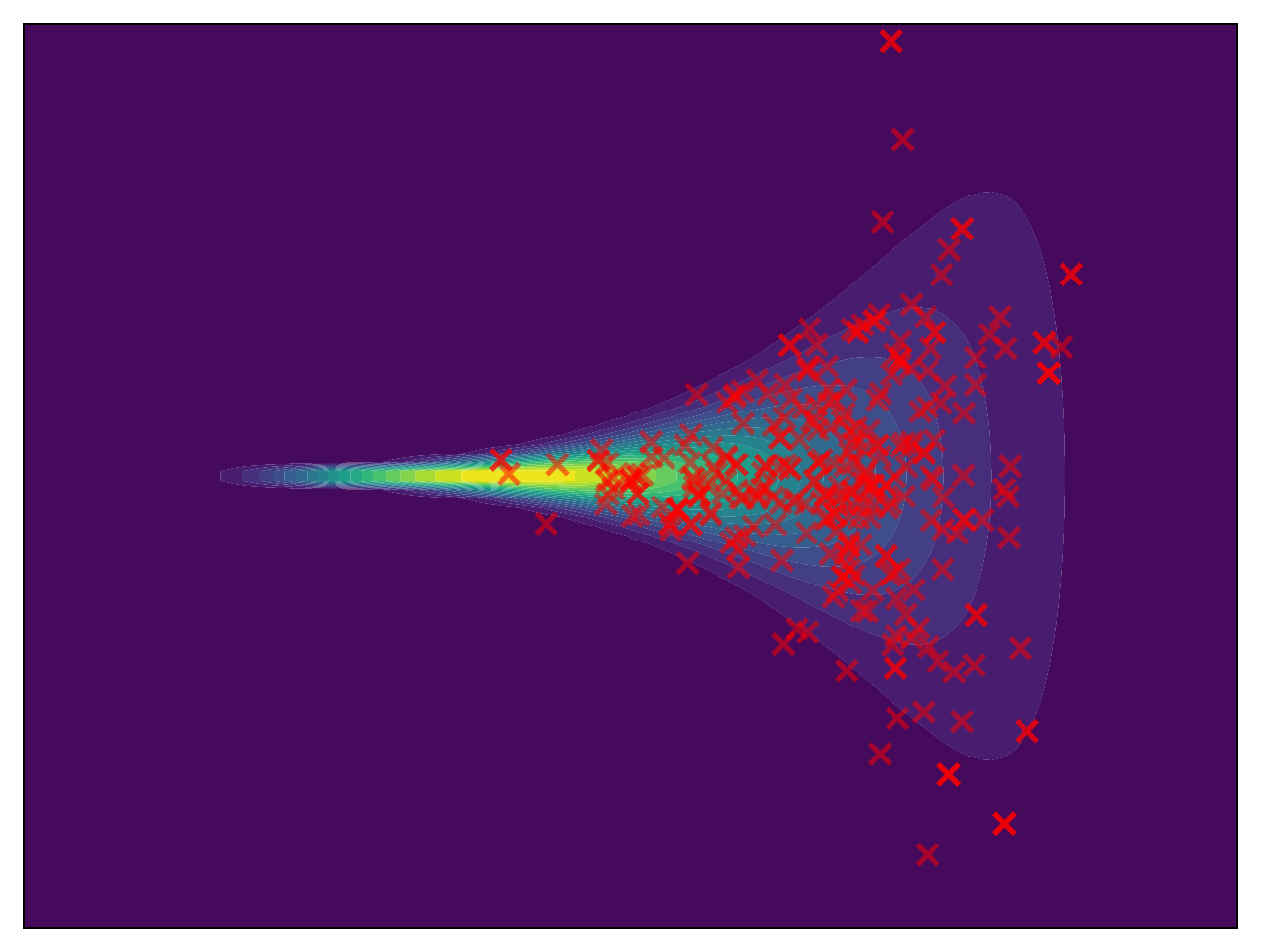} &     
  \\
  \small CRAFT &
  \small FAB &
  \small MCD &
  \small LDVI &
  \\
  \includegraphics[width=0.18\textwidth]{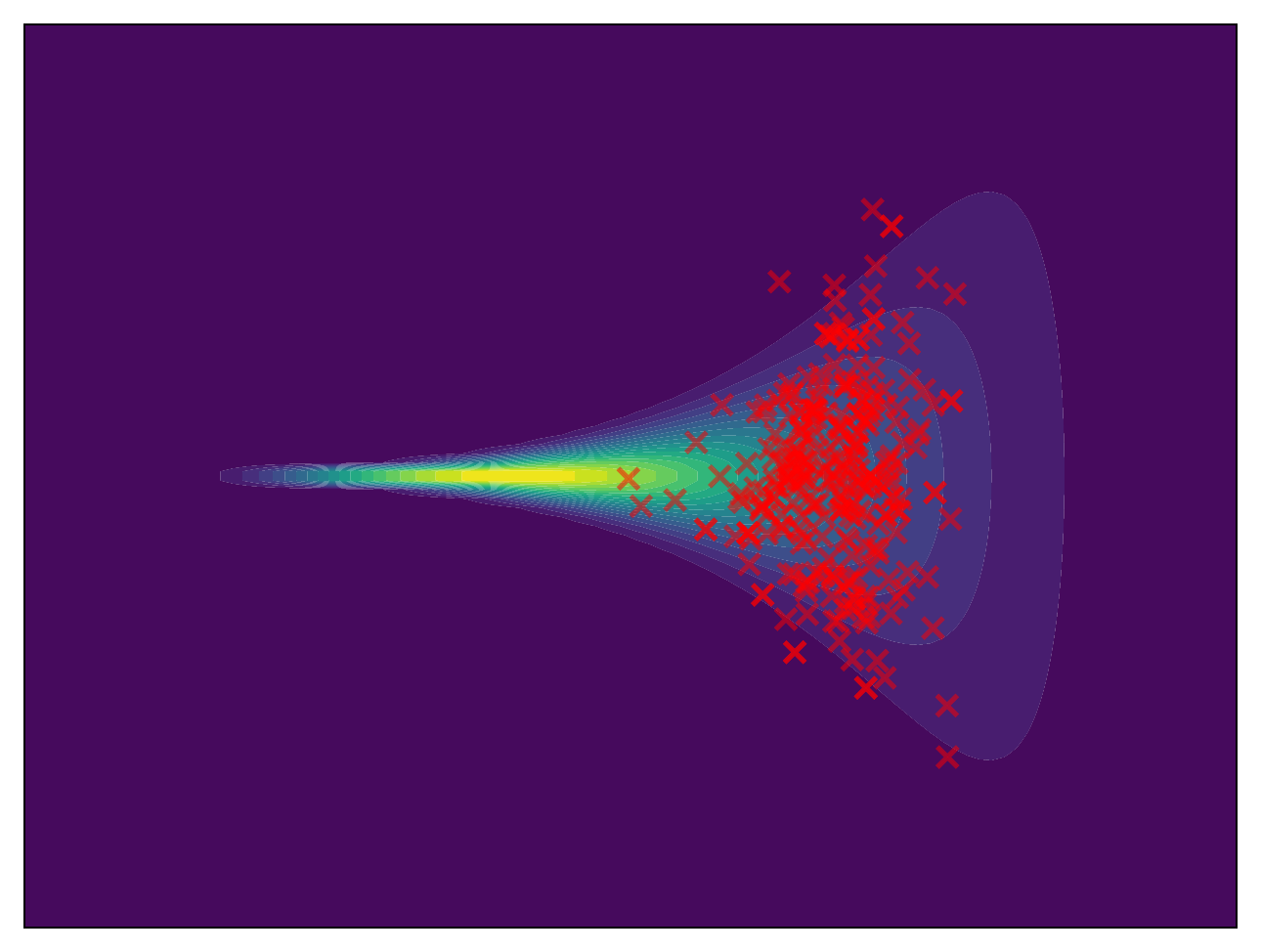} &   
  \includegraphics[width=0.18\textwidth]{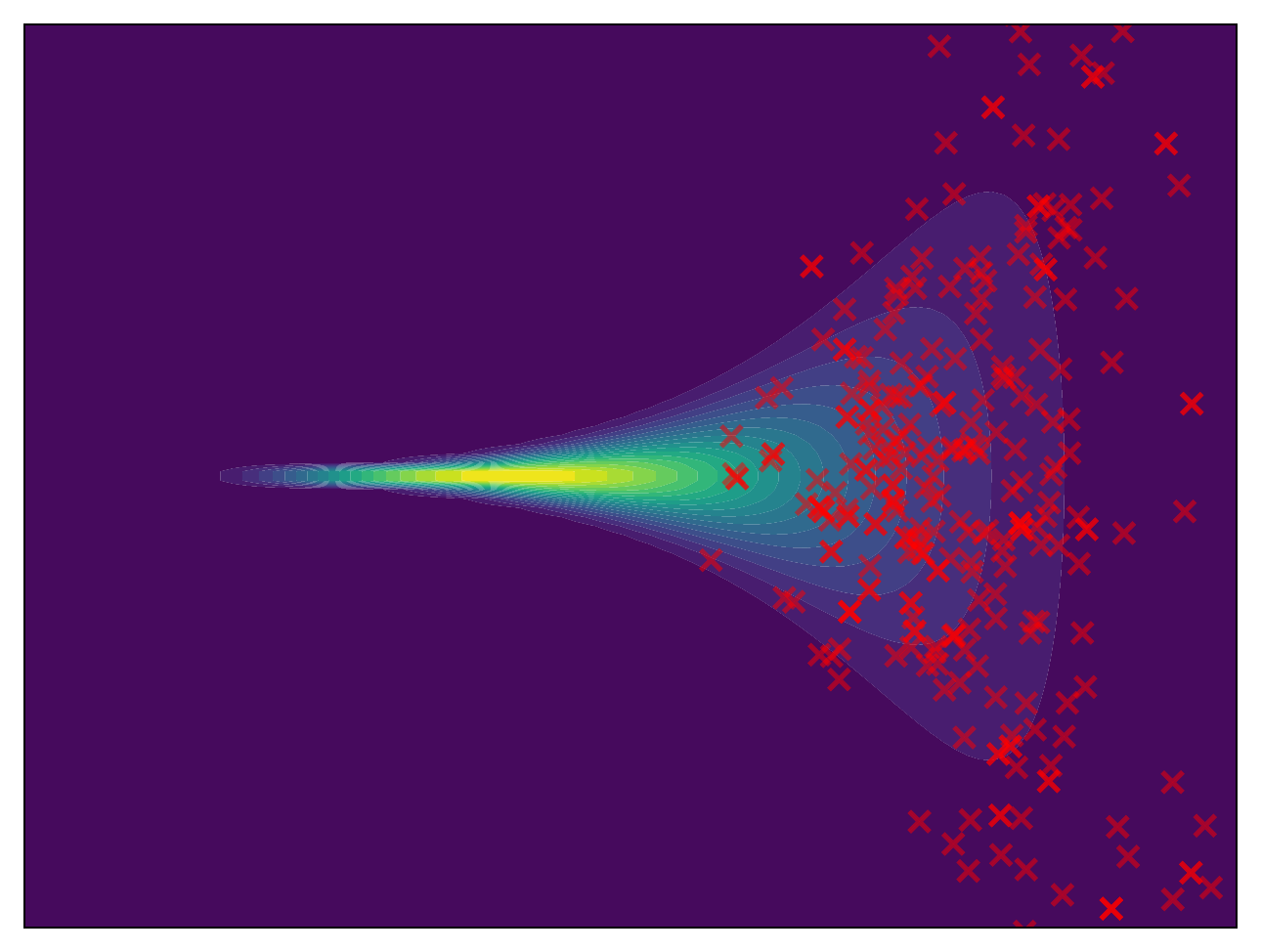} &   
  \includegraphics[width=0.18\textwidth]{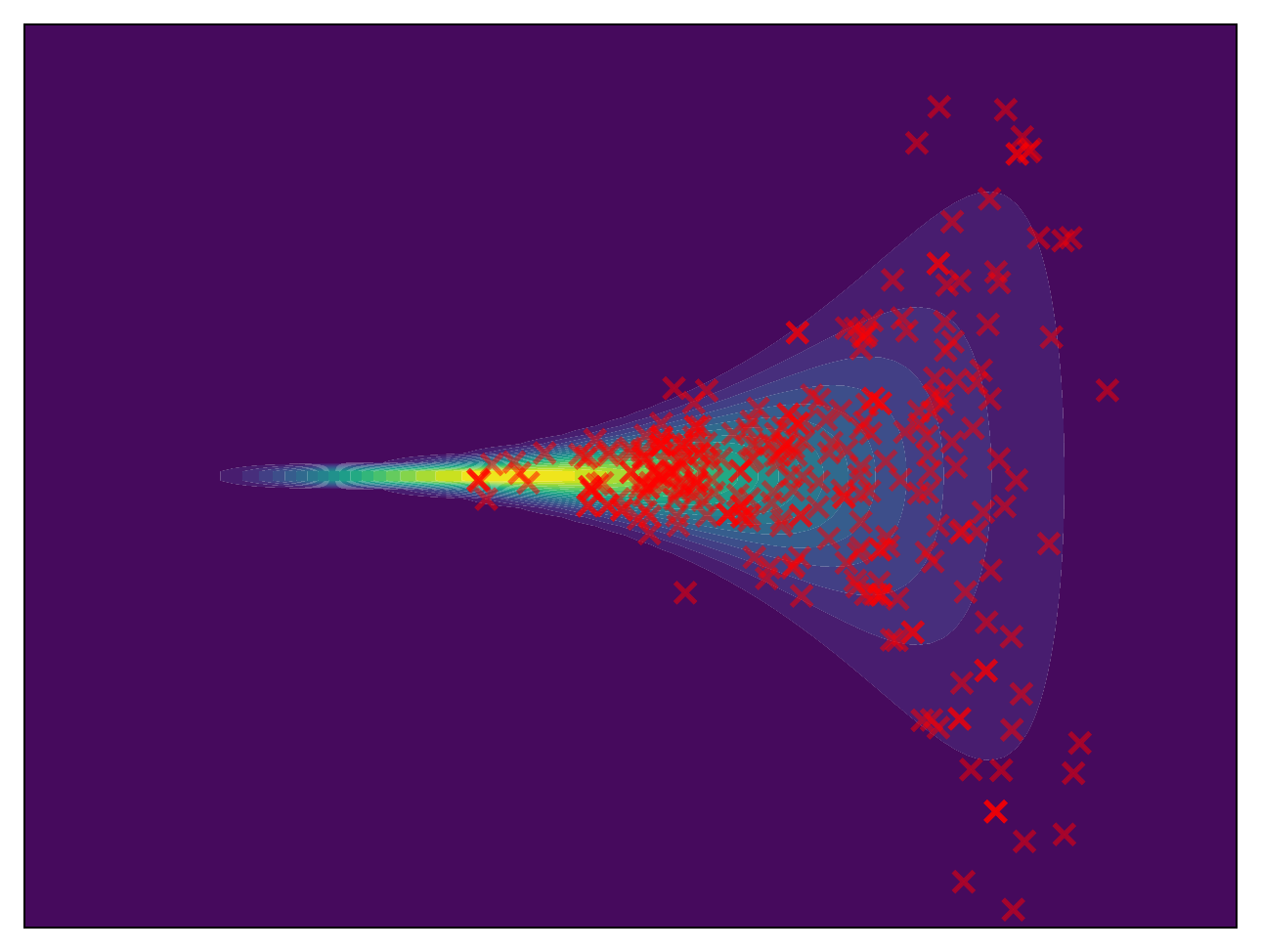} &   
  \includegraphics[width=0.18\textwidth]{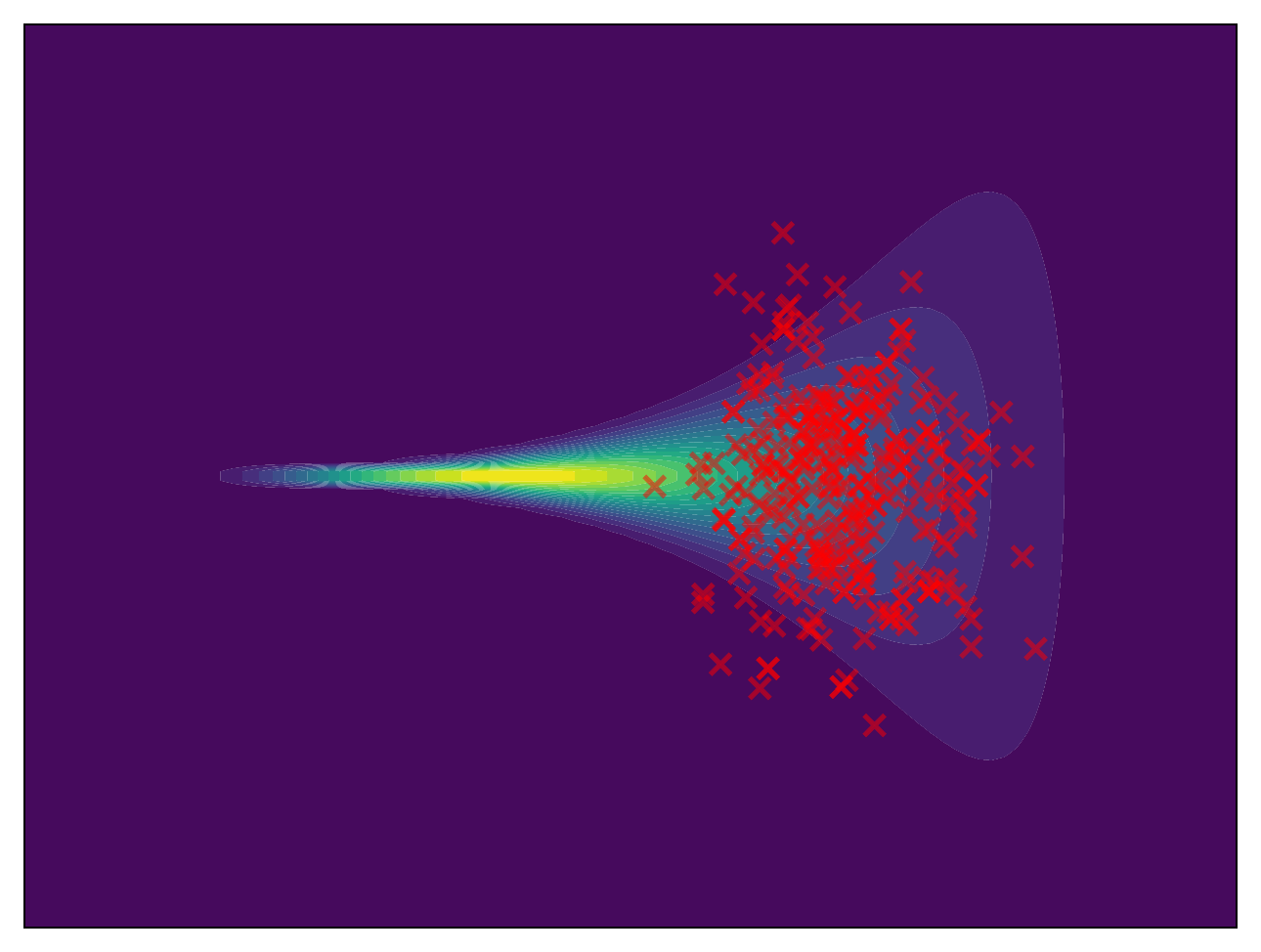} &      
  \\
  \small PIS &
  \small DIS &  
  \small DDS &  
  \small GBS &  
  \\
  \midrule
  \includegraphics[width=0.18\textwidth]{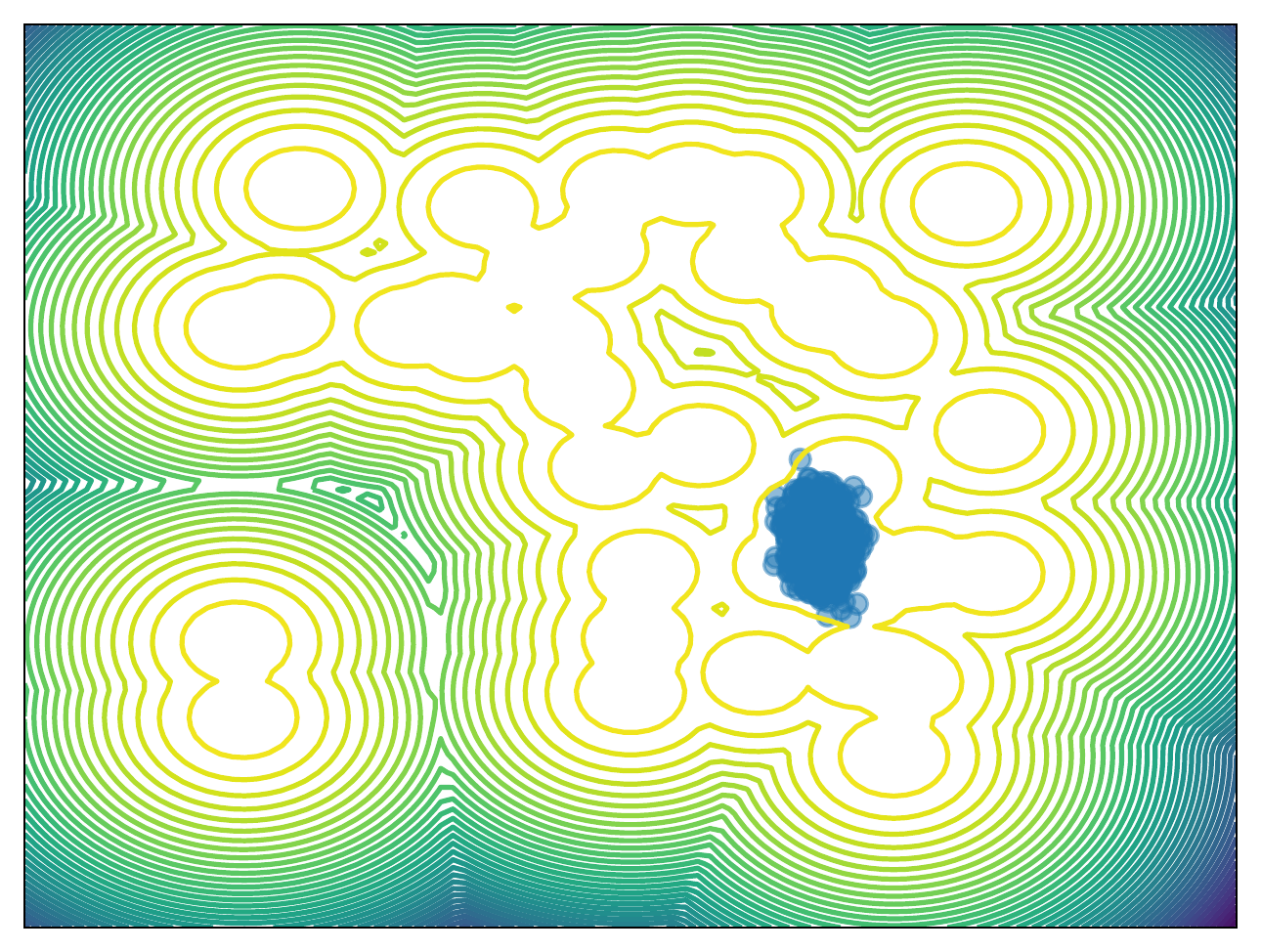} &   
  \includegraphics[width=0.18\textwidth]{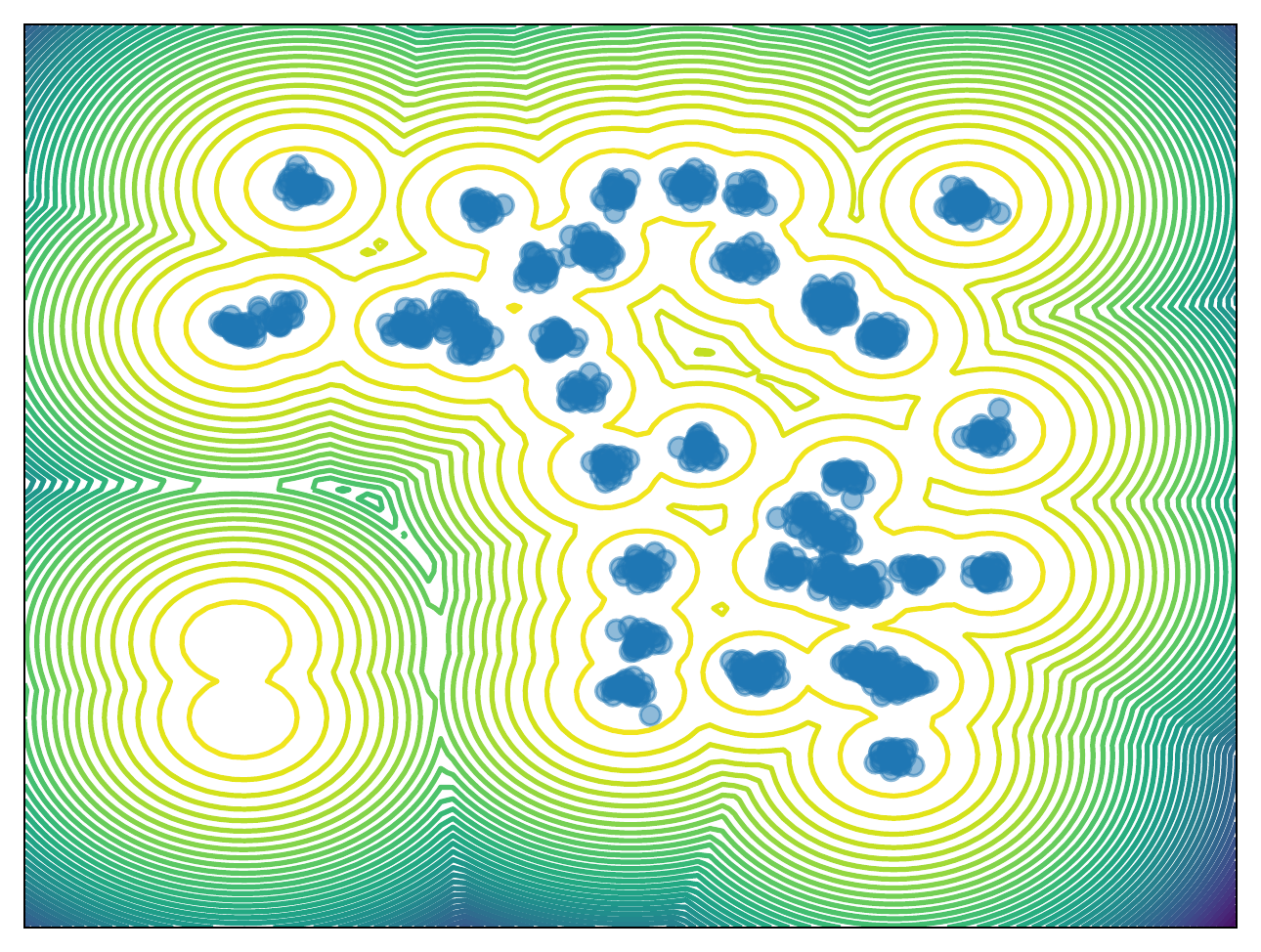} &   
  \includegraphics[width=0.18\textwidth]{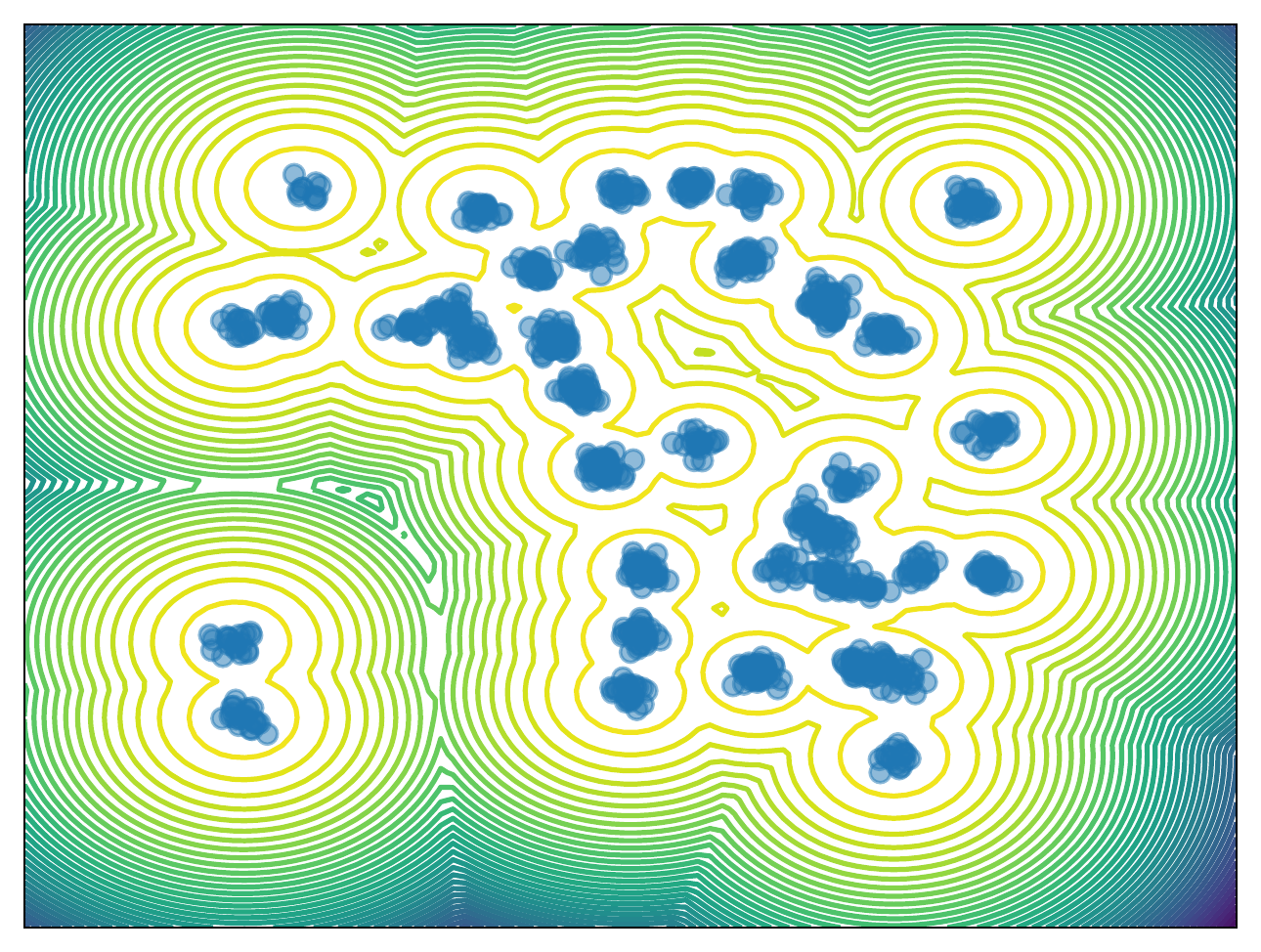} &   
  \includegraphics[width=0.18\textwidth]{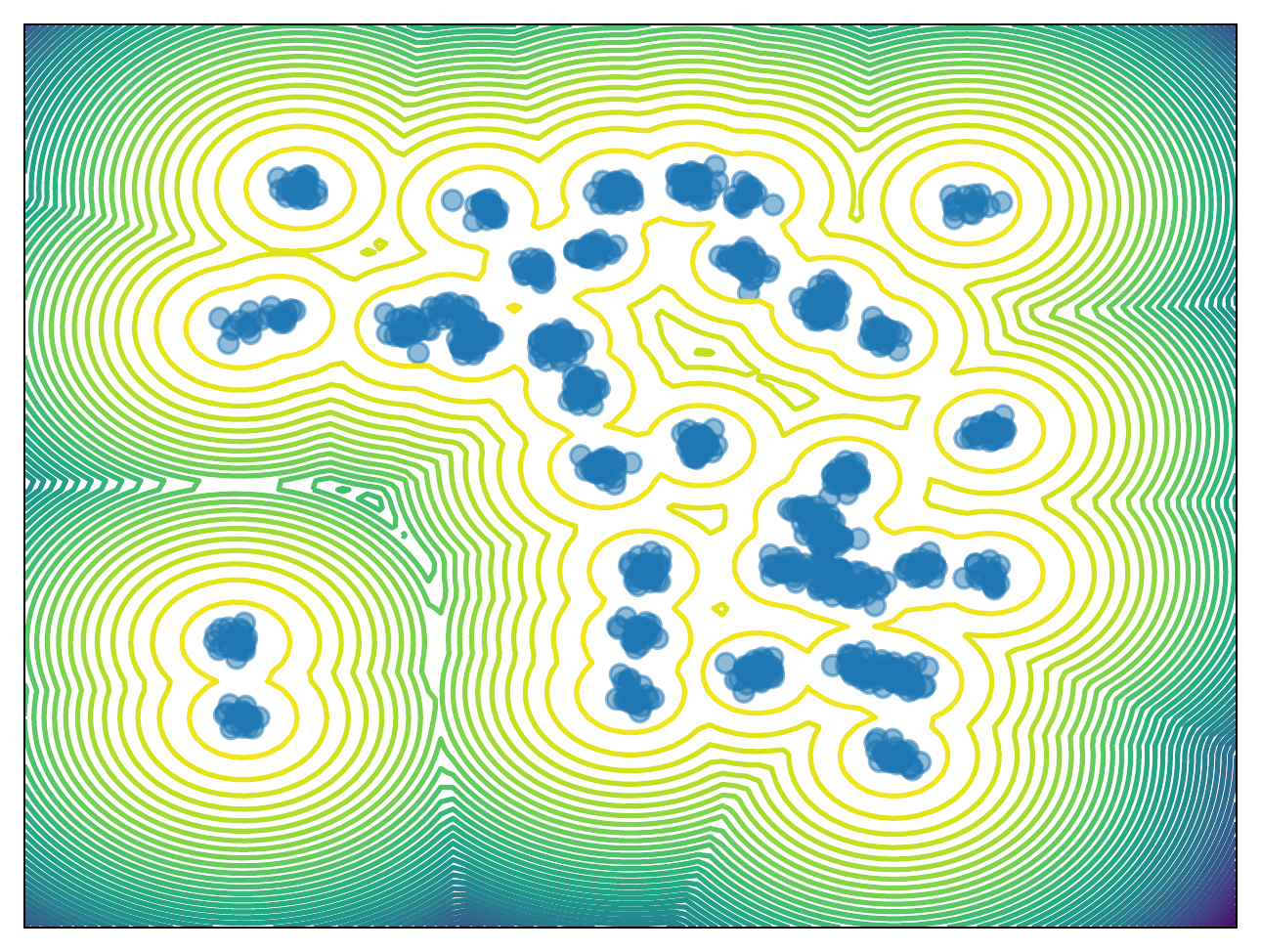} &   
  \\
  \small MFVI &
  \small GMMVI &
  \small SMC &
  \small AFT &
  \\
  \includegraphics[width=0.18\textwidth]{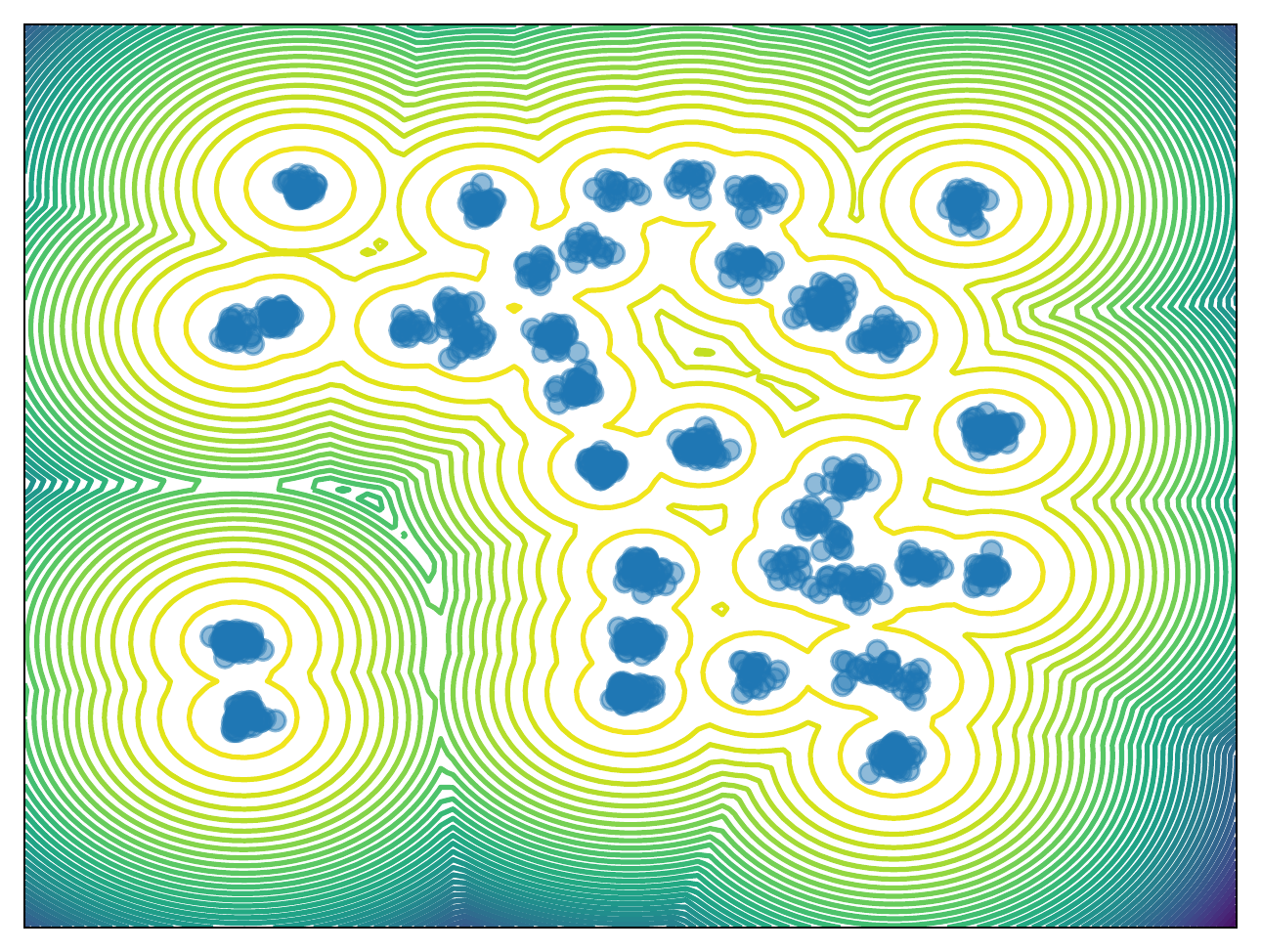} &   
  \includegraphics[width=0.18\textwidth]{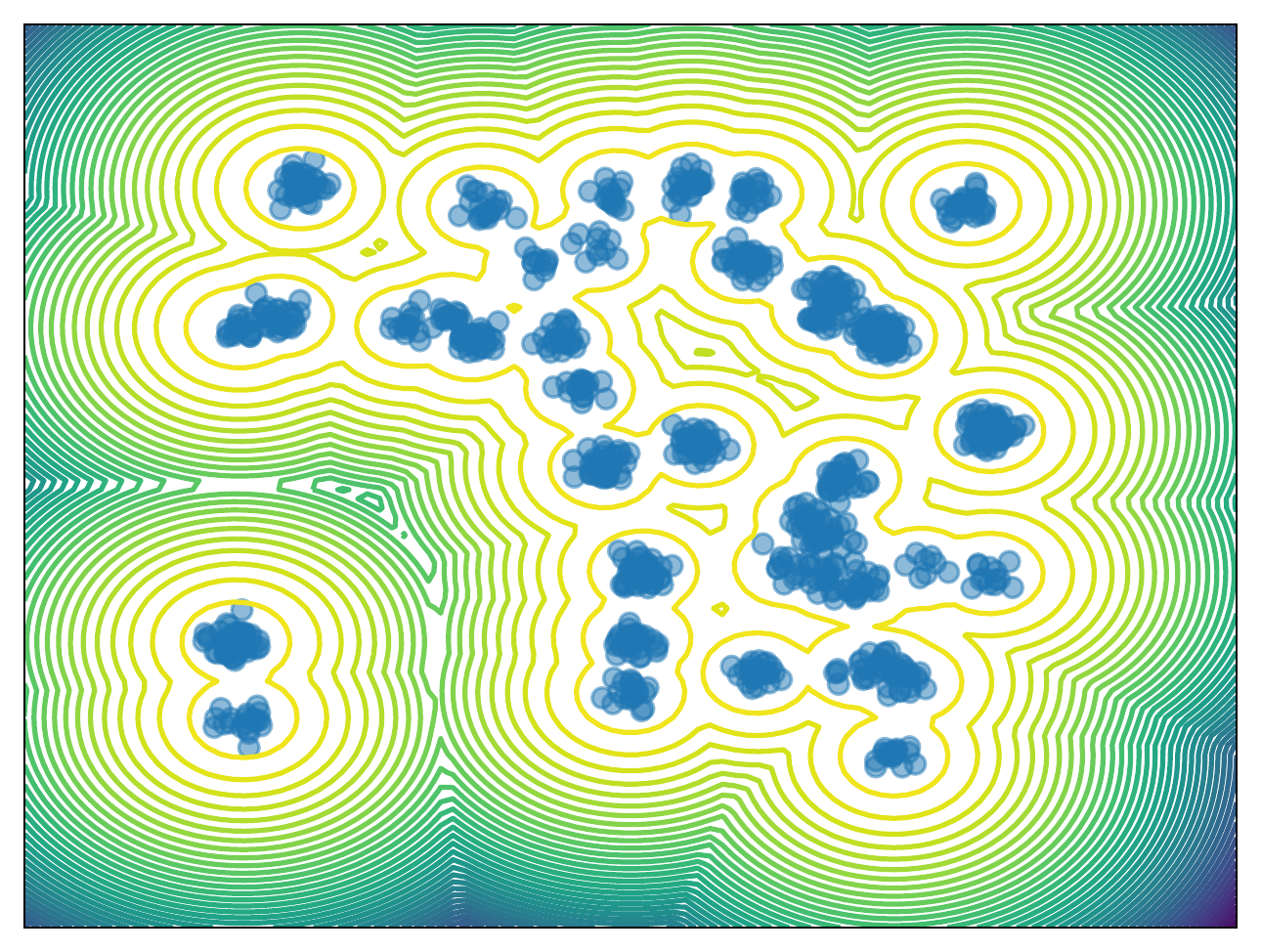} &   
  \includegraphics[width=0.18\textwidth]{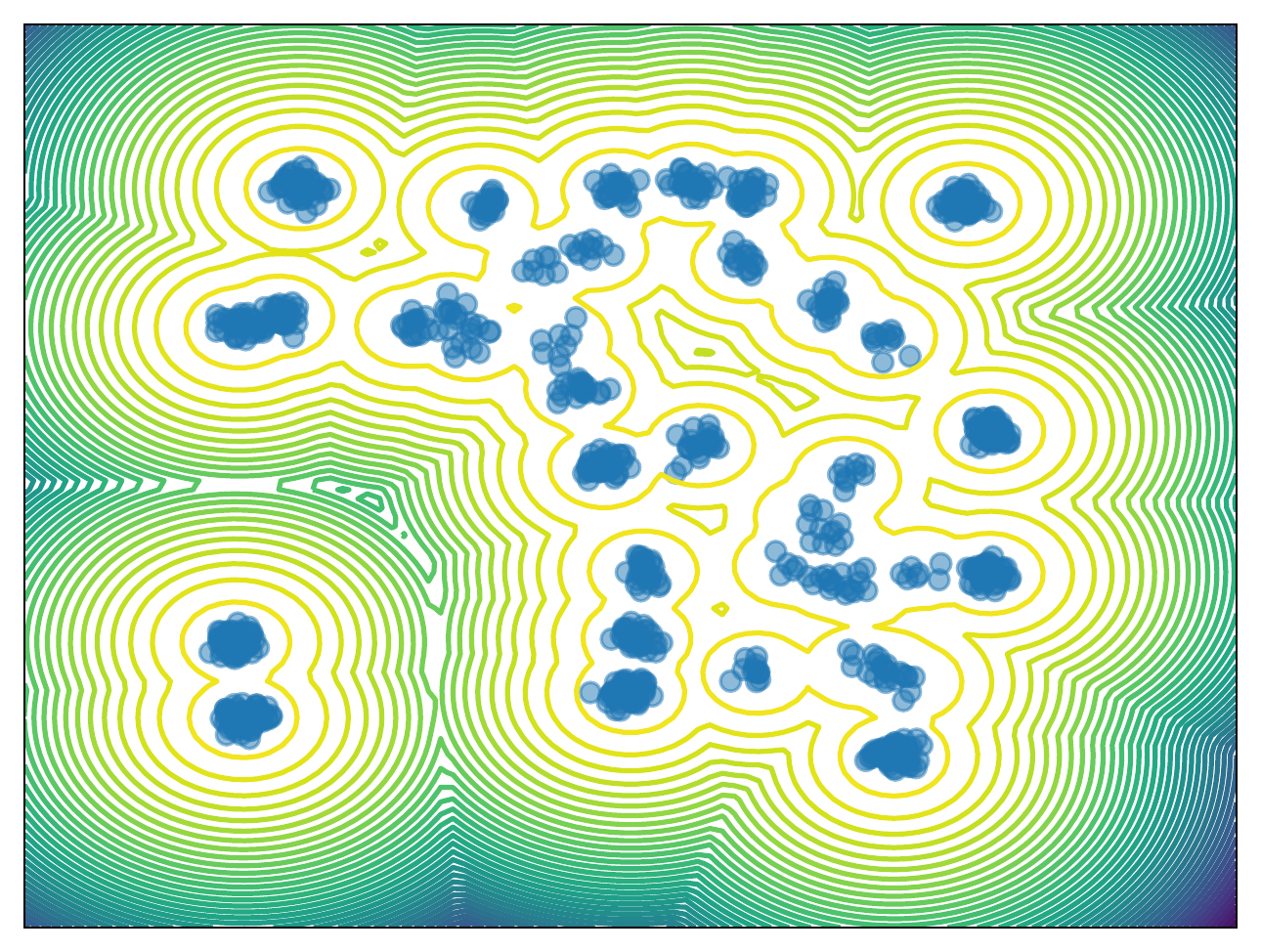} &   
  \includegraphics[width=0.18\textwidth]{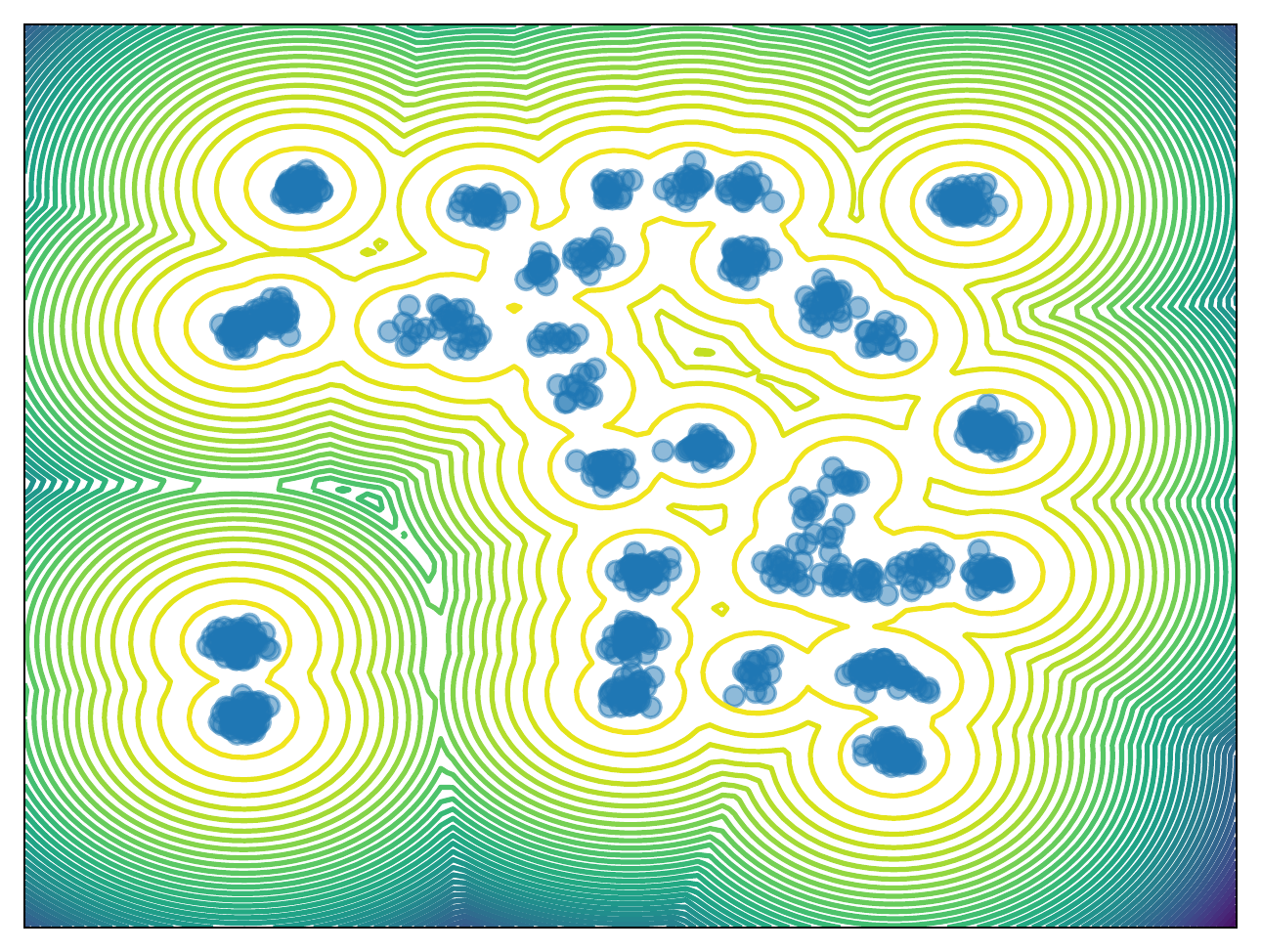} &   
  \\
  \small CRAFT &
  \small FAB &
  \small MCD &
  \small LDVI &
  \\
  \includegraphics[width=0.18\textwidth]{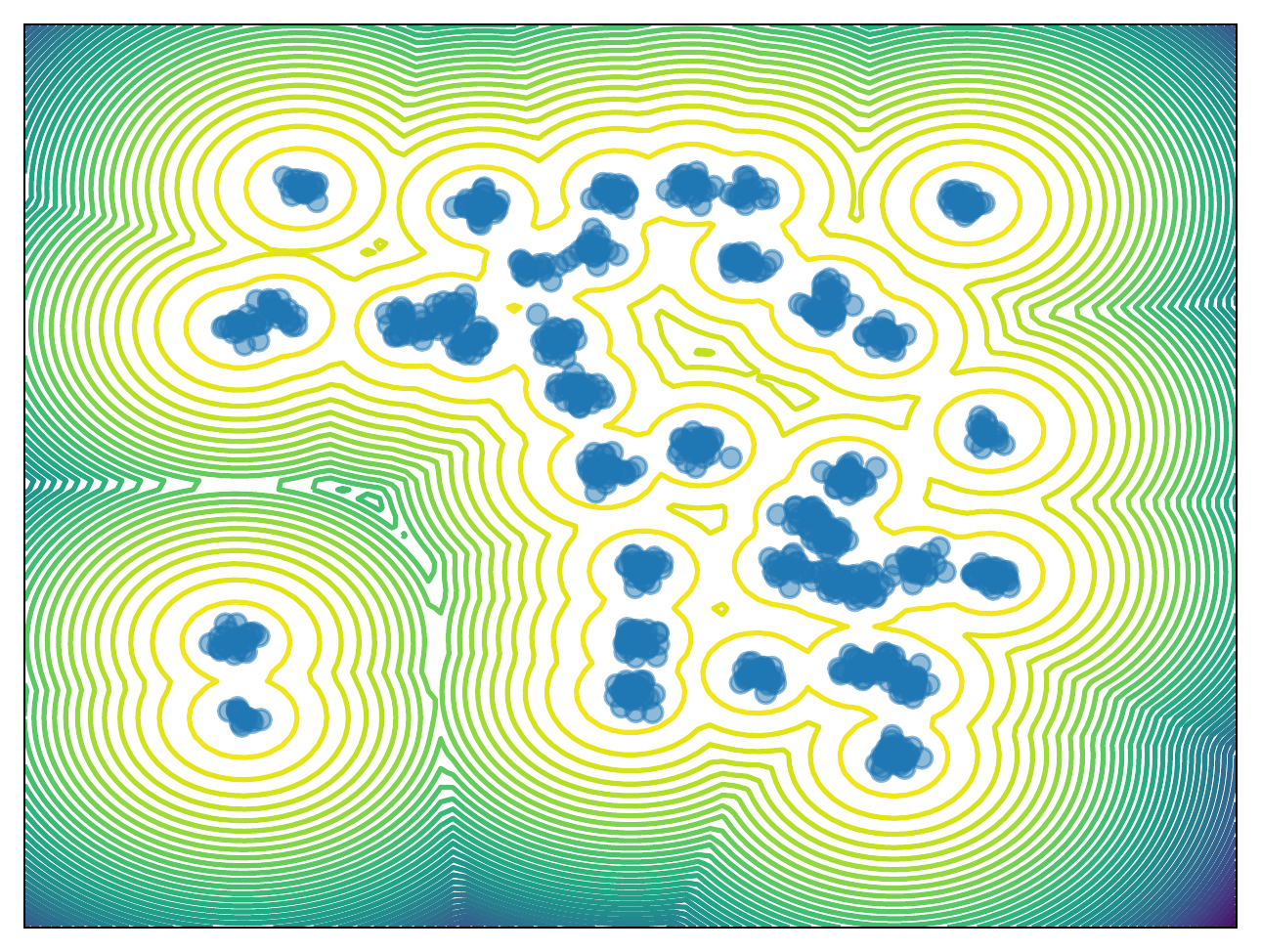} &   
  \includegraphics[width=0.18\textwidth]{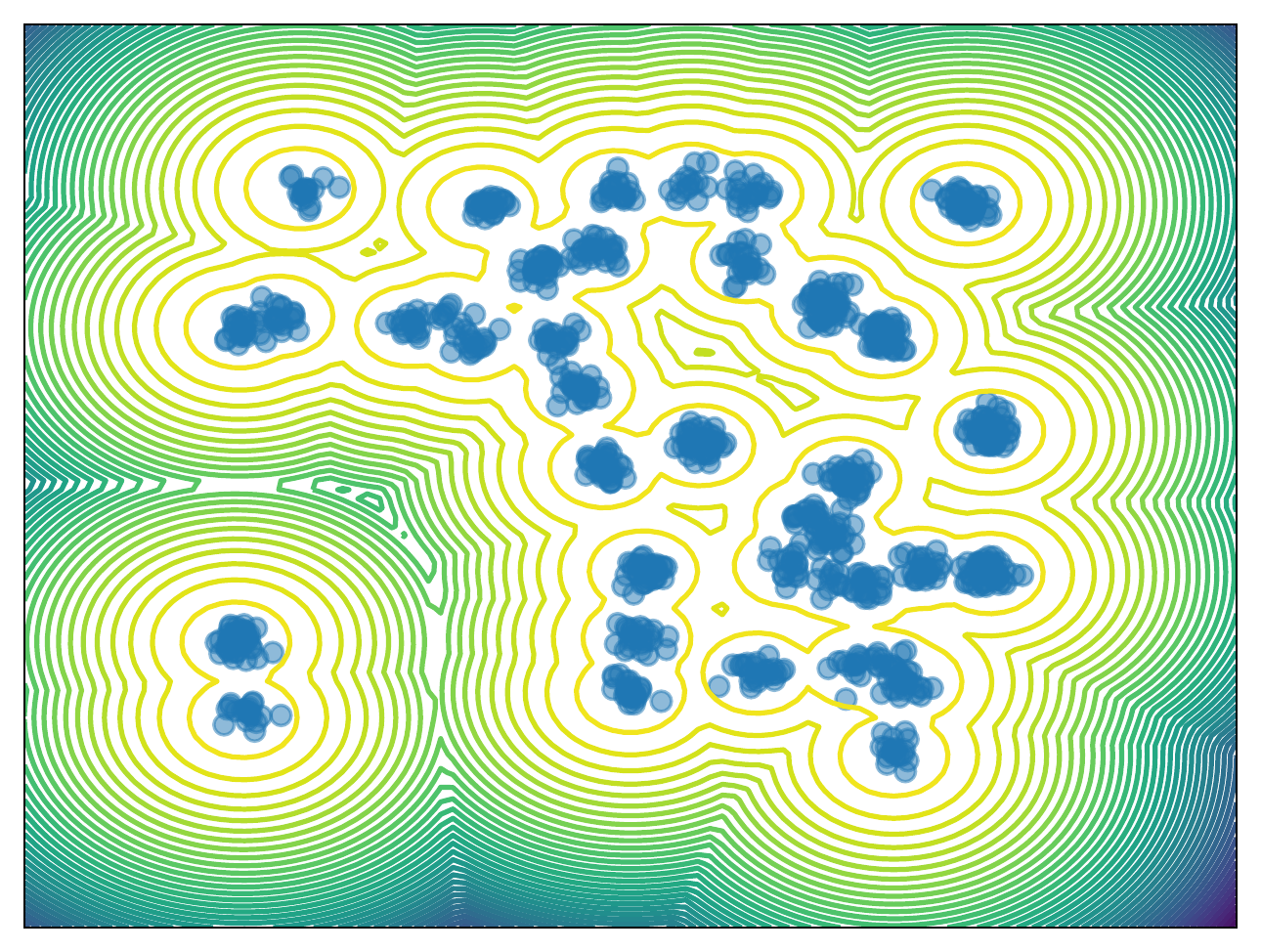} &   
  \includegraphics[width=0.18\textwidth]{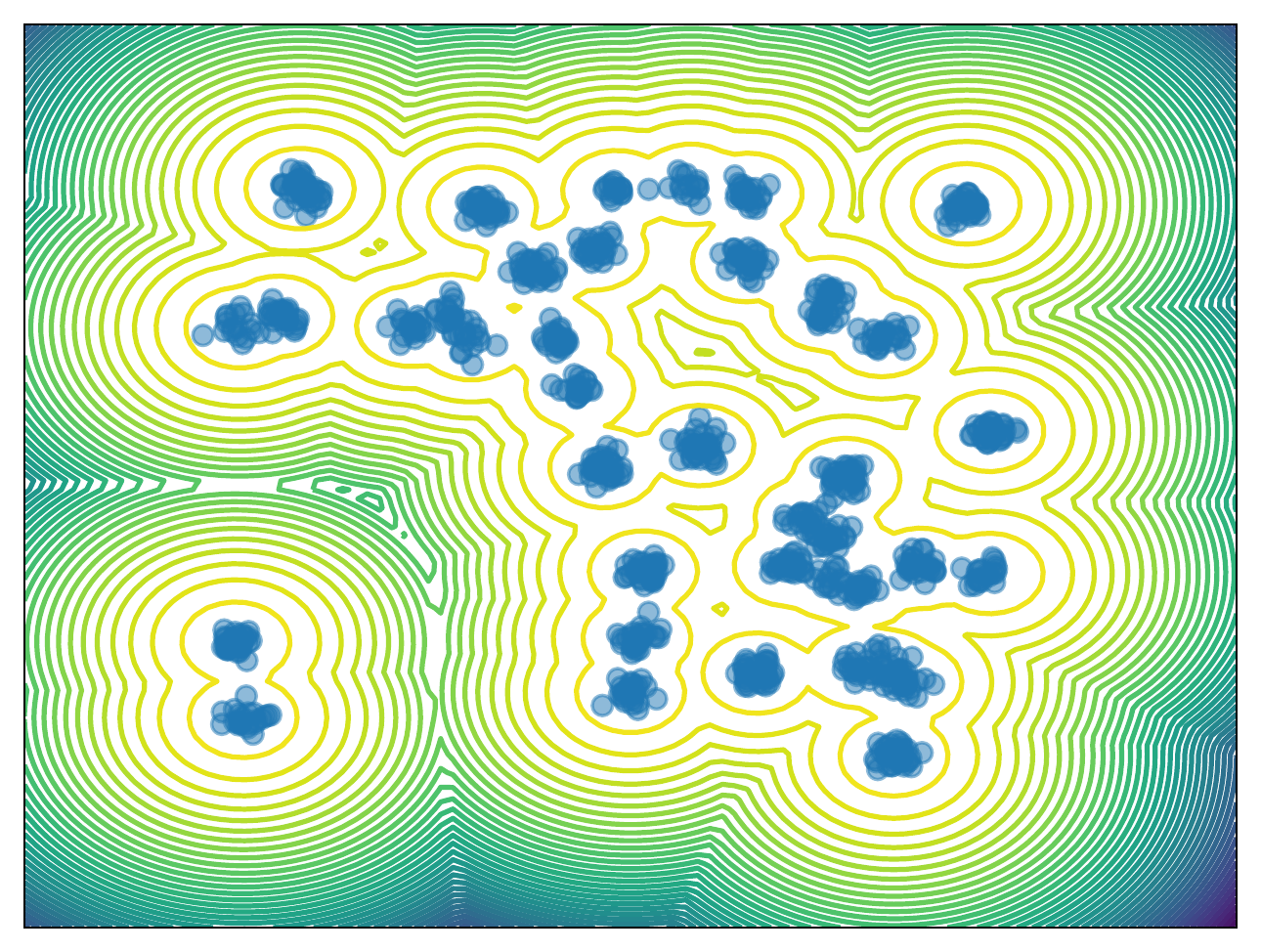} &   
  \includegraphics[width=0.18\textwidth]{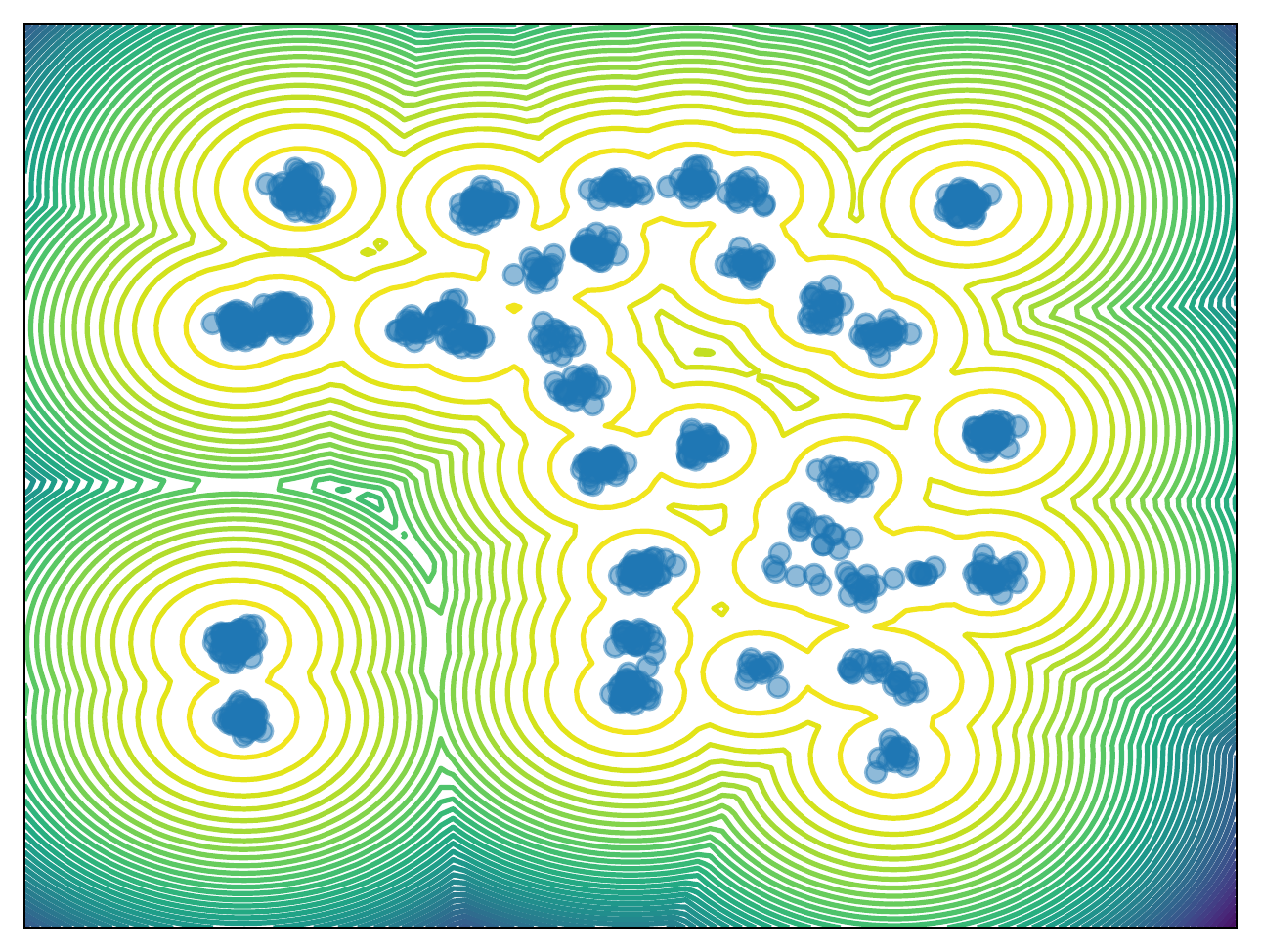} &    
  \\
  \small PIS &
  \small DIS &  
  \small DDS &  
  \small GBS &  
  \\
  \bottomrule
\end{tabular}
\caption{Visualization of samples drawn from different sampling methods for Funnel (top) and MoG (bottom).}
\label{fig:samples_mog_funnel}
\end{figure*}

\begin{figure*}
\centering
\setlength{\tabcolsep}{1pt}
\begin{tabular}{ccccc}
\toprule
  \includegraphics[width=0.18\textwidth]{figures/illustrations/digits/mfvi_digitsnice.pdf} &   
  \includegraphics[width=0.18\textwidth]{figures/illustrations/digits/gmmvi_jax_digitsnice.pdf} &   
  \includegraphics[width=0.18\textwidth]{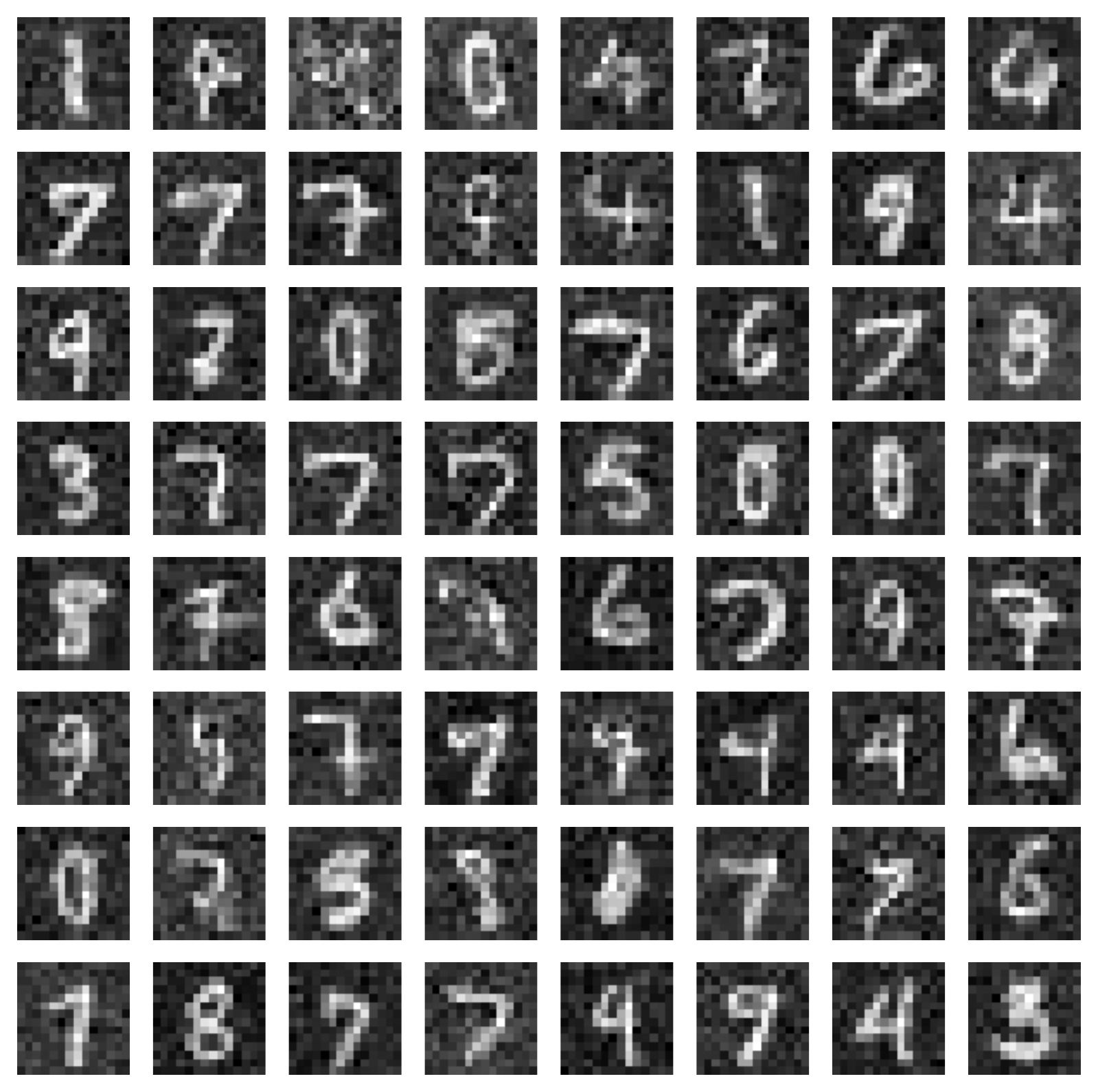} &   
  \includegraphics[width=0.18\textwidth]{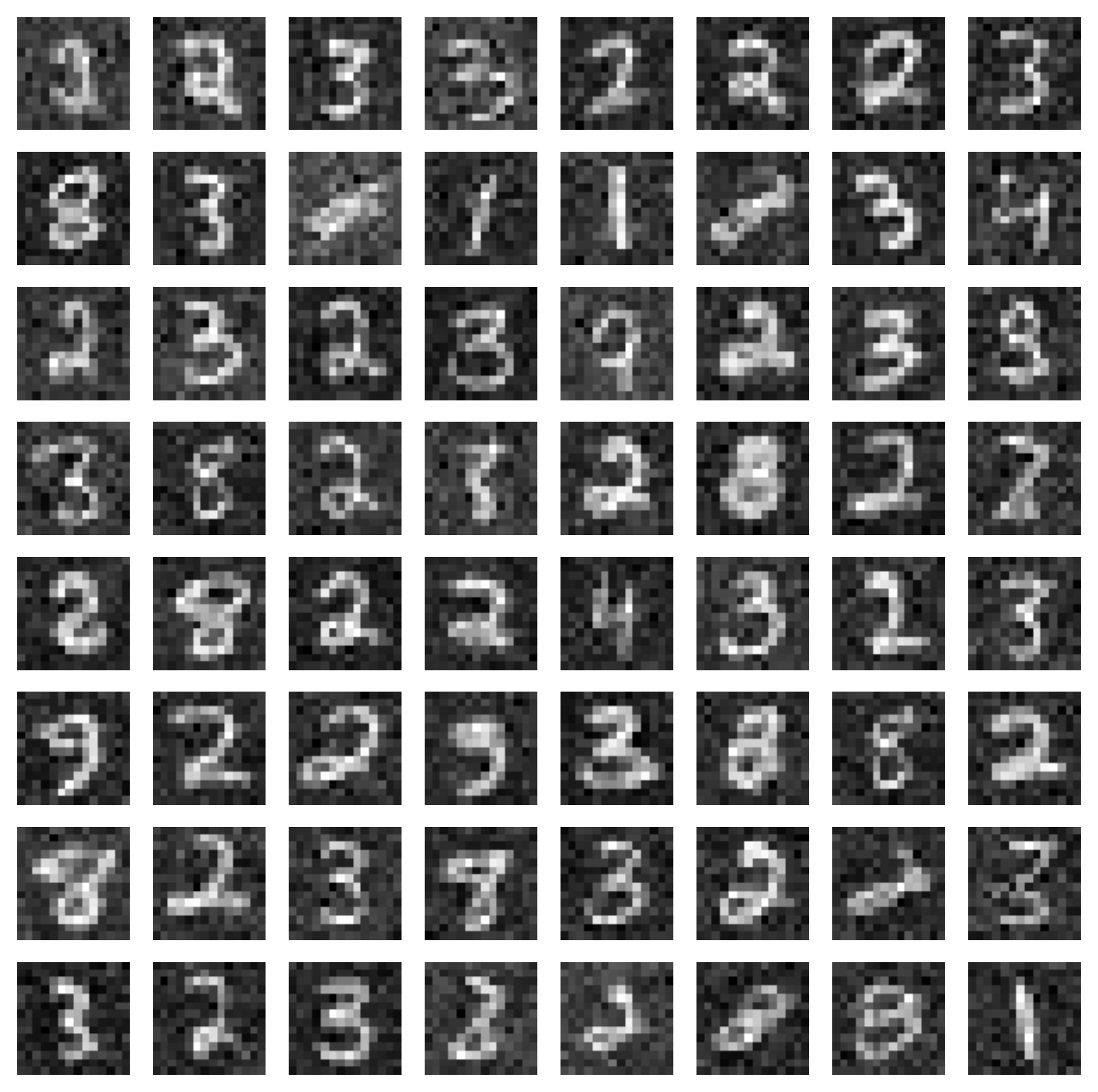} &   
  \\
  \small MFVI &
  \small GMMVI &
  \small SMC &
  \small AFT &
  \\
  \includegraphics[width=0.18\textwidth]{figures/illustrations/digits/craft_digitsnice.pdf} &   
  \includegraphics[width=0.18\textwidth]{figures/illustrations/digits/fab_digitsnice.pdf} &   
  \includegraphics[width=0.18\textwidth]{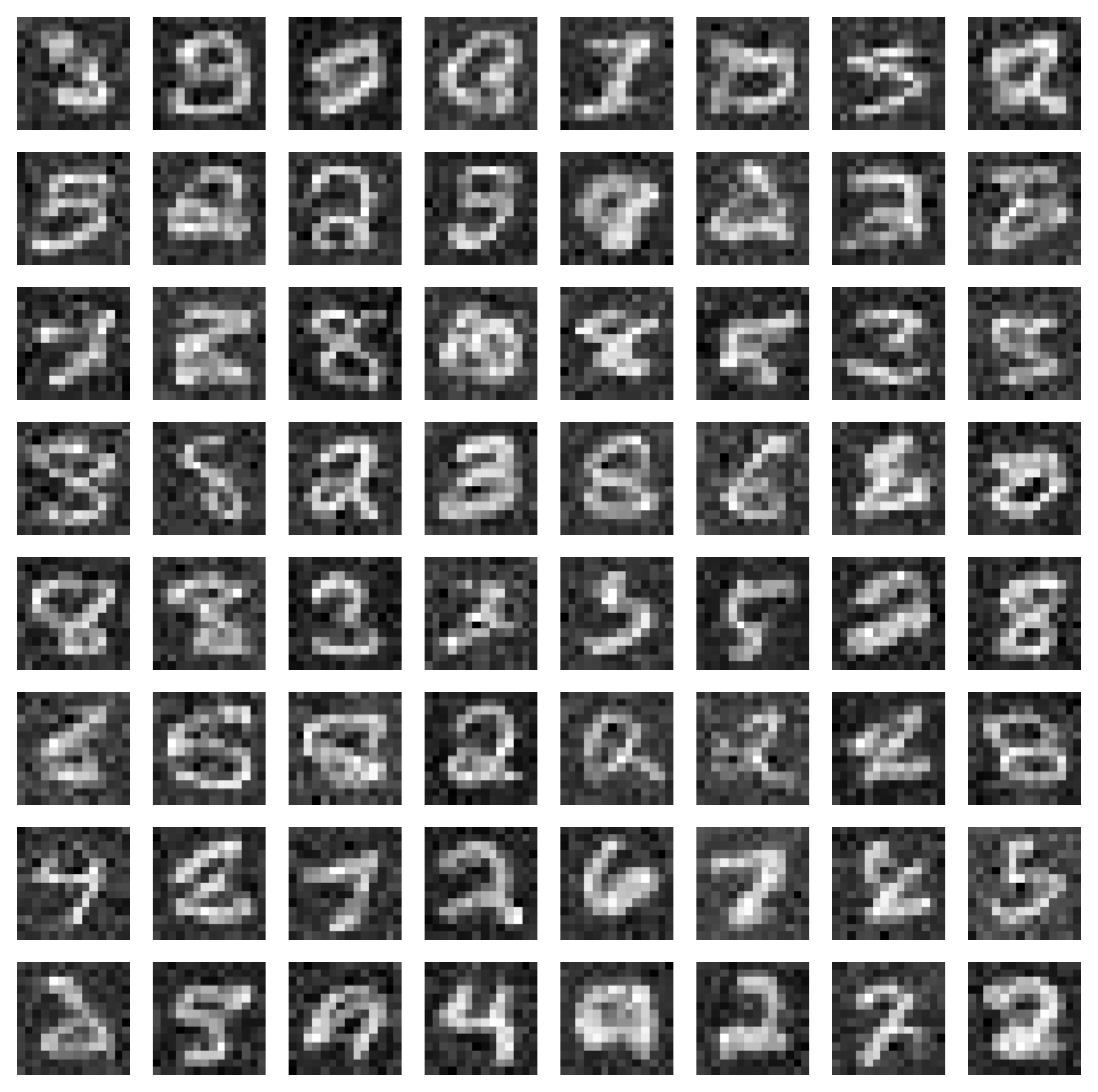} &   
  \includegraphics[width=0.18\textwidth]{figures/illustrations/digits/ldvi_digitsnice.pdf} &   
  \\
  \small CRAFT &
  \small FAB &
  \small MCD &
  \small LDVI &
  \\
  \includegraphics[width=0.18\textwidth]{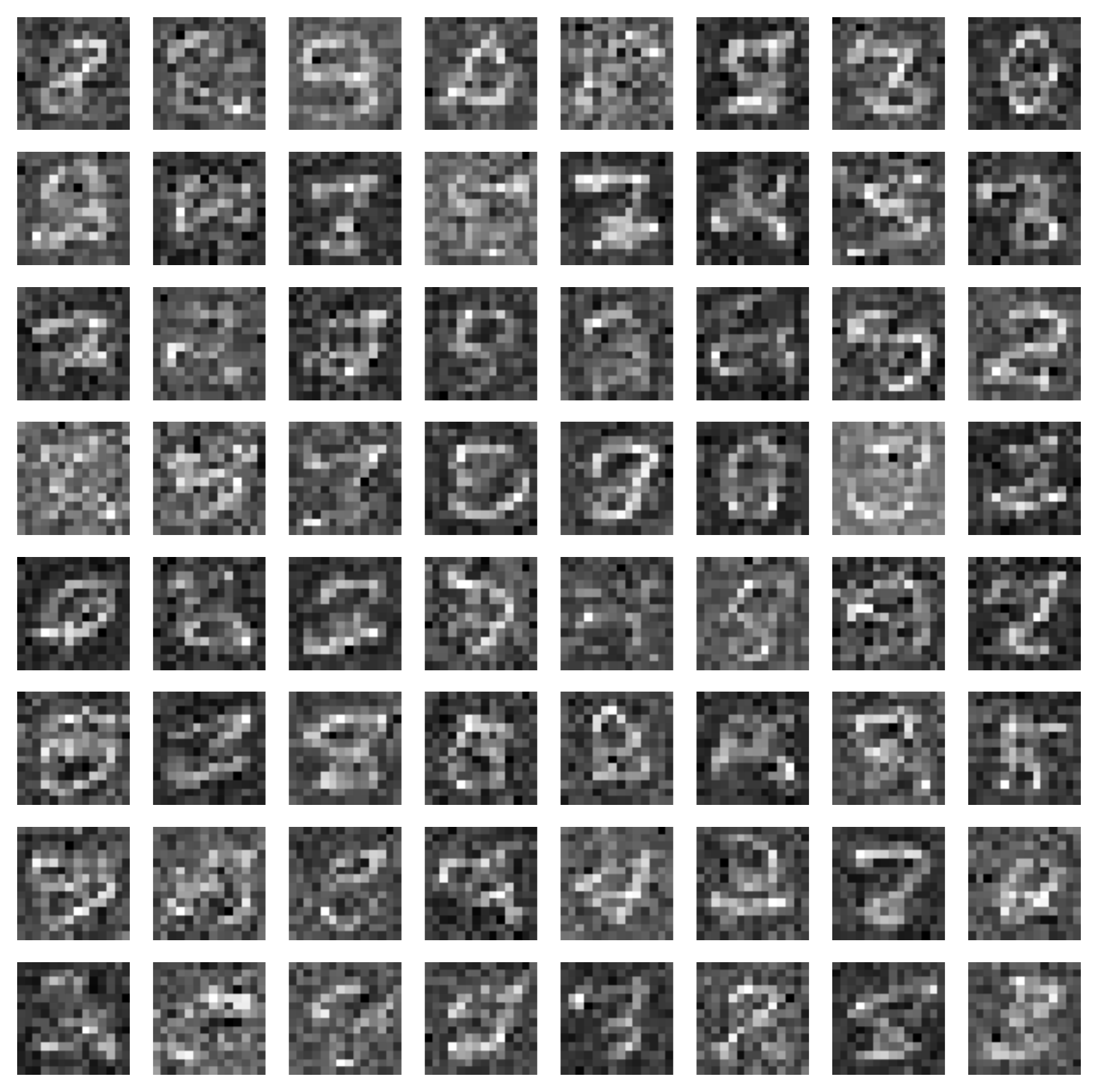} &   
  \includegraphics[width=0.18\textwidth]{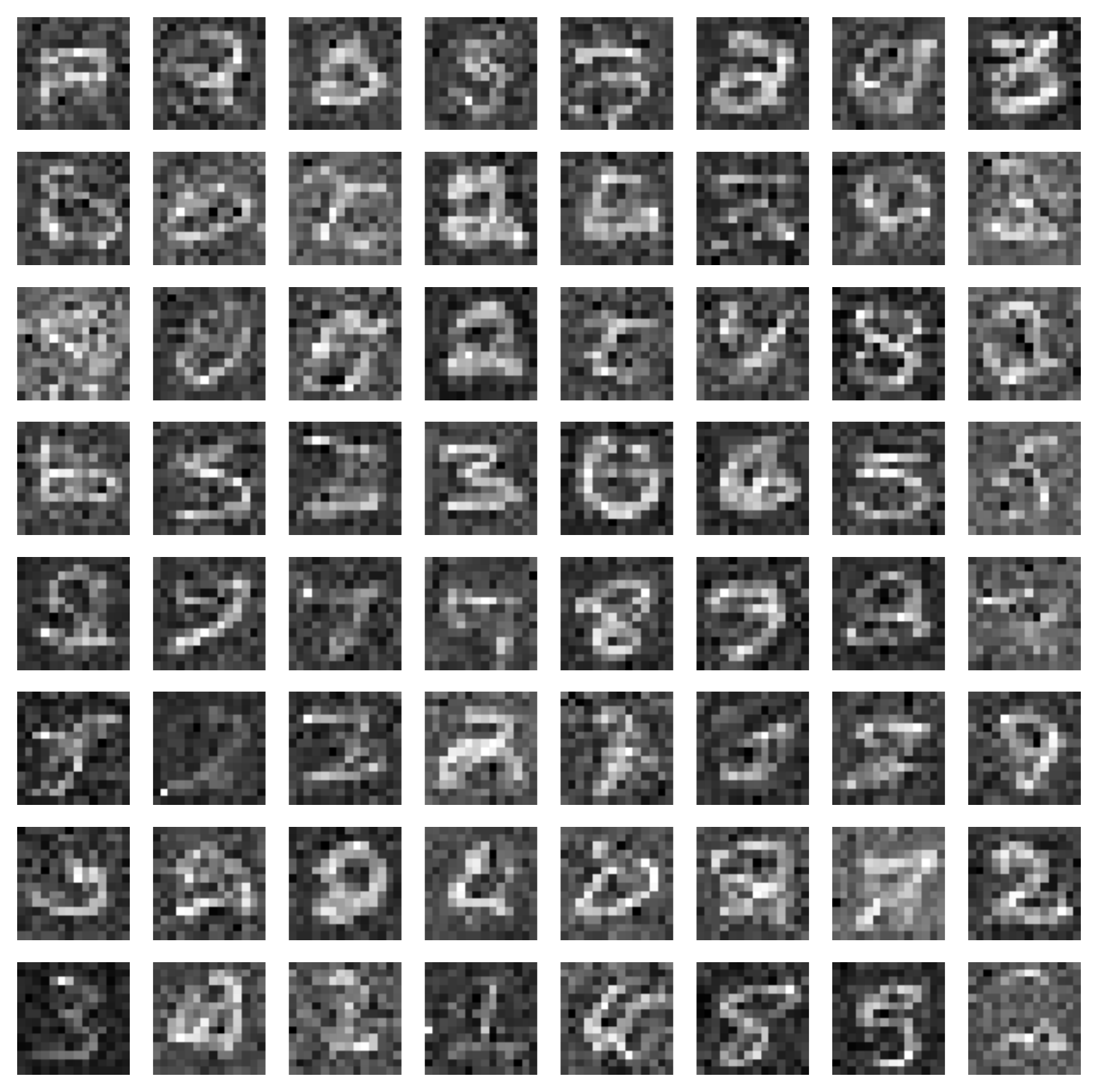} &   
  \includegraphics[width=0.18\textwidth]{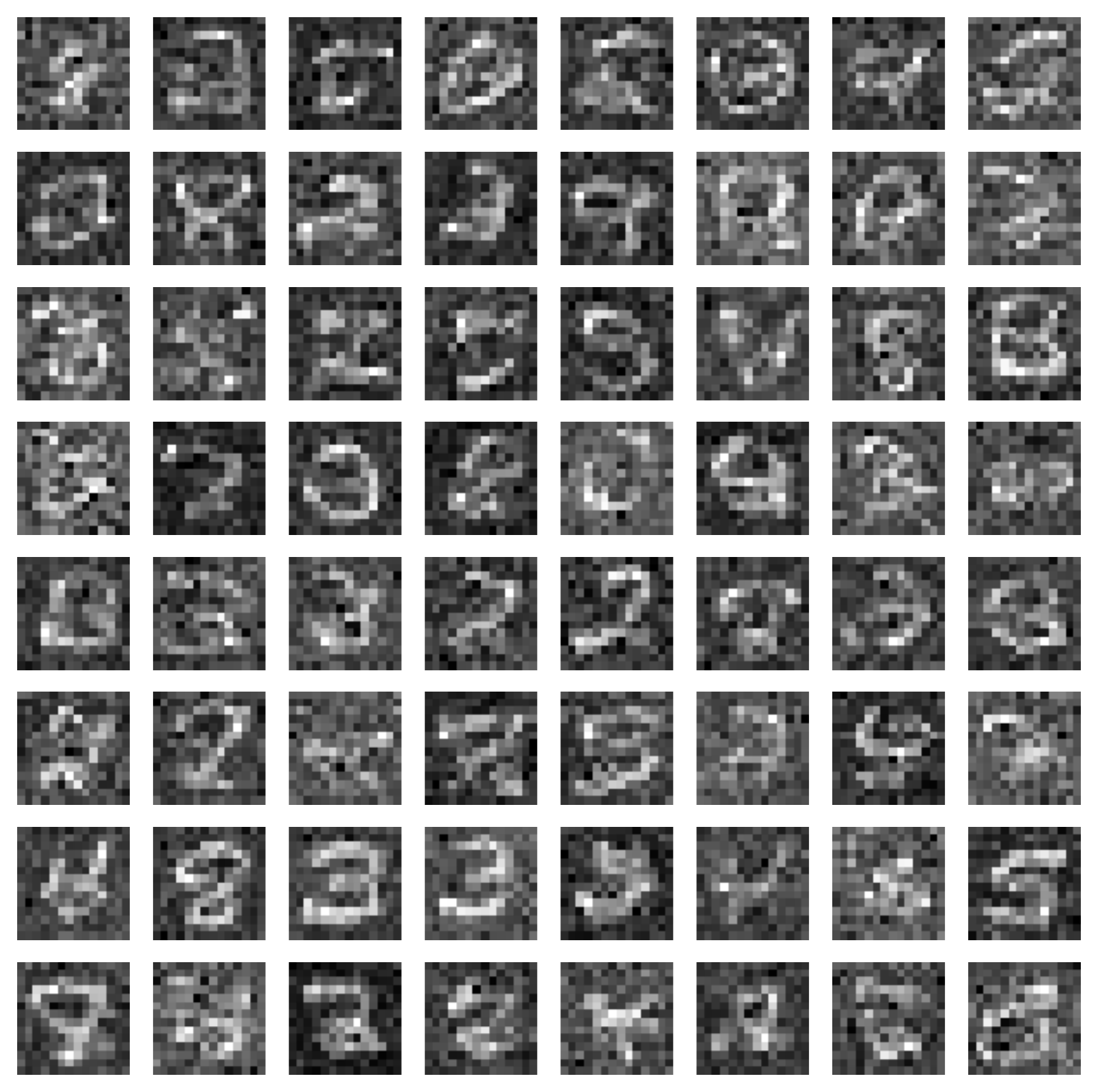} &   
  \includegraphics[width=0.18\textwidth]{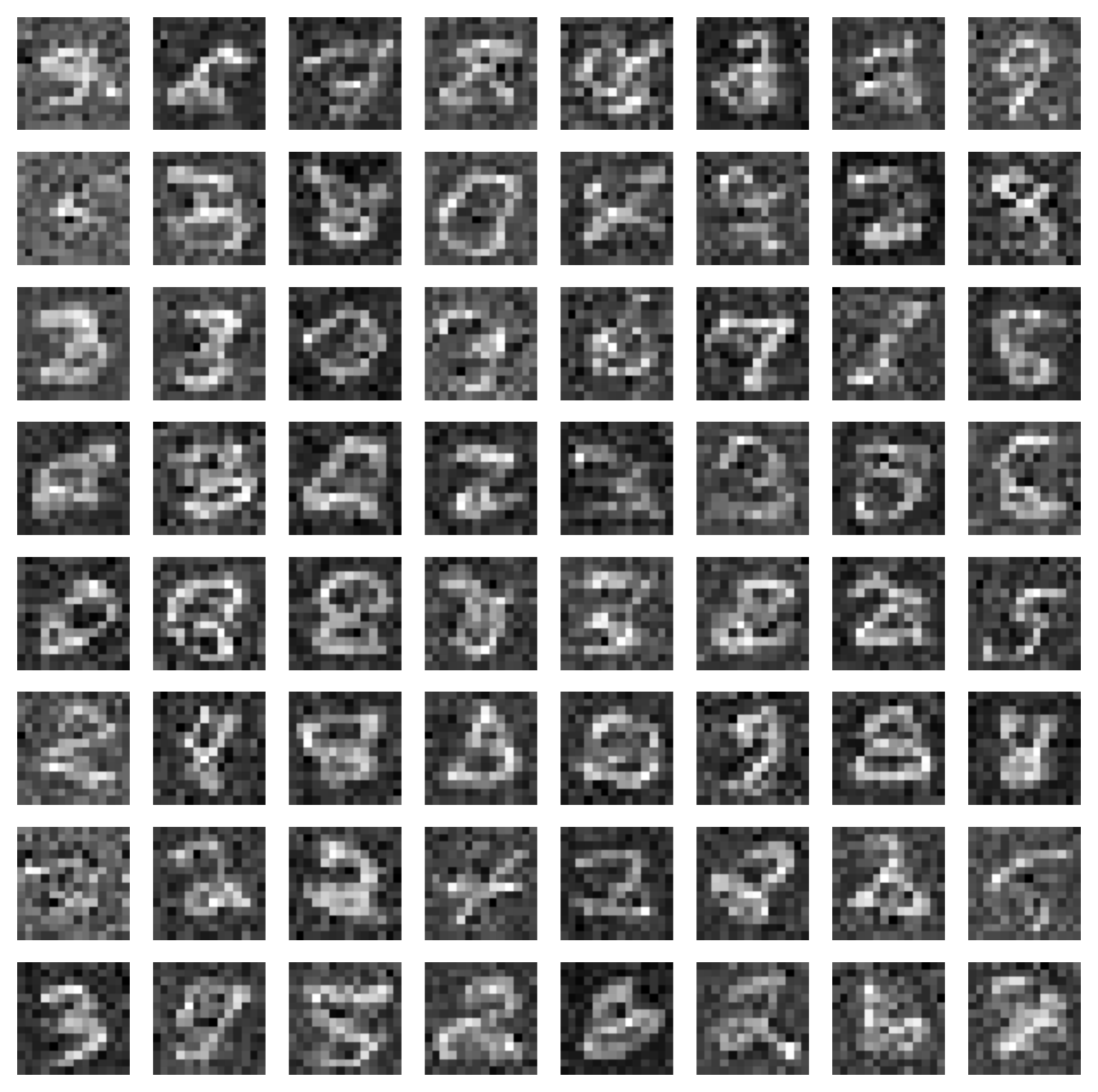} &   
  \\
  \small PIS &
  \small DIS &  
  \small DDS &  
  \small GBS &  
  \\
  \midrule
  \includegraphics[width=0.18\textwidth]{figures/illustrations/fashion/mfvi_fashionnice.pdf} &   
  \includegraphics[width=0.18\textwidth]{figures/illustrations/fashion/gmmvi_jax_fashionnice.pdf} &   
  \includegraphics[width=0.18\textwidth]{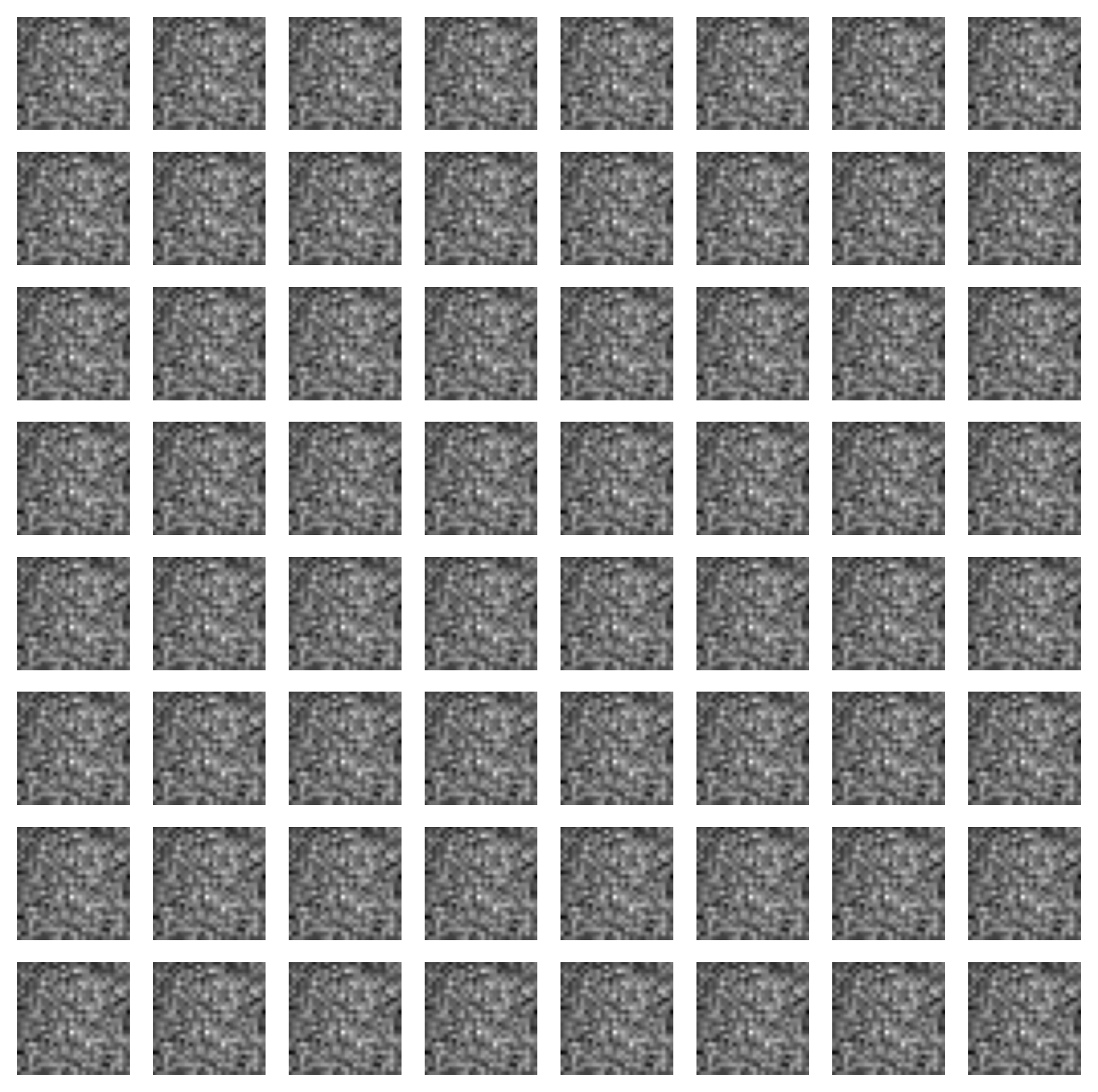} &   
  \includegraphics[width=0.18\textwidth]{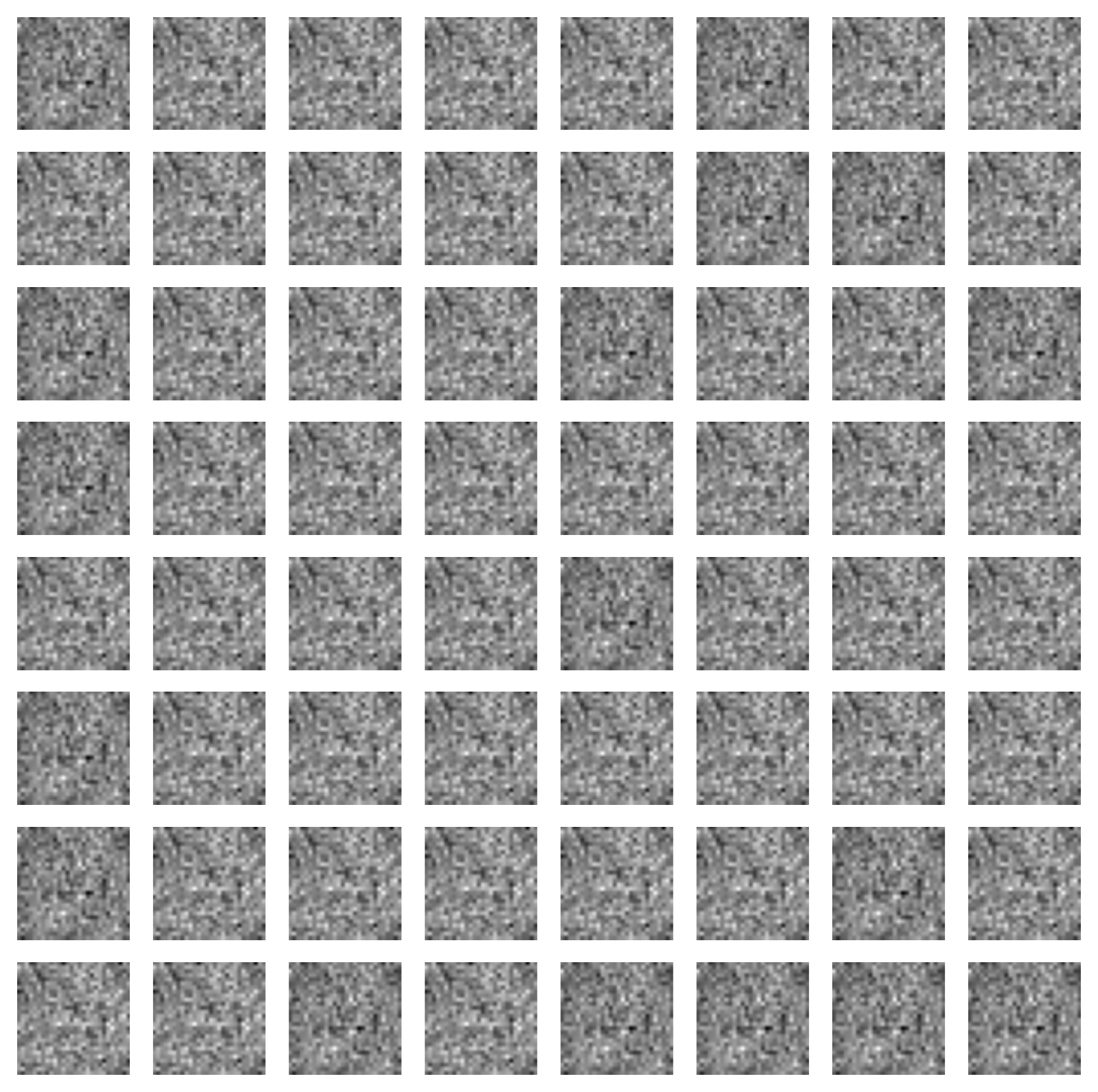} &    
  \\
  \small MFVI &
  \small GMMVI &
  \small SMC &
  \small AFT &
  \\
  \includegraphics[width=0.18\textwidth]{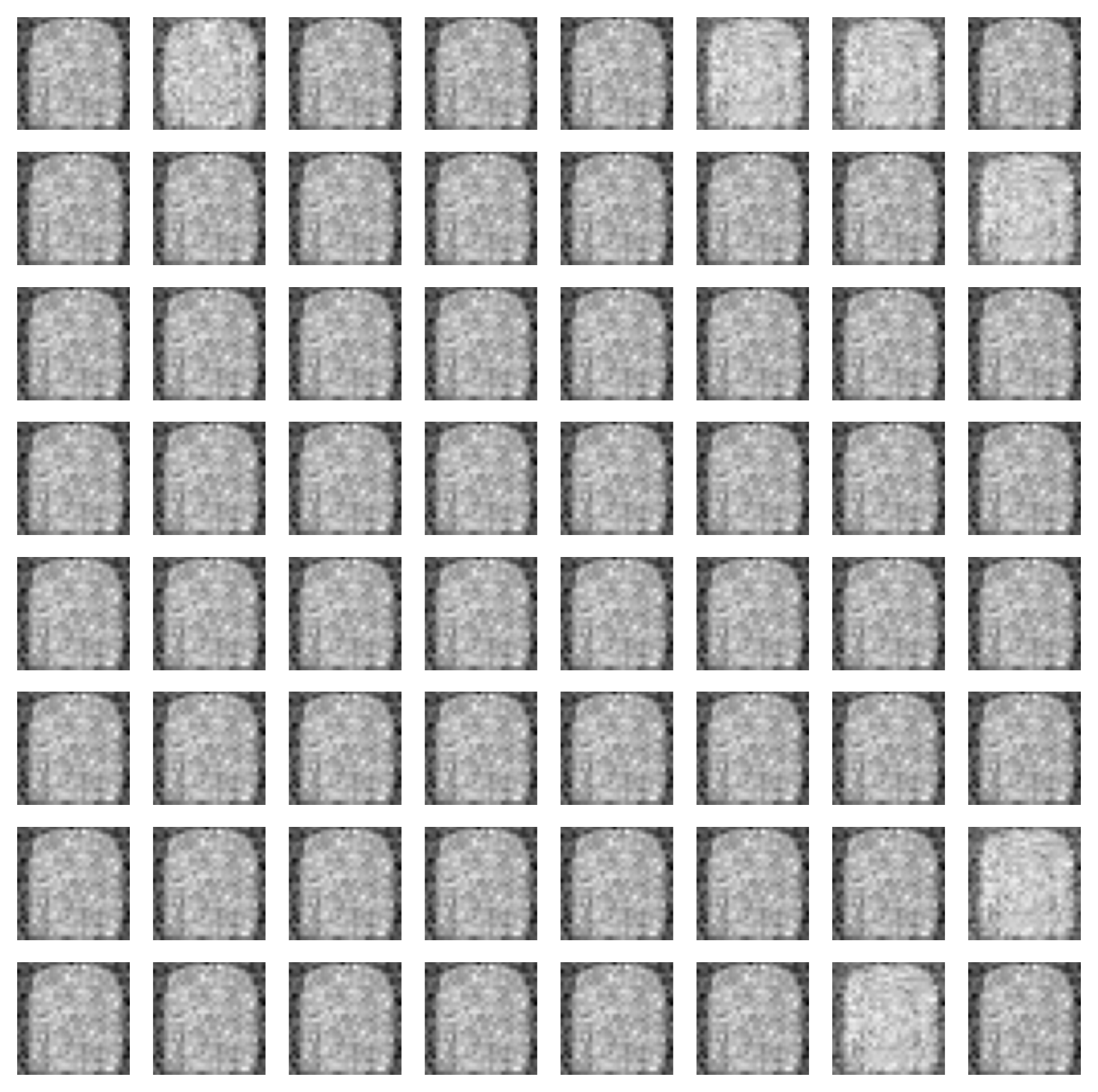} &   
  \includegraphics[width=0.18\textwidth]{figures/illustrations/fashion/fab_fashionnice.pdf} &  
  \includegraphics[width=0.18\textwidth]{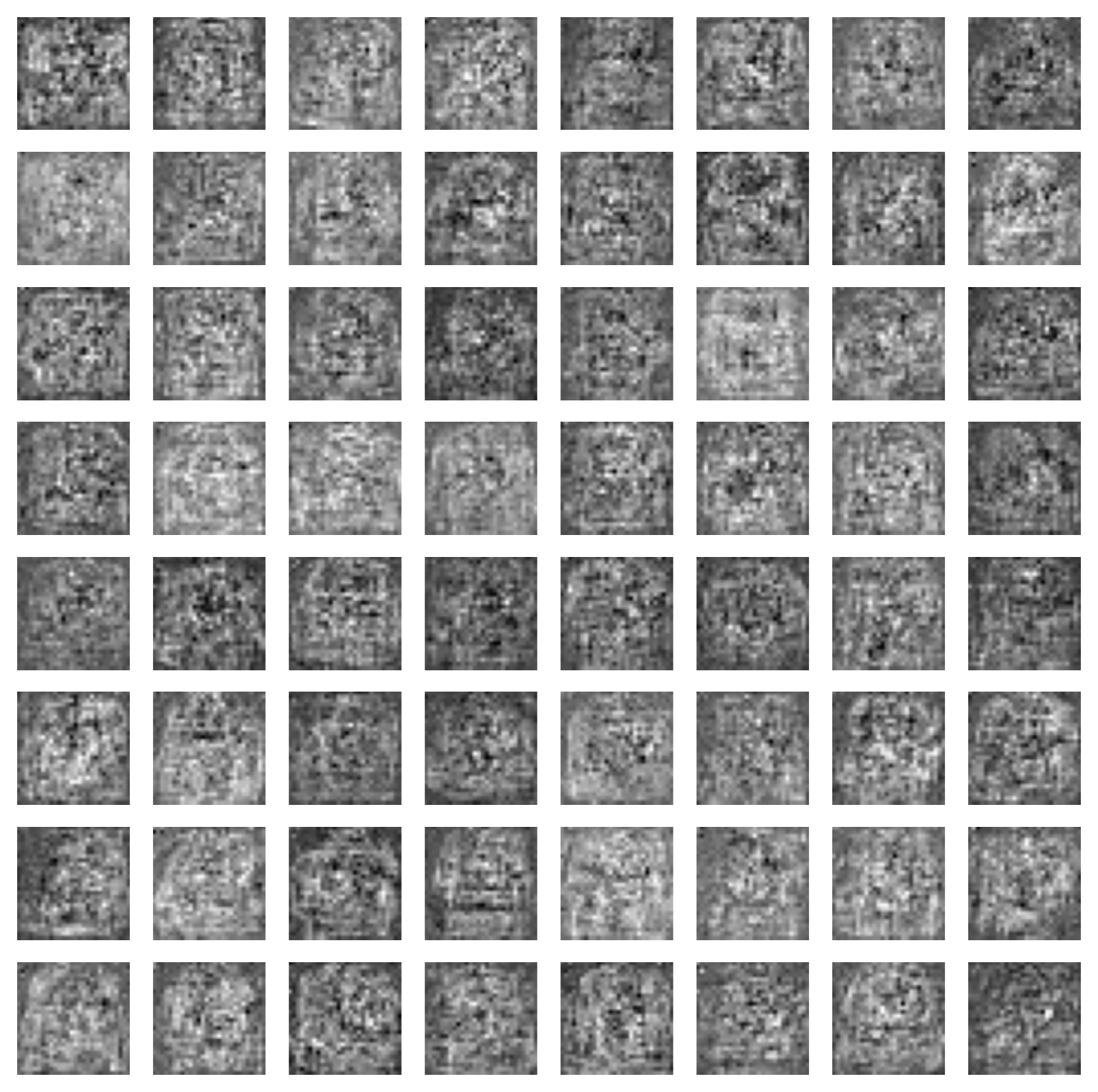} &   
  \includegraphics[width=0.18\textwidth]{figures/illustrations/fashion/ldvi_fashionnice.pdf} &   
  \\
  \small CRAFT &
  \small FAB &
  \small MCD &
  \small LDVI &
  \\
  \includegraphics[width=0.18\textwidth]{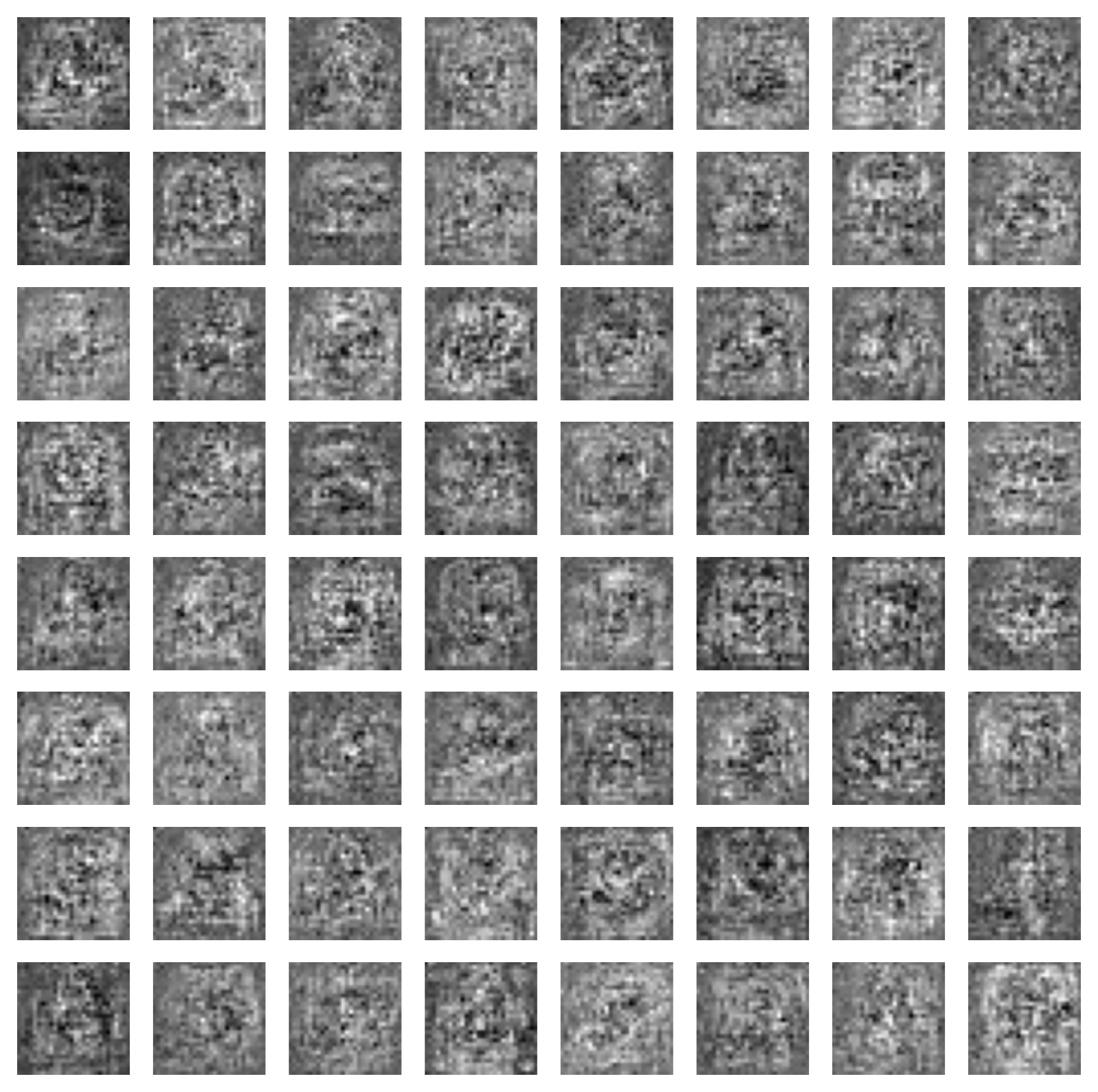} &   
  \includegraphics[width=0.18\textwidth]{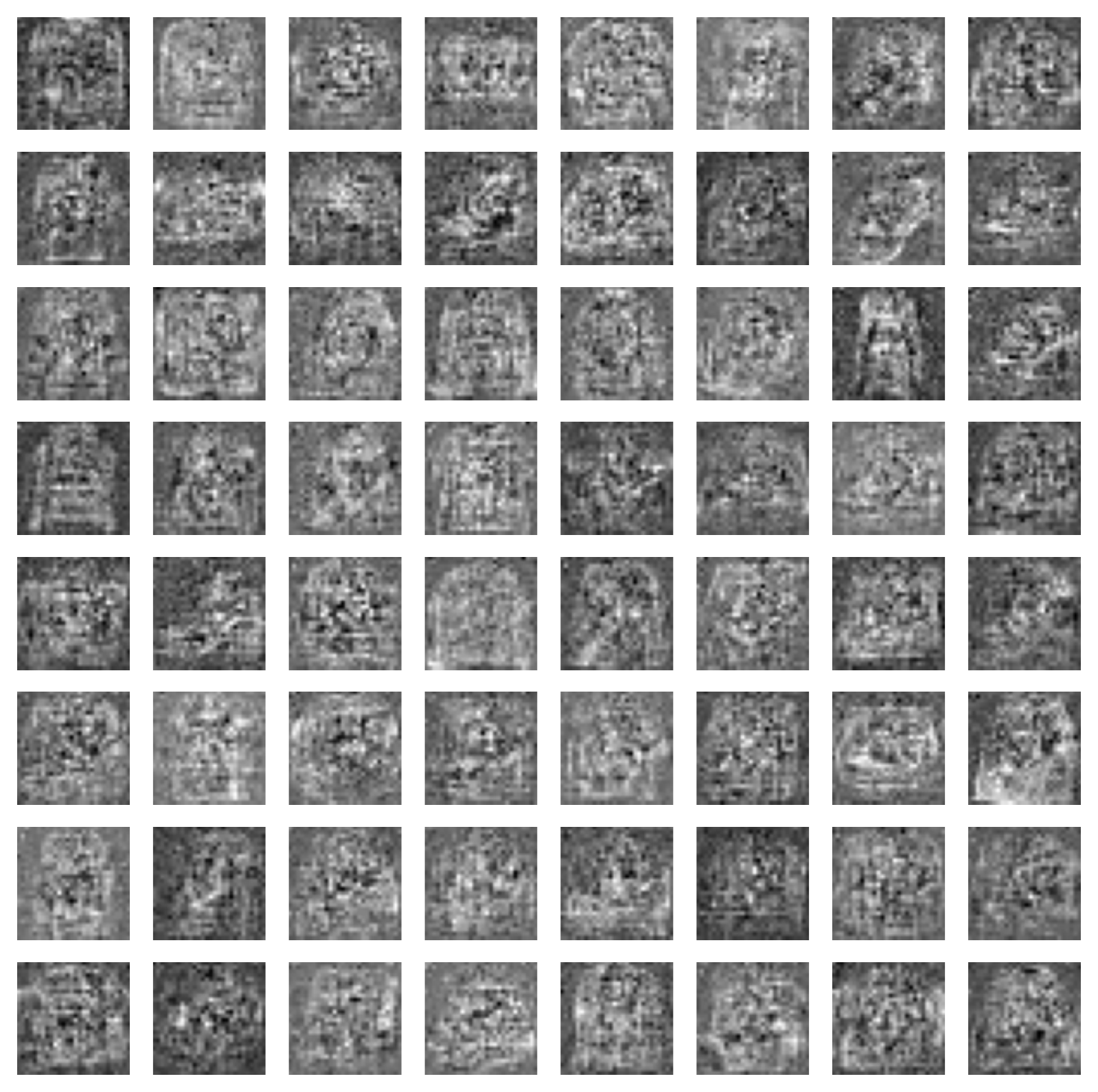} &   
  \includegraphics[width=0.18\textwidth]{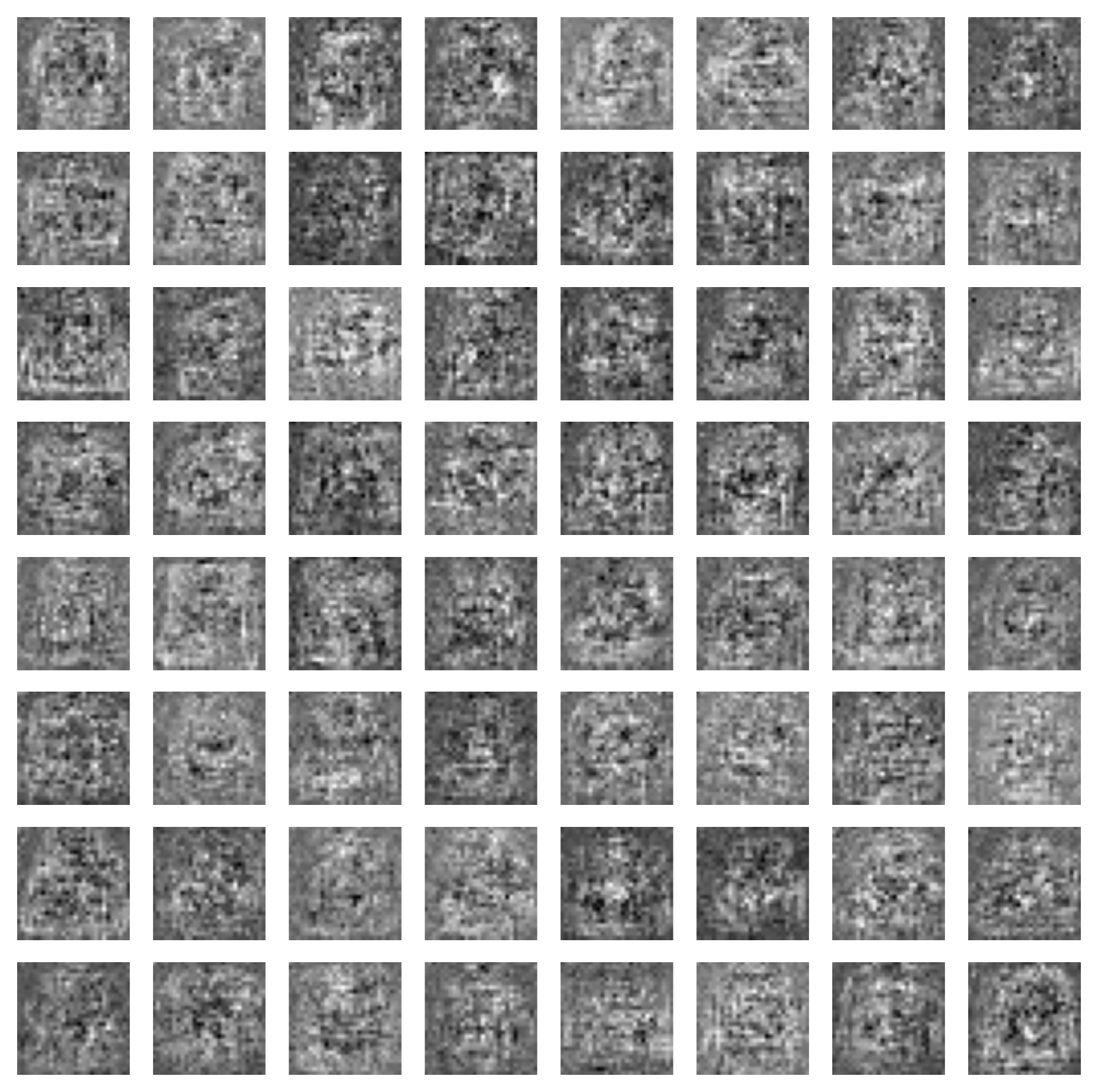} &   
  \includegraphics[width=0.18\textwidth]{figures/illustrations/fashion/gsb_fashionnice.pdf} &   
  \\
  \small PIS &
  \small DIS &  
  \small DDS &  
  \small GBS &  
  \\
  \bottomrule
\end{tabular}
\caption{Visualization of samples drawn from different sampling methods for Digits (top) and Fashion (bottom).}
\label{fig:nice}
\end{figure*}

We additionally provide sample visualizations for \textit{Funnel} and  \textit{MoG} in Figure \ref{fig:samples_mog_funnel}, and \textit{Digits} and \textit{Fashion} in Figure \ref{fig:nice}. We also report additional evaluation criteria for \textit{MoG} and \textit{MoS}, including 2-Wasserstein distance, maximum mean discrepancy, reverse and forward partition function error, lower and upper evidence bounds, and reverse and forward effective sample size in Table \ref{tab:dim}. Lastly, we provide insights into the models efficiency by providing values for the number of target queries and wallclock time needed, for obtaining the best ELBO value. These results are shown in Table \ref{tab:efficiency}.

\begin{table*}[t!]
\centering
\resizebox{\textwidth}{!}{%
\begin{tabular}{l|rrr|rrr|rrr|rrr}
\toprule
   & \multicolumn{6}{c}{\textbf{MoG}} & \multicolumn{6}{c}{\textbf{MoS}}   \\ 
   \\ 
& $d=2$ & $d=50$  & $d=200$ & $d=2$ & $d=50$  & $d=200$
& $d=2$ & $d=50$  & $d=200$ & $d=2$ & $d=50$  & $d=200$
\\ 
\midrule
   & \multicolumn{3}{c}{ $\mathcal{W}_2 \downarrow$} & \multicolumn{3}{c}{MMD $\downarrow$}   
   & \multicolumn{3}{c}{ $\mathcal{W}_2 \downarrow$} & \multicolumn{3}{c}{MMD $\downarrow$}
   \\ 
\midrule
MFVI
& $506.967 \scriptstyle \pm 7.385$
& $36158.898 \scriptstyle \pm 8.765$
& $148945.539 \scriptstyle \pm 14.42$
& $0.251 \scriptstyle \pm 0.002$
& $0.209 \scriptstyle \pm 0.000$
& $0.211 \scriptstyle \pm 0.000$
& $24.688 \scriptstyle \pm 0.225$
& $2282.540 \scriptstyle \pm 1.959$
& $12956.415 \scriptstyle \pm 6.530$
& $0.162 \scriptstyle \pm 0.001$
& $0.187 \scriptstyle \pm 0.001$
& $0.195 \scriptstyle \pm 0.000$
\\
GMMVI
& $76.474 \scriptstyle \pm 20.60$
& $31983.344 \scriptstyle \pm 1065.$
& $140166.746 \scriptstyle \pm 3020.$
& $0.052 \scriptstyle \pm 0.010$
& $0.202 \scriptstyle \pm 0.013$
& $0.214 \scriptstyle \pm 0.012$
& $2.851 \scriptstyle \pm 0.128$
& $1249.010 \scriptstyle \pm 297.3$
& $10402.243 \scriptstyle \pm 870.9$
& $0.036 \scriptstyle \pm 0.000$
& $0.133 \scriptstyle \pm 0.018$
& $0.211 \scriptstyle \pm 0.026$
\\
SMC
& $32.387 \scriptstyle \pm 9.219$
& $46351.236 \scriptstyle \pm 4.795$
& $176586.789 \scriptstyle \pm 3.638$
& $0.047 \scriptstyle \pm 0.004$
& $0.631 \scriptstyle \pm 0.000$
& $0.611 \scriptstyle \pm 0.000$
& $34.963 \scriptstyle \pm 2.833$
& $3297.640 \scriptstyle \pm 1372.$
& $17612.889 \scriptstyle \pm 2423.$
& $0.069 \scriptstyle \pm 0.003$
& $0.431 \scriptstyle \pm 0.161$
& $0.509 \scriptstyle \pm 0.113$
\\
AFT
& $21.571 \scriptstyle \pm 6.374$
& $44914.194 \scriptstyle \pm 1154.$
& $184075.172 \scriptstyle \pm 4347.$
& $0.040 \scriptstyle \pm 0.003$
& $0.622 \scriptstyle \pm 0.009$
& $0.622 \scriptstyle \pm 0.008$
& $41.299 \scriptstyle \pm 11.27$
& $2648.410 \scriptstyle \pm 301.3$
& $20207.756 \scriptstyle \pm 998.6$
& $0.077 \scriptstyle \pm 0.011$
& $0.395 \scriptstyle \pm 0.082$
& $0.611 \scriptstyle \pm 0.019$
\\
CRAFT
& $24.554 \scriptstyle \pm 4.216$
& $42953.544 \scriptstyle \pm 389.9$
& $177039.500 \scriptstyle \pm 329.3$
& $0.041 \scriptstyle \pm 0.003$
& $0.600 \scriptstyle \pm 0.003$
& $0.609 \scriptstyle \pm 0.002$
& $10.108 \scriptstyle \pm 0.186$
& $1806.321 \scriptstyle \pm 117.4$
& $14411.712 \scriptstyle \pm 305.9$
& $0.048 \scriptstyle \pm 0.000$
& $0.233 \scriptstyle \pm 0.021$
& $0.425 \scriptstyle \pm 0.024$
\\
FAB
& $57.111 \scriptstyle \pm 24.53$
& $9567.319 \scriptstyle \pm 626.1$
& $58832.370 \scriptstyle \pm 1092.$
& $0.047 \scriptstyle \pm 0.007$
& $0.073 \scriptstyle \pm 0.005$
& $0.099 \scriptstyle \pm 0.001$
& $8.868 \scriptstyle \pm 1.673$
& $\mathbf{1193.455 \scriptstyle \pm 152.3}$
& $\mathbf{7490.803 \scriptstyle \pm 433.9}$
& $\mathbf{0.035 \scriptstyle \pm 0.003}$
& $\mathbf{0.093 \scriptstyle \pm 0.014}$
& $\mathbf{0.102 \scriptstyle \pm 0.012}$
\\
MCD
& $211.657 \scriptstyle \pm 3.504$
& $4892.591 \scriptstyle \pm 71.26$
& $\mathbf{30977.775 \scriptstyle \pm 276.6}$
& $0.136 \scriptstyle \pm 0.001$
& $0.043 \scriptstyle \pm 0.000$
& $\mathbf{0.054 \scriptstyle \pm 0.000}$
& $102.002 \scriptstyle \pm 0.338$
& $6406.902 \scriptstyle \pm 20.87$
& $32034.058 \scriptstyle \pm 40.86$
& $0.215 \scriptstyle \pm 0.001$
& $0.256 \scriptstyle \pm 0.000$
& $0.257 \scriptstyle \pm 0.000$
\\
LDVI
& $178.241 \scriptstyle \pm 3.129$
& $4931.898 \scriptstyle \pm 87.43$
& $31019.831 \scriptstyle \pm 278.6$
& $0.118 \scriptstyle \pm 0.003$
& $0.043 \scriptstyle \pm 0.000$
& $\mathbf{0.054 \scriptstyle \pm 0.000}$
& $38.758 \scriptstyle \pm 4.940$
& $2899.472 \scriptstyle \pm 102.9$
& $17435.914 \scriptstyle \pm 299.8$
& $0.084 \scriptstyle \pm 0.008$
& $0.181 \scriptstyle \pm 0.003$
& $0.183 \scriptstyle \pm 0.002$
\\
PIS
& $\mathbf{10.398 \scriptstyle \pm 1.599}$
& $10405.749 \scriptstyle \pm 69.41$
& $92623.455 \scriptstyle \pm 1219.$
& $\mathbf{0.031 \scriptstyle \pm 0.001}$
& $0.082 \scriptstyle \pm 0.000$
& $0.168 \scriptstyle \pm 0.003$
& $\mathbf{2.476 \scriptstyle \pm 0.236}$
& $2078.751 \scriptstyle \pm 41.51$
& $32415.244 \scriptstyle \pm 63.11$
& $\mathbf{0.033 \scriptstyle \pm 0.001}$
& $0.205 \scriptstyle \pm 0.008$
& $0.258 \scriptstyle \pm 0.001$
\\
DIS
& $65.162 \scriptstyle \pm 35.72$
& $\mathbf{3044.733 \scriptstyle \pm 464.7}$
& $31573.015 \scriptstyle \pm 702.4$
& $0.071 \scriptstyle \pm 0.017$
& $\mathbf{0.034 \scriptstyle \pm 0.003}$
& $\mathbf{0.055 \scriptstyle \pm 0.001}$
& $3.486 \scriptstyle \pm 0.214$
& $2200.590 \scriptstyle \pm 18.73$
& $13059.766 \scriptstyle \pm 72.12$
& $0.037 \scriptstyle \pm 0.002$
& $0.155 \scriptstyle \pm 0.001$
& $0.152 \scriptstyle \pm 0.001$
\\
DDS
& $16.217 \scriptstyle \pm 3.202$
& $5435.177 \scriptstyle \pm 172.2$
& $38576.259 \scriptstyle \pm 392.9$
& $0.035 \scriptstyle \pm 0.002$
& $0.045 \scriptstyle \pm 0.001$
& $0.065 \scriptstyle \pm 0.001$
& $3.641 \scriptstyle \pm 0.224$
& $2145.188 \scriptstyle \pm 3.960$
& $24187.186 \scriptstyle \pm 256.4$
& $\mathbf{0.034 \scriptstyle \pm 0.003}$
& $0.124 \scriptstyle \pm 0.001$
& $0.219 \scriptstyle \pm 0.003$
\\
GBS
& $140.138 \scriptstyle \pm 39.76$
& $5027.819 \scriptstyle \pm 103.7$
& $31970.248 \scriptstyle \pm 1177.$
& $0.108 \scriptstyle \pm 0.021$
& $0.043 \scriptstyle \pm 0.000$
& $\mathbf{0.055 \scriptstyle \pm 0.001}$
& $2.572 \scriptstyle \pm 0.099$
& $5708.871 \scriptstyle \pm 20.91$
& $22914.911 \scriptstyle \pm 300.4$
& $\mathbf{0.034 \scriptstyle \pm 0.001}$
& $0.232 \scriptstyle \pm 0.000$
& $0.203 \scriptstyle \pm 0.001$
\\
\midrule
   & \multicolumn{3}{c}{$\mathbf{\Delta \log Z}_{r} \downarrow$} & \multicolumn{3}{c}{$\mathbf{\Delta \log Z}_{f} \downarrow$}   
   & \multicolumn{3}{c}{$\mathbf{\Delta \log Z}_{r} \downarrow$} & \multicolumn{3}{c}{$\mathbf{\Delta \log Z}_{f} \downarrow$}
   \\ 
\midrule
MFVI
& $0.084 \scriptstyle \pm 0.066$
& $3.658 \scriptstyle \pm 0.040$
& $3.676 \scriptstyle \pm 0.0130$
& $0.150 \scriptstyle \pm 0.002$
& $0.185 \scriptstyle \pm 0.002$
& $0.176 \scriptstyle \pm 0.0050$
& $0.018 \scriptstyle \pm 0.003$
& $3.009 \scriptstyle \pm 0.291$
& $8.048 \scriptstyle \pm 0.758$
& $0.114 \scriptstyle \pm 0.000$
& $\mathbf{0.048 \scriptstyle \pm 0.002}$
& $5.982 \scriptstyle \pm 0.019$
\\
GMMVI
& $0.044 \scriptstyle \pm 0.011$
& $\mathbf{1.715 \scriptstyle \pm 0.119}$
& $\mathbf{1.709 \scriptstyle \pm 0.0580}$
& $\mathbf{0.003 \scriptstyle \pm 0.002}$
& $\mathbf{0.048 \scriptstyle \pm 0.007}$
& $\mathbf{0.028 \scriptstyle \pm 0.0270}$
& $\mathbf{0.000 \scriptstyle \pm 0.000}$
& $1.282 \scriptstyle \pm 0.221$
& $\mathbf{7.126 \scriptstyle \pm 0.377}$
& $\mathbf{0.000 \scriptstyle \pm 0.000}$
& $0.084 \scriptstyle \pm 0.055$
& $\mathbf{5.708 \scriptstyle \pm 0.478}$
\\
SMC
& $0.069 \scriptstyle \pm 0.010$
& $690.721 \scriptstyle \pm 11.21$
& $6326.621 \scriptstyle \pm 51.428$
& $2.728 \scriptstyle \pm 0.000$
& $161.796 \scriptstyle \pm 0.000$
& $661.945 \scriptstyle \pm 0.0000$
& $0.016 \scriptstyle \pm 0.009$
& $3.880 \scriptstyle \pm 1.105$
& $49.846 \scriptstyle \pm 7.638$
& $1.262 \scriptstyle \pm 0.000$
& $80.992 \scriptstyle \pm 0.000$
& $338.745 \scriptstyle \pm 0.000$
\\
AFT
& $0.023 \scriptstyle \pm 0.015$
& $765.624 \scriptstyle \pm 108.0$
& $5567.272 \scriptstyle \pm 277.52$
& $1.157 \scriptstyle \pm 0.038$
& $110.955 \scriptstyle \pm 18.37$
& $420.932 \scriptstyle \pm 12.987$
& $0.024 \scriptstyle \pm 0.014$
& $4.081 \scriptstyle \pm 1.579$
& $47.121 \scriptstyle \pm 6.693$
& $0.639 \scriptstyle \pm 0.095$
& $205.297 \scriptstyle \pm 23.91$
& $12765.117 \scriptstyle \pm 2877.$
\\
CRAFT
& $0.008 \scriptstyle \pm 0.001$
& $337.094 \scriptstyle \pm 9.296$
& $2504.363 \scriptstyle \pm 64.970$
& $0.901 \scriptstyle \pm 0.007$
& $100.987 \scriptstyle \pm 0.065$
& $415.277 \scriptstyle \pm 0.4550$
& $0.004 \scriptstyle \pm 0.001$
& $\mathbf{0.822 \scriptstyle \pm 0.087}$
& $19.738 \scriptstyle \pm 0.342$
& $0.333 \scriptstyle \pm 0.013$
& $210.245 \scriptstyle \pm 6.098$
& $12516.502 \scriptstyle \pm 631.8$
\\
FAB
& $\mathbf{0.007 \scriptstyle \pm 0.003}$
& $2.952 \scriptstyle \pm 0.247$
& $3.331 \scriptstyle \pm 0.2290$
& $1.193 \scriptstyle \pm 0.125$
& $126.363 \scriptstyle \pm 1.789$
& $545.226 \scriptstyle \pm 5.6200$
& $0.005 \scriptstyle \pm 0.001$
& $3.358 \scriptstyle \pm 1.062$
& $43.419 \scriptstyle \pm 4.690$
& $0.268 \scriptstyle \pm 0.093$
& $84.592 \scriptstyle \pm 22.64$
& $13514.417 \scriptstyle \pm 101.9$
\\
MCD
& $0.010 \scriptstyle \pm 0.002$
& $31.319 \scriptstyle \pm 1.793$
& $2354.020 \scriptstyle \pm 60.855$
& $0.009 \scriptstyle \pm 0.005$
& $21.148 \scriptstyle \pm 1.478$
& $305.656 \scriptstyle \pm 2.6620$
& $0.010 \scriptstyle \pm 0.002$
& $28.607 \scriptstyle \pm 1.275$
& $210.536 \scriptstyle \pm 1.393$
& $0.068 \scriptstyle \pm 0.010$
& $24.757 \scriptstyle \pm 0.841$
& $147.321 \scriptstyle \pm 1.272$
\\
LDVI
& $0.038 \scriptstyle \pm 0.015$
& $8.159 \scriptstyle \pm 0.775$
& $647.953 \scriptstyle \pm 7.2120$
& $0.031 \scriptstyle \pm 0.008$
& $15.477 \scriptstyle \pm 0.815$
& $282.699 \scriptstyle \pm 4.1050$
& $0.004 \scriptstyle \pm 0.001$
& $4.360 \scriptstyle \pm 0.741$
& $103.224 \scriptstyle \pm 2.118$
& $0.017 \scriptstyle \pm 0.003$
& $5.472 \scriptstyle \pm 0.938$
& $83.029 \scriptstyle \pm 0.819$
\\
CMCD
& $0.026 \scriptstyle \pm 0.011$
& $51.218 \scriptstyle \pm 2.809$
& $306.127 \scriptstyle \pm 24.673$
& $0.030 \scriptstyle \pm 0.008$
& $79.227 \scriptstyle \pm 3.758$
& $440.341 \scriptstyle \pm 4.2520$
& $0.004 \scriptstyle \pm 0.001$
& $10.533 \scriptstyle \pm 0.404$
& $167.654 \scriptstyle \pm 1.564$
& $0.005 \scriptstyle \pm 0.002$
& $12.835 \scriptstyle \pm 0.275$
& $148.676 \scriptstyle \pm 2.851$
\\
PIS
& $0.267 \scriptstyle \pm 0.006$
& $7.122 \scriptstyle \pm 0.630$
& $40.699 \scriptstyle \pm 0.5430$
& $0.094 \scriptstyle \pm 0.038$
& $3113.492 \scriptstyle \pm 1.978$
& $16071.743 \scriptstyle \pm 3.0460$
& $0.275 \scriptstyle \pm 0.016$
& $12.248 \scriptstyle \pm 0.326$
& $209.981 \scriptstyle \pm 2.573$
& $0.342 \scriptstyle \pm 0.001$
& $54.090 \scriptstyle \pm 0.151$
& $304.178 \scriptstyle \pm 0.329$
\\
DIS
& $0.058 \scriptstyle \pm 0.030$
& $87.709 \scriptstyle \pm 8.942$
& $11646.394 \scriptstyle \pm 15938.$
& $1.390 \scriptstyle \pm 0.458$
& $369.352 \scriptstyle \pm 16.29$
& $14376.906 \scriptstyle \pm 17877.$
& $0.049 \scriptstyle \pm 0.005$
& $10.448 \scriptstyle \pm 0.607$
& $658.634 \scriptstyle \pm 4.952$
& $3.212 \scriptstyle \pm 0.028$
& $87.897 \scriptstyle \pm 5.255$
& $433.741 \scriptstyle \pm 10.78$
\\
DDS
& $0.012 \scriptstyle \pm 0.005$
& $1.739 \scriptstyle \pm 0.442$
& $27.506 \scriptstyle \pm 2.5840$
& $1.698 \scriptstyle \pm 0.029$
& $207.545 \scriptstyle \pm 1.163$
& $1052.805 \scriptstyle \pm 1.7320$
& $0.005 \scriptstyle \pm 0.001$
& $7.952 \scriptstyle \pm 0.299$
& $155.502 \scriptstyle \pm 3.594$
& $0.315 \scriptstyle \pm 0.000$
& $53.411 \scriptstyle \pm 0.024$
& $291.566 \scriptstyle \pm 0.102$
\\
GBS
& $\mathbf{0.007 \scriptstyle \pm 0.001}$
& $8.103 \scriptstyle \pm 1.696$
& $87.971 \scriptstyle \pm 14.656$
& $0.008 \scriptstyle \pm 0.001$
& $9.321 \scriptstyle \pm 0.776$
& $72.634 \scriptstyle \pm 12.301$
& $0.002 \scriptstyle \pm 0.000$
& $53.767 \scriptstyle \pm 0.732$
& $157.791 \scriptstyle \pm 2.947$
& $0.010 \scriptstyle \pm 0.002$
& $47.441 \scriptstyle \pm 0.098$
& $101.874 \scriptstyle \pm 2.214$
\\
\midrule
   & \multicolumn{3}{c}{\textbf{ELBO} $\uparrow$} & \multicolumn{3}{c}{\textbf{EUBO} $\downarrow$}   
   & \multicolumn{3}{c}{\textbf{ELBO} $\uparrow$} & \multicolumn{3}{c}{\textbf{EUBO} $\downarrow$}
   \\ 
\midrule
MFVI
& $-3.011 \scriptstyle \pm 0.002$
& $-3.690 \scriptstyle \pm 0.000$
& $-3.695 \scriptstyle \pm 0.0010$
& $3.089 \scriptstyle \pm 0.000$
& $164.114 \scriptstyle \pm 0.000$
& $666.954 \scriptstyle \pm 0.0000$
& $-1.038 \scriptstyle \pm 0.007$
& $-5.957 \scriptstyle \pm 0.007$
& $-16.969 \scriptstyle \pm 0.011$
& $1.218 \scriptstyle \pm 0.001$
& $72.663 \scriptstyle \pm 0.005$
& $324.202 \scriptstyle \pm 0.044$
\\
GMMVI
& $\mathbf{-0.045 \scriptstyle \pm 0.011}$
& $\mathbf{-1.715 \scriptstyle \pm 0.119}$
& $\mathbf{-1.709 \scriptstyle \pm 0.0580}$
& $3.619 \scriptstyle \pm 1.308$
& $240.459 \scriptstyle \pm 51.13$
& $645.405 \scriptstyle \pm 6.3090$
& $\mathbf{-0.001 \scriptstyle \pm 0.000}$
& $-3.890 \scriptstyle \pm 0.122$
& $\mathbf{-15.649 \scriptstyle \pm 0.173}$
& $\mathbf{0.002 \scriptstyle \pm 0.001}$
& $57.746 \scriptstyle \pm 1.928$
& $268.513 \scriptstyle \pm 17.65$
\\
SMC
& $-2.095 \scriptstyle \pm 0.009$
& $-877.034 \scriptstyle \pm 10.23$
& $-6816.697 \scriptstyle \pm 44.195$
& $2.734 \scriptstyle \pm 0.000$
& $161.921 \scriptstyle \pm 0.000$
& $662.404 \scriptstyle \pm 0.0000$
& $-0.010 \scriptstyle \pm 0.016$
& $-4.634 \scriptstyle \pm 1.088$
& $-52.535 \scriptstyle \pm 7.564$
& $1.272 \scriptstyle \pm 0.000$
& $81.325 \scriptstyle \pm 0.000$
& $340.984 \scriptstyle \pm 0.000$
\\
AFT
& $-1.778 \scriptstyle \pm 0.090$
& $-927.16 \scriptstyle \pm 103.8$
& $-6053.823 \scriptstyle \pm 260.99$
& $1.248 \scriptstyle \pm 0.045$
& $117.63 \scriptstyle \pm 22.16$
& $439.434 \scriptstyle \pm 16.788$
& $-0.041 \scriptstyle \pm 0.031$
& $-4.923 \scriptstyle \pm 1.546$
& $-50.328 \scriptstyle \pm 6.627$
& $0.67 \scriptstyle \pm 0.100$
& $207.625 \scriptstyle \pm 24.14$
& $12801.561 \scriptstyle \pm 2892.$
\\
CRAFT
& $-0.666 \scriptstyle \pm 0.026$
& $-451.399 \scriptstyle \pm 7.561$
& $-2836.471 \scriptstyle \pm 57.695$
& $0.976 \scriptstyle \pm 0.007$
& $103.674 \scriptstyle \pm 0.069$
& $425.500 \scriptstyle \pm 0.7070$
& $\mathbf{-0.002 \scriptstyle \pm 0.002}$
& $\mathbf{-0.339 \scriptstyle \pm 0.180}$
& $-22.687 \scriptstyle \pm 0.358$
& $0.346 \scriptstyle \pm 0.013$
& $212.210 \scriptstyle \pm 6.160$
& $12553.883 \scriptstyle \pm 645.7$
\\
FAB
& $-19.932 \scriptstyle \pm 12.80$
& $-299.916 \scriptstyle \pm 253.4$
& $-63.212 \scriptstyle \pm 56.191$
& $0.865 \scriptstyle \pm 0.113$
& $93.560 \scriptstyle \pm 5.086$
& $386.884 \scriptstyle \pm 12.161$
& $-0.257 \scriptstyle \pm 0.075$
& $75.735 \scriptstyle \pm 175.8$
& $-98.558 \scriptstyle \pm 7.688$
& $0.162 \scriptstyle \pm 0.055$
& $\mathbf{18.088 \scriptstyle \pm 2.503}$
& $227.514 \scriptstyle \pm 0.320$
\\
MCD
& $-0.651 \scriptstyle \pm 0.014$
& $-185.021 \scriptstyle \pm 0.743$
& $-4017.832 \scriptstyle \pm 20.356$
& $0.652 \scriptstyle \pm 0.008$
& $\mathbf{43.670 \scriptstyle \pm 0.457}$
& $358.687 \scriptstyle \pm 2.1120$
& $-1.215 \scriptstyle \pm 0.005$
& $-69.358 \scriptstyle \pm 0.633$
& $-308.728 \scriptstyle \pm 0.450$
& $0.734 \scriptstyle \pm 0.002$
& $47.834 \scriptstyle \pm 0.820$
& $208.626 \scriptstyle \pm 0.525$
\\
LDVI
& $-0.986 \scriptstyle \pm 0.136$
& $-29.034 \scriptstyle \pm 0.591$
& $-956.576 \scriptstyle \pm 6.2700$
& $1.072 \scriptstyle \pm 0.242$
& $51.137 \scriptstyle \pm 0.177$
& $375.527 \scriptstyle \pm 3.1100$
& $-0.311 \scriptstyle \pm 0.034$
& $-28.471 \scriptstyle \pm 1.018$
& $-173.716 \scriptstyle \pm 2.629$
& $0.198 \scriptstyle \pm 0.008$
& $20.887 \scriptstyle \pm 1.042$
& $\mathbf{132.711 \scriptstyle \pm 1.817}$
\\
PIS
& $-0.585 \scriptstyle \pm 0.016$
& $-16.881 \scriptstyle \pm 0.026$
& $-65.700 \scriptstyle \pm 0.2010$
& $7.344 \scriptstyle \pm 0.004$
& $3626.120 \scriptstyle \pm 1.360$
& $16979.347 \scriptstyle \pm 4.4700$
& $-0.387 \scriptstyle \pm 0.009$
& $-29.261 \scriptstyle \pm 1.743$
& $-306.678 \scriptstyle \pm 0.548$
& $1.868 \scriptstyle \pm 0.000$
& $88.192 \scriptstyle \pm 0.005$
& $363.435 \scriptstyle \pm 0.030$
\\
DIS
& $-1.850 \scriptstyle \pm 0.359$
& $-181.348 \scriptstyle \pm 15.47$
& $-14142.693 \scriptstyle \pm 17807.$
& $6.653 \scriptstyle \pm 0.357$
& $546.335 \scriptstyle \pm 30.86$
& $15792.004 \scriptstyle \pm 18966.$
& $-0.157 \scriptstyle \pm 0.023$
& $-36.704 \scriptstyle \pm 0.629$
& $-819.959 \scriptstyle \pm 6.264$
& $4.778 \scriptstyle \pm 0.038$
& $193.270 \scriptstyle \pm 3.293$
& $658.575 \scriptstyle \pm 7.820$
\\
DDS
& $-0.527 \scriptstyle \pm 0.022$
& $-13.284 \scriptstyle \pm 0.460$
& $-60.642 \scriptstyle \pm 2.3330$
& $4.176 \scriptstyle \pm 0.000$
& $291.867 \scriptstyle \pm 0.047$
& $1224.926 \scriptstyle \pm 2.4850$
& $-0.110 \scriptstyle \pm 0.007$
& $-31.681 \scriptstyle \pm 0.363$
& $-244.188 \scriptstyle \pm 3.504$
& $1.783 \scriptstyle \pm 0.000$
& $86.014 \scriptstyle \pm 0.001$
& $351.204 \scriptstyle \pm 0.005$
\\
GBS
& $-0.473 \scriptstyle \pm 0.061$
& $-35.771 \scriptstyle \pm 1.105$
& $-161.259 \scriptstyle \pm 20.704$
& $\mathbf{0.485 \scriptstyle \pm 0.047}$
& $67.819 \scriptstyle \pm 2.157$
& $\mathbf{204.498 \scriptstyle \pm 48.539}$
& $-0.064 \scriptstyle \pm 0.004$
& $-99.369 \scriptstyle \pm 0.158$
& $-258.263 \scriptstyle \pm 2.639$
& $0.064 \scriptstyle \pm 0.004$
& $73.545 \scriptstyle \pm 0.107$
& $147.412 \scriptstyle \pm 1.504$
\\
\midrule
   & \multicolumn{3}{c}{$\textbf{ESS}_r$  $\uparrow$} & \multicolumn{3}{c}{$\textbf{ESS}_f$ $\uparrow$}   
   & \multicolumn{3}{c}{$\textbf{ESS}_r$  $\uparrow$} & \multicolumn{3}{c}{$\textbf{ESS}_f$ $\uparrow$}
   \\ 
\midrule
MFVI
& $0.077 \scriptstyle \pm 0.016$
& $0.997 \scriptstyle \pm 0.000$
& $0.988 \scriptstyle \pm 0.001$
& $0.286 \scriptstyle \pm 0.000$
& $0.000 \scriptstyle \pm 0.000$
& $0.000 \scriptstyle \pm 0.000$
& $0.180 \scriptstyle \pm 0.007$
& $\mathbf{0.031 \scriptstyle \pm 0.007}$
& $\mathbf{0.006 \scriptstyle \pm 0.001}$
& $0.163 \scriptstyle \pm 0.000$
& $0.000 \scriptstyle \pm 0.000$
& $0.000 \scriptstyle \pm 0.000$
\\
GMMVI
& $\mathbf{1.000 \scriptstyle \pm 0.000}$
& $\mathbf{1.000 \scriptstyle \pm 0.000}$
& $\mathbf{1.000 \scriptstyle \pm 0.000}$
& $0.000 \scriptstyle \pm 0.000$
& $0.000 \scriptstyle \pm 0.000$
& $0.000 \scriptstyle \pm 0.000$
& $\mathbf{0.997 \scriptstyle \pm 0.000}$
& $0.027 \scriptstyle \pm 0.004$
& $\mathbf{0.006 \scriptstyle \pm 0.001}$
& $\mathbf{0.997 \scriptstyle \pm 0.000}$
& $0.000 \scriptstyle \pm 0.000$
& $0.000 \scriptstyle \pm 0.000$
\\
MCD
& $0.311 \scriptstyle \pm 0.013$
& $0.001 \scriptstyle \pm 0.000$
& $0.000 \scriptstyle \pm 0.000$
& $0.289 \scriptstyle \pm 0.010$
& $0.000 \scriptstyle \pm 0.000$
& $0.000 \scriptstyle \pm 0.000$
& $0.332 \scriptstyle \pm 0.004$
& $0.001 \scriptstyle \pm 0.000$
& $0.000 \scriptstyle \pm 0.000$
& $0.352 \scriptstyle \pm 0.004$
& $0.000 \scriptstyle \pm 0.000$
& $0.000 \scriptstyle \pm 0.000$
\\
LDVI
& $0.207 \scriptstyle \pm 0.044$
& $0.002 \scriptstyle \pm 0.000$
& $0.000 \scriptstyle \pm 0.000$
& $0.269 \scriptstyle \pm 0.046$
& $0.000 \scriptstyle \pm 0.000$
& $0.000 \scriptstyle \pm 0.000$
& $0.742 \scriptstyle \pm 0.006$
& $0.002 \scriptstyle \pm 0.000$
& $0.000 \scriptstyle \pm 0.000$
& $0.761 \scriptstyle \pm 0.004$
& $0.000 \scriptstyle \pm 0.000$
& $0.000 \scriptstyle \pm 0.000$
\\
PIS
& $0.529 \scriptstyle \pm 0.012$
& $0.006 \scriptstyle \pm 0.001$
& $0.002 \scriptstyle \pm 0.000$
& $0.000 \scriptstyle \pm 0.000$
& $0.000 \scriptstyle \pm 0.000$
& $0.000 \scriptstyle \pm 0.000$
& $0.840 \scriptstyle \pm 0.004$
& $0.003 \scriptstyle \pm 0.000$
& $0.000 \scriptstyle \pm 0.000$
& $0.042 \scriptstyle \pm 0.000$
& $0.000 \scriptstyle \pm 0.000$
& $0.000 \scriptstyle \pm 0.000$
\\
DIS
& $0.078 \scriptstyle \pm 0.025$
& $0.000 \scriptstyle \pm 0.000$
& $0.000 \scriptstyle \pm 0.000$
& $0.000 \scriptstyle \pm 0.000$
& $0.000 \scriptstyle \pm 0.000$
& $0.000 \scriptstyle \pm 0.000$
& $0.580 \scriptstyle \pm 0.003$
& $0.002 \scriptstyle \pm 0.000$
& $0.000 \scriptstyle \pm 0.000$
& $0.000 \scriptstyle \pm 0.000$
& $0.000 \scriptstyle \pm 0.000$
& $0.000 \scriptstyle \pm 0.000$
\\
DDS
& $0.338 \scriptstyle \pm 0.003$
& $0.003 \scriptstyle \pm 0.000$
& $0.000 \scriptstyle \pm 0.000$
& $0.000 \scriptstyle \pm 0.000$
& $0.000 \scriptstyle \pm 0.000$
& $0.000 \scriptstyle \pm 0.000$
& $0.780 \scriptstyle \pm 0.010$
& $0.002 \scriptstyle \pm 0.000$
& $0.000 \scriptstyle \pm 0.000$
& $0.032 \scriptstyle \pm 0.000$
& $0.000 \scriptstyle \pm 0.000$
& $0.000 \scriptstyle \pm 0.000$
\\
GBS
& $0.405 \scriptstyle \pm 0.029$
& $0.002 \scriptstyle \pm 0.000$
& $0.000 \scriptstyle \pm 0.000$
& $\mathbf{0.380 \scriptstyle \pm 0.027}$
& $0.000 \scriptstyle \pm 0.000$
& $0.000 \scriptstyle \pm 0.000$
& $0.879 \scriptstyle \pm 0.002$
& $0.001 \scriptstyle \pm 0.000$
& $0.000 \scriptstyle \pm 0.000$
& $0.721 \scriptstyle \pm 0.161$
& $0.000 \scriptstyle \pm 0.000$
& $0.000 \scriptstyle \pm 0.000$
\\
\bottomrule
\end{tabular}
}
\caption{
Results for various sampling methods for MoG and MoS with varying dimensions $d$. Evaluation criteria include 2-Wasserstein distance ($\mathcal{W}_2$), maximum mean discrepancy (MMD), reverse and forward partition function error ($\Delta \log Z_{r}$, $\Delta \log Z_{f}$), lower and upper evidence bounds (ELBO, EUBO), reverse and forward effective sample size ($\text{ESS}_{r}$, $\text{ESS}_{f}$). The best results are highlighted in bold. Arrows ($\uparrow$, $\downarrow$) indicate whether higher or lower values are preferable, respectively.
}
\label{tab:dim}
\end{table*}


\begin{table}[h!]
\label{table:efficieny}
\begin{center}
\begin{small}
\begin{sc} {
\resizebox{.3\textwidth}{!}{%
\begin{tabular}{l|ccc}
\toprule
   & \multicolumn{3}{c}{\textbf{NFE} $\downarrow$}  \\ 
\textbf{Method}   & $d=2$ & $d=50$  & $d=200$  \\ 
\midrule
MFVI
& $6.5 \times 10^{6}$ 
& $2.3 \times 10^{6}$ 
& $1.9 \times 10^{6}$ 
\\
GMMVI
& $1.4 \times 10^{5}$ 
& $5.9 \times 10^{5}$ 
& $7.9 \times 10^{5}$ 
\\
SMC
& $2.8 \times 10^{6}$ 
& $2.8 \times 10^{6}$ 
& $2.8 \times 10^{6}$ 
\\
AFT
& $2.0 \times 10^{5}$ 
& $2.0 \times 10^{5}$ 
& $2.0 \times 10^{5}$ 
\\
CRAFT
& $4.5 \times 10^{9}$ 
& $4.4 \times 10^{9}$ 
& $4.5 \times 10^{9}$ 
\\
FAB
& $1.5 \times 10^{7}$ 
& $3.4 \times 10^{7}$ 
& $3.4 \times 10^{7}$ 
\\
DDS
& $6.0 \times 10^{8}$ 
& $4.8 \times 10^{8}$ 
& $3.6 \times 10^{8}$ 
\\
MCD
& $1.3 \times 10^{9}$ 
& $1.3 \times 10^{9}$ 
& $1.2 \times 10^{9}$ 
\\
LDVI
& $1.3 \times 10^{9}$ 
& $1.3 \times 10^{9}$ 
& $1.3 \times 10^{9}$ 
\\
\bottomrule
\end{tabular}
}}
\end{sc}
\end{small}
\end{center}
\vskip -0.1in
\caption{Number of function evaluations (NFE), that is number of times a sampling method queries $\gamma(\x)$ until achieving the highest ELBO value for varying dimensions $d$ on MoG.}
\label{tab:efficiency}
\end{table}

\section{Ablation Studies} \label{appendix:ablations}
\subsection{Ablation Study: Batchsize and Number of Particles} \label{abl:batchsize}
\textbf{Experimental Setup.} We test the influence of different batchsizes/number of particles on ELBO and EMC on the MoG experiment for various methods. We use the parameters detailed in Appendix \ref{appendix:algos_params}.

\textbf{Discussion.} The results for the ablation study for the batchsize can be found in Table \ref{abl_batchsize}. We find that increasing batchsizes do not yield significant performance increases for simple methods such as MFVI. For more complex methods such as MCD or DDS, larger batchsizes tend to yield consistently better ELBO values across varying dimensionalities of the target density. In contrast, EMC values are unaffected by larger batchsizes (cf. MCD $d=200$). 

The results for the number of particles can be found in Table \ref{abl_particles}. Surprisingly, ELBO values do often not improve beyond $512$ particles, despite particle interactions through resampling \cite{del2006sequential}. Moreover, similar to the batch size, EMC does not change significantly when using a larger number of particles.

\begin{table*}[h!]
\begin{center}
\begin{small}
\begin{sc} {
\resizebox{\textwidth}{!}{%
\begin{tabular}{l|l|ccc|ccc}
\toprule
   & & \multicolumn{3}{c|}{\textbf{ELBO} $\uparrow$} & \multicolumn{3}{c}{\textbf{EMC} $\uparrow$}   \\ 
\textbf{Method} & \textbf{Batchsize}  & $d=2$ & $d=50$  & $d=200$ & $d=2$ & $d=50$  & $d=200$   \\ 
\midrule
MFVI
& 64
& $\mathbf{-3.011 \scriptstyle \pm 0.003}$ 
& $-3.707 \scriptstyle \pm 0.002$ 
& $-3.746 \scriptstyle \pm 0.001$ 
& $\mathbf{0.383 \scriptstyle \pm 0.002}$ 
& $0.0 \scriptstyle \pm 0.0$ 
& $0.0 \scriptstyle \pm 0.0$ 
\\
& 128
& $\mathbf{-3.012 \scriptstyle \pm 0.004}$ 
& $\mathbf{-3.698 \scriptstyle \pm 0.002}$ 
& $-3.731 \scriptstyle \pm 0.002$ 
& $\mathbf{0.382 \scriptstyle \pm 0.003}$ 
& $0.0 \scriptstyle \pm 0.0$ 
& $0.0 \scriptstyle \pm 0.0$ 
\\
& 512
& $\mathbf{-3.011 \scriptstyle \pm 0.004}$ 
& $\mathbf{-3.694 \scriptstyle \pm 0.0}$ 
& $\mathbf{-3.706 \scriptstyle \pm 0.0}$ 
& $\mathbf{0.382 \scriptstyle \pm 0.002}$ 
& $0.0 \scriptstyle \pm 0.0$ 
& $0.0 \scriptstyle \pm 0.0$ 
\\
& 1024
& $\mathbf{-3.012 \scriptstyle \pm 0.003}$ 
& $\mathbf{-3.692 \scriptstyle \pm 0.001}$ 
& $\mathbf{-3.701 \scriptstyle \pm 0.002}$ 
& $\mathbf{0.382 \scriptstyle \pm 0.002}$ 
& $0.0 \scriptstyle \pm 0.0$ 
& $0.0 \scriptstyle \pm 0.0$ 
\\
& 2048
& $\mathbf{-3.012 \scriptstyle \pm 0.003}$ 
& $\mathbf{-3.691 \scriptstyle \pm 0.0}$ 
& $\mathbf{-3.697 \scriptstyle \pm 0.001}$ 
& $\mathbf{0.383 \scriptstyle \pm 0.002}$ 
& $0.0 \scriptstyle \pm 0.0$ 
& $0.0 \scriptstyle \pm 0.0$ 
\\
\midrule
MCD
& 64
& $-3.017 \scriptstyle \pm 0.2$ 
& $-942.74 \scriptstyle \pm 8.447$ 
& $-4699.422 \scriptstyle \pm 269.44$ 
& $\mathbf{0.796 \scriptstyle \pm 0.003}$ 
& $\mathbf{0.994 \scriptstyle \pm 0.001}$ 
& $\mathbf{0.989 \scriptstyle \pm 0.0}$ 
\\
& 128
& $-2.685 \scriptstyle \pm 0.168$ 
& $-889.472 \scriptstyle \pm 7.41$ 
& $-4145.279 \scriptstyle \pm 179.564$ 
& $\mathbf{0.798 \scriptstyle \pm 0.001}$ 
& $\mathbf{0.994 \scriptstyle \pm 0.001}$ 
& $\mathbf{0.988 \scriptstyle \pm 0.0}$ 
\\
& 512
& $-2.409 \scriptstyle \pm 0.05$ 
& $-876.718 \scriptstyle \pm 6.132$ 
& $\mathbf{-3442.883 \scriptstyle \pm 260.824}$ 
& $\mathbf{0.796 \scriptstyle \pm 0.002}$ 
& $\mathbf{0.994 \scriptstyle \pm 0.0}$ 
& $\mathbf{0.988 \scriptstyle \pm 0.0}$ 
\\
& 1024
& $-2.277 \scriptstyle \pm 0.131$ 
& $-844.588 \scriptstyle \pm 10.761$ 
& OOM 
& $\mathbf{0.797 \scriptstyle \pm 0.002}$ 
& $\mathbf{0.994 \scriptstyle \pm 0.0}$ 
& OOM 
\\
& 2048
& $\mathbf{-2.257 \scriptstyle \pm 0.075}$ 
& $\mathbf{-823.443 \scriptstyle \pm 18.151}$ 
& OOM 
& $\mathbf{0.796 \scriptstyle \pm 0.002}$ 
& $\mathbf{0.994 \scriptstyle \pm 0.0}$ 
& OOM 
\\
\midrule
DDS
& 64
& $-0.807 \scriptstyle \pm 0.036$ 
& $-16.83 \scriptstyle \pm 0.404$ 
& $-67.053 \scriptstyle \pm 0.993$ 
& $0.973 \scriptstyle \pm 0.002$ 
& $\mathbf{0.992 \scriptstyle \pm 0.0}$ 
& $\mathbf{0.984 \scriptstyle \pm 0.001}$ 
\\
& 128
& $-0.716 \scriptstyle \pm 0.009$ 
& $-16.092 \scriptstyle \pm 0.247$ 
& $-65.232 \scriptstyle \pm 0.4$ 
& $0.978 \scriptstyle \pm 0.002$ 
& $\mathbf{0.991 \scriptstyle \pm 0.001}$ 
& $\mathbf{0.983 \scriptstyle \pm 0.001}$ 
\\
& 512
& $-0.611 \scriptstyle \pm 0.022$ 
& $-15.61 \scriptstyle \pm 0.206$ 
& $-63.135 \scriptstyle \pm 0.348$ 
& $\mathbf{0.984 \scriptstyle \pm 0.001}$ 
& $\mathbf{0.992 \scriptstyle \pm 0.001}$ 
& $\mathbf{0.983 \scriptstyle \pm 0.001}$ 
\\
& 1024
& $-0.593 \scriptstyle \pm 0.011$ 
& $-15.414 \scriptstyle \pm 0.161$ 
& $-62.086 \scriptstyle \pm 0.258$ 
& $\mathbf{0.985 \scriptstyle \pm 0.001}$ 
& $\mathbf{0.992 \scriptstyle \pm 0.0}$ 
& $\mathbf{0.982 \scriptstyle \pm 0.002}$ 
\\
& 2048
& $\mathbf{-0.556 \scriptstyle \pm 0.009}$ 
& $\mathbf{-15.313 \scriptstyle \pm 0.165}$ 
& $\mathbf{-61.576 \scriptstyle \pm 0.384}$ 
& $\mathbf{0.988 \scriptstyle \pm 0.001}$ 
& $\mathbf{0.992 \scriptstyle \pm 0.001}$ 
& $\mathbf{0.979 \scriptstyle \pm 0.001}$ 
\\
\bottomrule
\end{tabular}
}}
\end{sc}
\end{small}
\end{center}
\vskip -0.1in
\caption{ELBO and EMC values for varying batch sizes for different methods,  and dimensions of the MoG target density. Best values are marked with bold font. Here, OOM refers to `out of memory'.}
\label{abl_batchsize}
\end{table*}

\begin{table*}[h!]
\label{table:abl_particles}
\begin{center}
\begin{small}
\begin{sc} {
\resizebox{\textwidth}{!}{%
\begin{tabular}{l|l|ccc|ccc}
\toprule
   & & \multicolumn{3}{c|}{\textbf{ELBO} $\uparrow$} & \multicolumn{3}{c}{\textbf{EMC} $\uparrow$}   \\ 
\textbf{Method} & \textbf{Particles}  & $d=2$ & $d=50$  & $d=200$ & $d=2$ & $d=50$  & $d=200$   \\ 
\midrule
SMC
& 64
& $-9.267 \scriptstyle \pm 0.217$ 
& $-2622.073 \scriptstyle \pm 21.637$ 
& $-17904.276 \scriptstyle \pm 25.557$ 
& $0.824 \scriptstyle \pm 0.018$ 
& $0.0 \scriptstyle \pm 0.0$ 
& $0.0 \scriptstyle \pm 0.0$ 
\\
& 128
& $-9.08 \scriptstyle \pm 0.041$ 
& $-2647.733 \scriptstyle \pm 31.935$ 
& $-16999.909 \scriptstyle \pm 14.959$ 
& $0.879 \scriptstyle \pm 0.035$ 
& $0.0 \scriptstyle \pm 0.0$ 
& $0.0 \scriptstyle \pm 0.0$ 
\\
& 512
& $-8.823 \scriptstyle \pm 0.033$ 
& $\mathbf{-1911.449 \scriptstyle \pm 8.527}$ 
& $-16867.03 \scriptstyle \pm 44.663$ 
& $0.941 \scriptstyle \pm 0.004$ 
& $0.0 \scriptstyle \pm 0.0$ 
& $0.0 \scriptstyle \pm 0.0$ 
\\
& 1024
& $\mathbf{-8.595 \scriptstyle \pm 0.035}$ 
& $-2323.482 \scriptstyle \pm 13.52$ 
& $-15565.314 \scriptstyle \pm 78.958$ 
& $\mathbf{0.971 \scriptstyle \pm 0.004}$ 
& $0.0 \scriptstyle \pm 0.0$ 
& $0.0 \scriptstyle \pm 0.0$ 
\\
& 2048
& $-10.317 \scriptstyle \pm 0.028$ 
& $-2041.686 \scriptstyle \pm 20.993$ 
& $\mathbf{-15032.371 \scriptstyle \pm 46.049}$ 
& $\mathbf{0.965 \scriptstyle \pm 0.002}$ 
& $0.0 \scriptstyle \pm 0.0$ 
& $0.0 \scriptstyle \pm 0.0$ 
\\
\midrule
CRAFT
& 64
& $-3.666 \scriptstyle \pm 0.048$ 
& $-793.354 \scriptstyle \pm 19.752$ 
& $-4646.891 \scriptstyle \pm 77.062$ 
& $\mathbf{0.986 \scriptstyle \pm 0.001}$ 
& $0.0 \scriptstyle \pm 0.0$ 
& $0.0 \scriptstyle \pm 0.0$ 
\\
& 128
& $-3.604 \scriptstyle \pm 0.039$ 
& $-790.385 \scriptstyle \pm 23.036$ 
& $-4656.227 \scriptstyle \pm 80.153$ 
& $\mathbf{0.986 \scriptstyle \pm 0.001}$ 
& $0.0 \scriptstyle \pm 0.0$ 
& $0.0 \scriptstyle \pm 0.0$ 
\\
& 512
& $-3.6 \scriptstyle \pm 0.061$ 
& $-784.881 \scriptstyle \pm 14.364$ 
& $\mathbf{-4624.869 \scriptstyle \pm 63.618}$ 
& $\mathbf{0.987 \scriptstyle \pm 0.0}$ 
& $0.0 \scriptstyle \pm 0.0$ 
& $0.0 \scriptstyle \pm 0.0$ 
\\
& 1024
& $\mathbf{-3.552 \scriptstyle \pm 0.041}$ 
& $-785.251 \scriptstyle \pm 16.847$ 
& $-4632.063 \scriptstyle \pm 68.715$ 
& $\mathbf{0.986 \scriptstyle \pm 0.002}$ 
& $0.0 \scriptstyle \pm 0.0$ 
& $0.0 \scriptstyle \pm 0.0$ 
\\
& 2048
& $\mathbf{-3.553 \scriptstyle \pm 0.05}$ 
& $\mathbf{-782.068 \scriptstyle \pm 13.855}$ 
& $-4625.769 \scriptstyle \pm 55.853$ 
& $\mathbf{0.987 \scriptstyle \pm 0.001}$ 
& $0.0 \scriptstyle \pm 0.0$ 
& $0.0 \scriptstyle \pm 0.0$ 
\\
\bottomrule
\end{tabular}
}}
\end{sc}
\end{small}
\end{center}
\vskip -0.1in
\caption{ELBO and EMC values for varying number of particles  and dimensions of the MoG target density.}
\label{abl_particles}
\end{table*}

\subsection{Ablation Study: Number of Temperatures / Timesteps T} \label{abl:num_steps}
\textbf{Experimental Setup.} We test the influence of different number of temperatures/timesteps $T$ for methods of sequential nature such as sequential importance sampling or SDE based methods. We use batch sizes of $512$. The remaining parameters are set according to Appendix \ref{appendix:algos_params}.

\begin{figure*}[h!]
\centering
\setlength{\tabcolsep}{1pt}
\begin{tabular}{ccccccc}
  \includegraphics[width=0.30\textwidth]{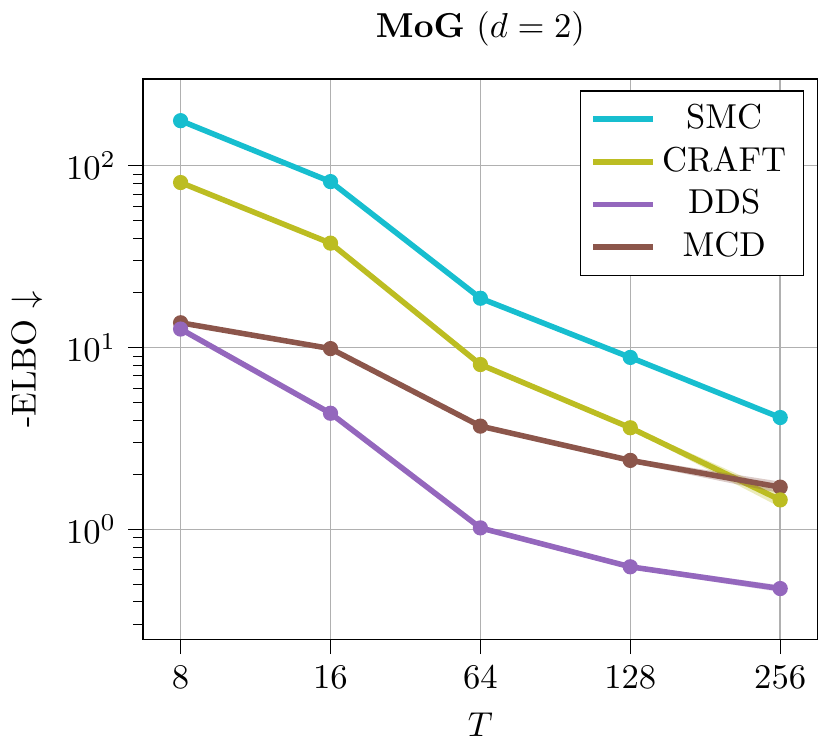} &   
  \includegraphics[width=0.30\textwidth]{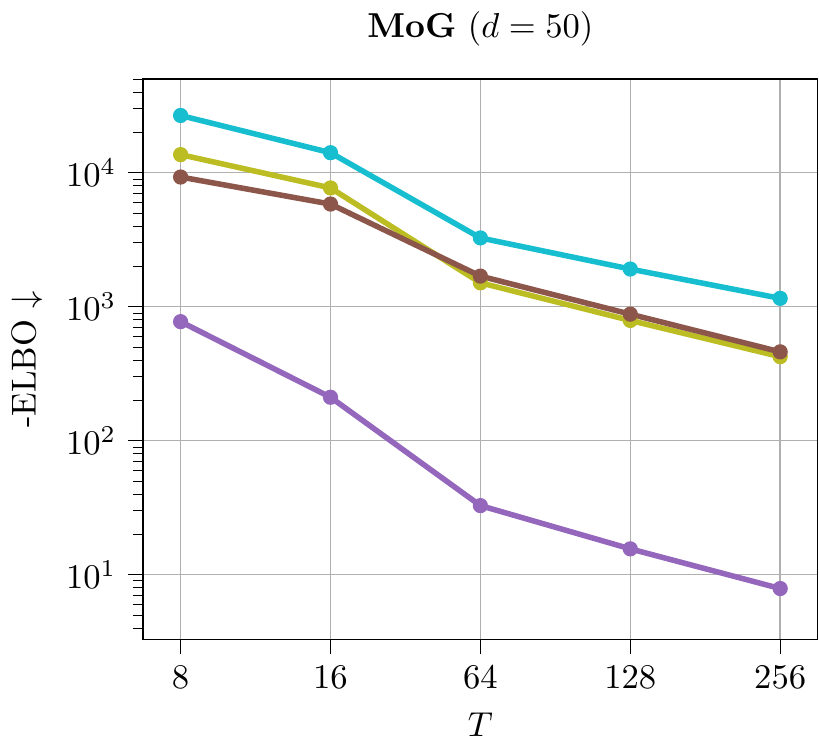} &   
  \includegraphics[width=0.30\textwidth]{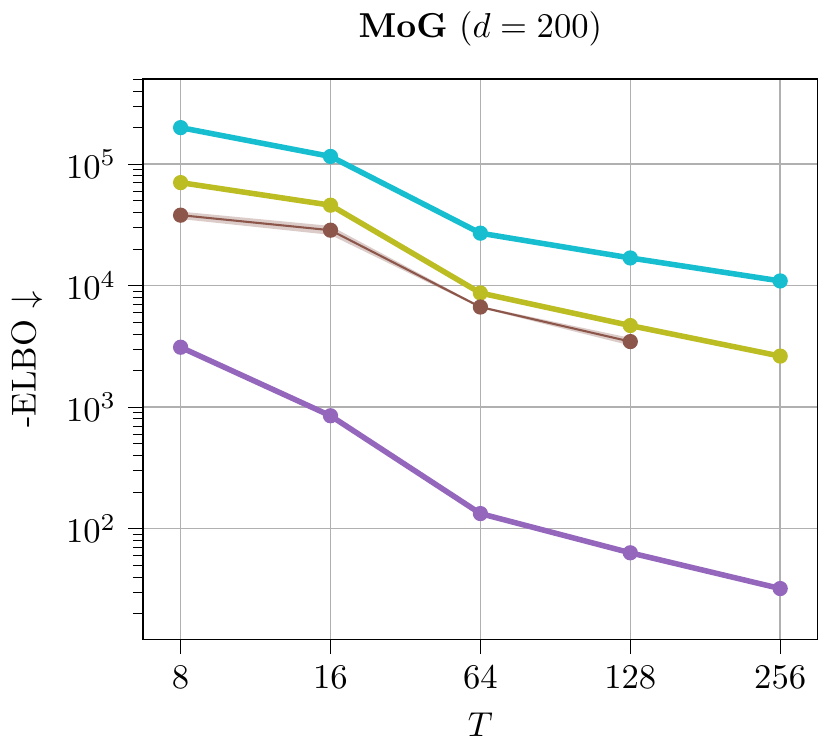} &   
  \\
  \includegraphics[width=0.30\textwidth]{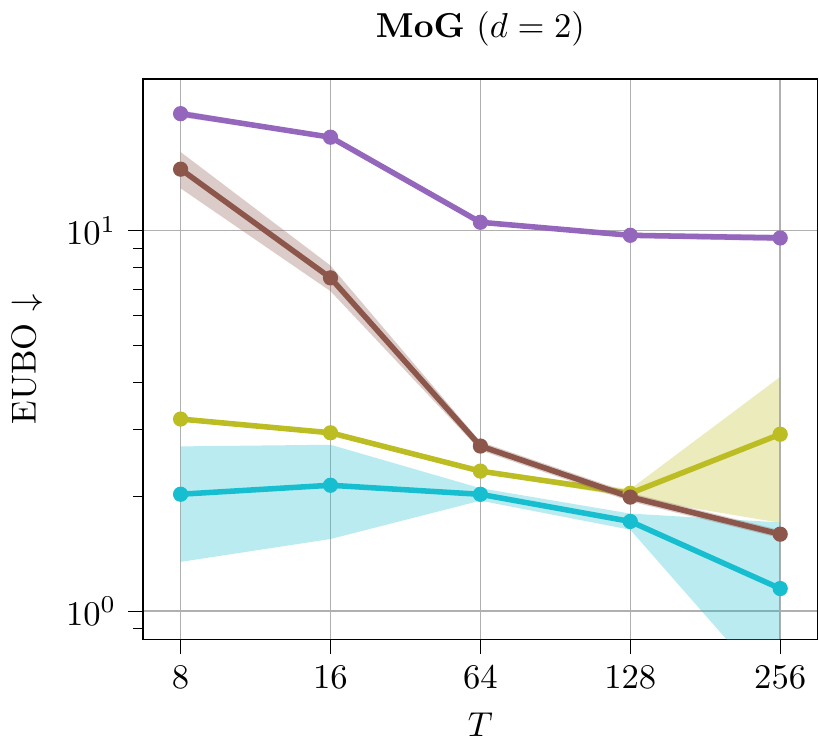} &   
  \includegraphics[width=0.30\textwidth]{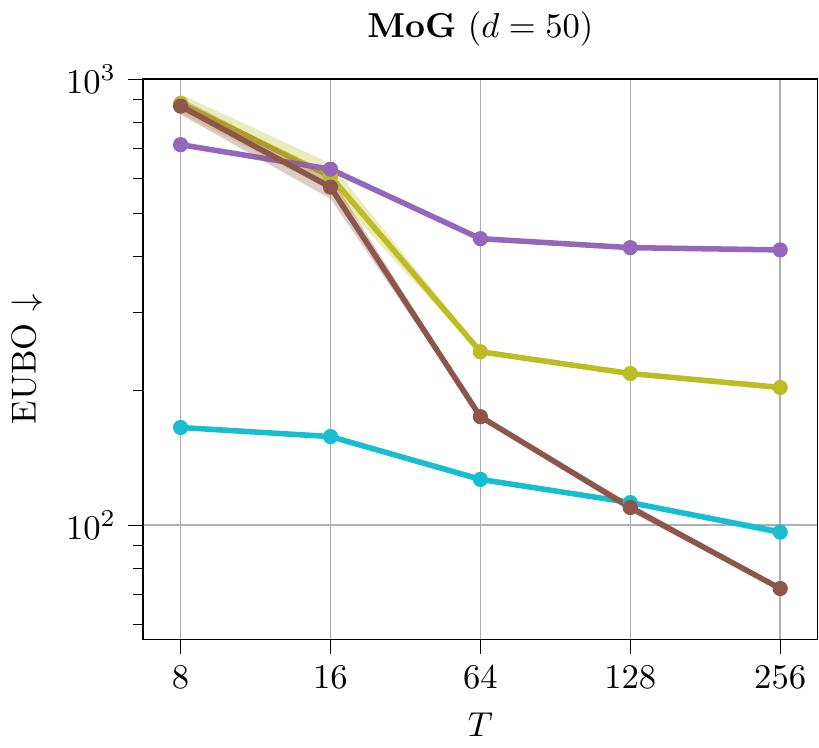} &   
  \includegraphics[width=0.30\textwidth]{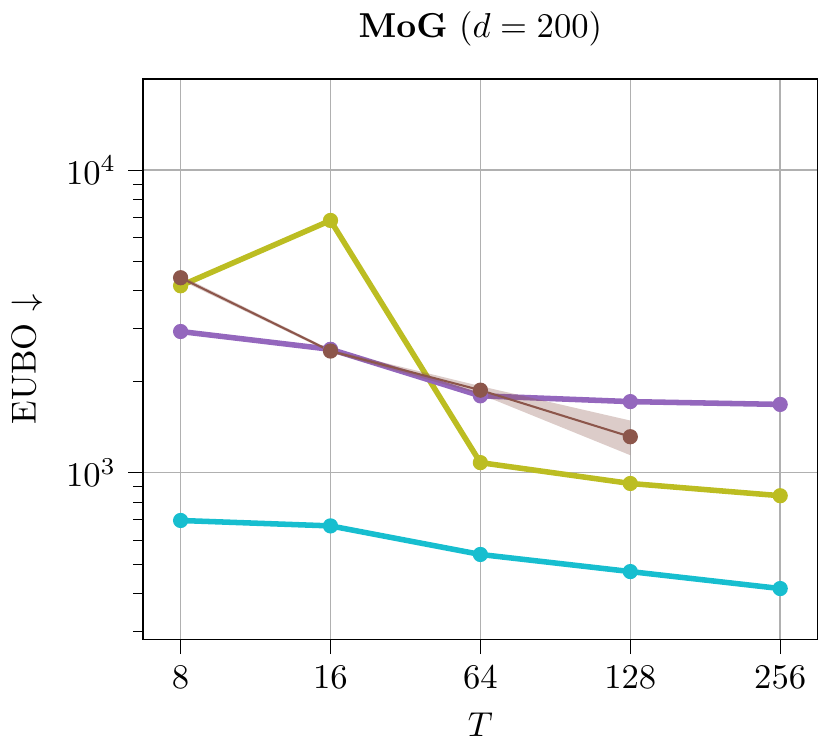} & 
  \\
\end{tabular}
\caption{Negative ELBO and EUBO values for varying temperatures/timesteps $T$ for different dimensions of the MoG target density. Best values are marked with bold font. Missing values for $T=256$ are caused by out-of-memory problems.}
\label{abl_timesteps}
\end{figure*}

\textbf{Discussion.} 
The results are illustrated in Figure \ref{abl_timesteps}. We can see that using larger values of $T$ tend to improves ELBO and EUBO values across all methods.  

\subsection{Ablation Study: Sequential Monte Carlo Design Choices} \label{abl:smc_choices}
\textbf{Experimental Setup.} As Sequential Monte Carlo is the basis for many sampling methods such as SNF \cite{wu2020stochastic}, AFT \cite{arbel2021annealed}, CRAFT \cite{matthews2022continual}, or FAB \cite{midgley2022flow} we perform a thorough ablation of its design choices. In particular, we ablate the influence of the MCMC kernel and whether or not resampling is used.  We tested Metropolis-Hastings (MH) and Hamiltonian Monte Carlo (HMC) MCMC kernels where we used the same number of function evaluations and hand-tuned the stepsizes such obtained a rejection rate $\approx 0.65$. The results are shown in Table \ref{abl_smc}.

\textbf{Discussion.} HMC outperforms MH across all dimensions with respect to both, ELBO and EMC values. Surprisingly, not using resampling avoids mode collapse entirely as indicated by $\text{EMC}\approx 1$.
\begin{table*}[h!]
\begin{center}
\begin{small}
\begin{sc}
\resizebox{\textwidth}{!}{%
\begin{tabular}{c|c|ccc|ccc}
\toprule
\textbf{MCMC}   & \textbf{Re- }& \multicolumn{3}{c|}{\textbf{ELBO} $\uparrow$} & \multicolumn{3}{c}{\textbf{EMC} $\uparrow$}   \\ 
\textbf{Kernel} & \textbf{sampling}  & $d=2$ & $d=50$  & $d=200$ & $d=2$ & $d=50$  & $d=200$   \\ 
\midrule
-
& $\tikzxmark$
& $-9.473 \scriptstyle{\pm 0.000}$ 
& $-32034.303 \scriptstyle{\pm 0.000}$ 
& $-292642.344 \scriptstyle{\pm 0.000}$ 
& $0.785 \scriptstyle{\pm 0.000}$ 
& $\mathbf{0.987 \scriptstyle{\pm 0.000}}$ 
& $\mathbf{0.988 \scriptstyle{\pm 0.000}}$ 
\\
-
& $\tikzcmark$
& $-9.28 \scriptstyle{\pm 0.2044}$ 
& $-27534.303 \scriptstyle{\pm 72.32}$ 
& $-288123.325 \scriptstyle{\pm 108.010}$ 
& $0.618 \scriptstyle{\pm 0.191}$ 
& $0 \scriptstyle{\pm 0}$ 
& $0 \scriptstyle{\pm 0}$ 
\\
MH
& $\tikzxmark$
& $-9.166 \scriptstyle{\pm 0.138}$ 
& $-26686.496 \scriptstyle{\pm 412.669}$ 
& $-275404.367 \scriptstyle{\pm 1375.306}$ 
& $0.785 \scriptstyle{\pm 0.003}$ 
& $\mathbf{0.987 \scriptstyle{\pm 0.000}}$ 
& $\mathbf{0.988 \scriptstyle{\pm 0.000}}$ 
\\
MH
& $\tikzcmark$
& $-9.064 \scriptstyle{\pm 0.034}$ 
& $-22411.798 \scriptstyle{\pm 69.874}$ 
& $-251904.734 \scriptstyle{\pm 422.895}$ 
& $0.864 \scriptstyle{\pm 0.021}$ 
& $0 \scriptstyle{\pm 0}$ 
& $0 \scriptstyle{\pm 0}$ 
\\
HMC
& $\tikzxmark$
& $\mathbf{-8.736 \scriptstyle{\pm 0.031}}$ 
& $-2272.619 \scriptstyle{\pm 96.639}$ 
& $-18270.795 \scriptstyle{\pm 91.703}$ 
& $0.798 \scriptstyle{\pm 0.006}$ 
& $\mathbf{0.986 \scriptstyle{\pm 0.000}}$ 
& $\mathbf{0.988 \scriptstyle{\pm 0.000}}$ 
\\
HMC
& $\tikzcmark$
& ${-8.850 \scriptstyle{\pm 0.110}}$ 
& $\mathbf{-1931.168 \scriptstyle{\pm 18.844}}$ 
& $\mathbf{-16952.94 \scriptstyle{\pm 49.119}}$ 
& $\mathbf{0.940 \scriptstyle{\pm 0.006}}$ 
& $0 \scriptstyle{\pm 0}$ 
& $0 \scriptstyle{\pm 0}$ 
\\
\bottomrule
\end{tabular}
}
\end{sc}
\end{small}
\end{center}
\vskip -0.1in
\caption{Ablation study for Sequential Monte Carlo \cite{del2006sequential}. ELBO and EUBO values for different MCMC kernels and whether or not resampling is used. Here, MH refers to Metropolis-Hastings and HMC to Hamiltonian Monte Carlo \cite{bishop2006pattern}. Results are reported for different dimensions of the MoG target density.}
\label{abl_smc}
\end{table*}

\subsection{Ablation Study: Initial Model Support} \label{abl:init_support}
\textbf{Experimental Setup.} We test the influence of the initial model support for different methods of sequential nature. In particular, we vary the scale $\sigma_0^2$ of the initial proposal/base distribution $\pi_0(\x) = \mathcal{N}(0, \sigma_0^2\I)$. To that end, we report ELBO and EUBO values on the MoG experiment for varying dimensions. We use the parameters detailed in Appendix \ref{appendix:algos_params}. The results are shown in Table \ref{abl_scale}.

\textbf{Discussion.} 
The results of the ablation study investigating varied initial standard deviations for parameterizing the base distribution can be found in Table \ref{abl_scale}. We observe that, in terms of the ELBO, most methods exhibit poor performance with a higher initial scale, particularly in higher dimensions. Conversely, EMC values tend to get $0$ for small initial scales and $1$ for large initial scales. 
\begin{table*}[h!]
\begin{center}
\begin{small}
\begin{sc} {
\resizebox{\textwidth}{!}{%
\begin{tabular}{l|c|ccc|ccc}
\toprule
\textbf{Method}   & \textbf{Initial}  & \multicolumn{3}{c|}{\textbf{ELBO} $\uparrow$} & \multicolumn{3}{c}{\textbf{EMC} $\uparrow$}   \\ 
 & \textbf{Scale}  & $d=2$ & $d=50$  & $d=200$ & $d=2$ & $d=50$  & $d=200$   \\ 
\midrule
SMC
& 1
& $-3.717 \scriptstyle \pm 0.056$ 
& $-1800.882 \scriptstyle \pm 38.348$ 
& $-12181.939 \scriptstyle \pm 110.871$ 
& $0.002 \scriptstyle \pm 0.002$ 
& $0.0 \scriptstyle \pm 0.0$ 
& $0.0 \scriptstyle \pm 0.0$ 
\\
& 10
& $\mathbf{-0.175 \scriptstyle \pm 0.408}$ 
& $\mathbf{-313.926 \scriptstyle \pm 4.847}$ 
& $\mathbf{-2008.238 \scriptstyle \pm 10.954}$ 
& $0.722 \scriptstyle \pm 0.022$ 
& $0.0 \scriptstyle \pm 0.0$ 
& $0.0 \scriptstyle \pm 0.0$ 
\\
& 30
& $-0.666 \scriptstyle \pm 0.081$ 
& $-674.1 \scriptstyle \pm 12.747$ 
& $-5014.372 \scriptstyle \pm 27.284$ 
& $\mathbf{0.957 \scriptstyle \pm 0.007}$ 
& $0.0 \scriptstyle \pm 0.0$ 
& $0.0 \scriptstyle \pm 0.0$ 
\\
& 60
& $-8.823 \scriptstyle \pm 0.033$ 
& $-1911.449 \scriptstyle \pm 8.527$ 
& $-16867.03 \scriptstyle \pm 44.663$ 
& $0.941 \scriptstyle \pm 0.004$ 
& $0.0 \scriptstyle \pm 0.0$ 
& $0.0 \scriptstyle \pm 0.0$ 
\\
\midrule
CRAFT
& 1
& $-2.675 \scriptstyle \pm 0.236$ 
& $\mathbf{-11.333 \scriptstyle \pm 0.644}$ 
& $\mathbf{-83.301 \scriptstyle \pm 1.267}$ 
& $0.143 \scriptstyle \pm 0.033$ 
& $0.0 \scriptstyle \pm 0.0$ 
& $0.0 \scriptstyle \pm 0.0$ 
\\
& 10
& $-0.633 \scriptstyle \pm 0.538$ 
& $-136.414 \scriptstyle \pm 2.482$ 
& $-1090.374 \scriptstyle \pm 22.117$ 
& $0.657 \scriptstyle \pm 0.233$ 
& $0.0 \scriptstyle \pm 0.0$ 
& $0.0 \scriptstyle \pm 0.0$ 
\\
& 30
& $\mathbf{-0.229 \scriptstyle \pm 0.018}$
& $-350.247 \scriptstyle \pm 11.605$ 
& $-2482.919 \scriptstyle \pm 12.176$ 
& $0.974 \scriptstyle \pm 0.004$ 
& $0.0 \scriptstyle \pm 0.0$ 
& $0.0 \scriptstyle \pm 0.0$ 
\\
& 60
& $-3.563 \scriptstyle \pm 0.057$ 
& $-784.881 \scriptstyle \pm 14.364$ 
& $-4624.869 \scriptstyle \pm 63.618$ 
& $\mathbf{0.987 \scriptstyle \pm 0.001}$ 
& $0.0 \scriptstyle \pm 0.0$ 
& $0.0 \scriptstyle \pm 0.0$ 
\\
\midrule
MCD
& 1
& $-3.676 \scriptstyle \pm 0.001$ 
& $\mathbf{-3.292 \scriptstyle \pm 0.011}$ 
& $\mathbf{-4.281 \scriptstyle \pm 0.039}$ 
& $0.005 \scriptstyle \pm 0.0$ 
& $0.187 \scriptstyle \pm 0.0$ 
& $0.005 \scriptstyle \pm 0.001$ 
\\
& 10
& $-1.653 \scriptstyle \pm 0.032$ 
& $-87.5 \scriptstyle \pm 0.519$ 
& $-144.237 \scriptstyle \pm 4.133$ 
& $0.613 \scriptstyle \pm 0.003$ 
& $0.658 \scriptstyle \pm 0.004$ 
& $0.647 \scriptstyle \pm 0.002$ 
\\
& 30
& $\mathbf{-1.138 \scriptstyle \pm 0.064}$ 
& $-441.73 \scriptstyle \pm 2.245$ 
& $-1265.551 \scriptstyle \pm 6.991$ 
& $\mathbf{0.94 \scriptstyle \pm 0.001}$ 
& $0.961 \scriptstyle \pm 0.0$ 
& $0.942 \scriptstyle \pm 0.002$ 
\\
& 60
& $-2.384 \scriptstyle \pm 0.059$ 
& $-878.12 \scriptstyle \pm 8.598$ 
& $-3458.28 \scriptstyle \pm 248.958$ 
& $0.798 \scriptstyle \pm 0.003$ 
& $\mathbf{0.994 \scriptstyle \pm 0.001}$ 
& $\mathbf{0.988 \scriptstyle \pm 0.0}$ 
\\
\midrule
DDS
& 1
& $-3.622 \scriptstyle \pm 0.012$ 
& $\mathbf{-6.053 \scriptstyle \pm 0.624}$ 
& $-49.0 \scriptstyle \pm 10.277$ 
& $0.0 \scriptstyle \pm 0.0$ 
& $0.187 \scriptstyle \pm 0.0$ 
& $0.0 \scriptstyle \pm 0.0$ 
\\
& 10
& $-0.737 \scriptstyle \pm 0.024$ 
& $-6.954 \scriptstyle \pm 0.146$ 
& $\mathbf{-20.149 \scriptstyle \pm 0.075}$ 
& $0.85 \scriptstyle \pm 0.001$ 
& $0.26 \scriptstyle \pm 0.031$ 
& $0.348 \scriptstyle \pm 0.011$ 
\\
& 30
& $\mathbf{-0.408 \scriptstyle \pm 0.01}$ 
& $-10.604 \scriptstyle \pm 0.165$ 
& $-42.396 \scriptstyle \pm 0.105$ 
& $\mathbf{0.989 \scriptstyle \pm 0.001}$ 
& $0.941 \scriptstyle \pm 0.003$ 
& $0.841 \scriptstyle \pm 0.011$ 
\\
& 60
& $-0.612 \scriptstyle \pm 0.019$ 
& $-15.598 \scriptstyle \pm 0.106$ 
& $-63.101 \scriptstyle \pm 0.253$ 
& $\mathbf{0.984 \scriptstyle \pm 0.001}$ 
& $\mathbf{0.992 \scriptstyle \pm 0.001}$ 
& $\mathbf{0.983 \scriptstyle \pm 0.002}$ 
\\
\bottomrule
\end{tabular}
}}
\end{sc}
\end{small}
\end{center}
\vskip -0.1in
\caption{ELBO and EMC values for varying initial scales, and dimensions of the MoG target density.}
\label{abl_scale}
\end{table*}

\subsection{Ablation Study: Langevin Methods} \label{abl:langevin_choices}
\textbf{Experimental Setup.} The augemented ELBO allows for end-to-end training of several parameters that otherwise need careful tuning. \cite{geffner2022langevin} showed that learning the mean and variance of the proposal distribution $\pi_0$, the time discretization stepsize $\Delta_t$ and annealing schedule $(\beta_t)_{t=1}^T$ by maximizing the extended ELBO. Here, we test the influence of training vs. fixing these paramters for MCD \cite{doucet2022annealed} on the MoG target for varying dimensions. The results are shown in Table \ref{abl_langevin}. The fixed parameters are chosen according to Table \ref{appendix:algos_params}.

\textbf{Discussion.} 
We observe that learning more parameters tend to yield higher ELBO values. However, especially learning the parameters of the proposal $\pi_0$ results in low EMC values.

\begin{table*}[h!]
\begin{center}
\begin{small}
\begin{sc} {
\resizebox{\textwidth}{!}{%
\begin{tabular}{ccc|ccc|ccc}
\toprule
\multicolumn{3}{c|}{\textbf{Trainable}}  & \multicolumn{3}{c|}{\textbf{ELBO} $\uparrow$} & \multicolumn{3}{c}{\textbf{EMC} $\uparrow$}   \\ 
 $\sigma_t$ & $\beta_t$ & $\target_0$ & $d=2$ &$d=50$  & $d=200$ & $d=2$ & $d=50$  & $d=200$   \\ 
\midrule
$\tikzxmark$
& $\tikzxmark$
& $\tikzxmark$
& $-3.519 \scriptstyle \pm 0.154$ 
& $-2513.292 \scriptstyle \pm 26.017$ 
& $-13575.6 \scriptstyle \pm 414.217$ 
& $0.799 \scriptstyle \pm 0.004$ 
& $\mathbf{0.994 \scriptstyle \pm 0.0}$ 
& $\mathbf{0.988 \scriptstyle \pm 0.001}$ 
\\
$\tikzcmark$
& $\tikzxmark$
& $\tikzxmark$
& $-2.441 \scriptstyle \pm 0.079$ 
& $-1141.639 \scriptstyle \pm 18.651$ 
& $-6574.401 \scriptstyle \pm 114.962$ 
& $0.819 \scriptstyle \pm 0.004$ 
& $\mathbf{0.994 \scriptstyle \pm 0.0}$ 
& $\mathbf{0.988 \scriptstyle \pm 0.001}$ 
\\
$\tikzxmark$
& $\tikzcmark$
& $\tikzxmark$
& $-2.384 \scriptstyle \pm 0.059$ 
& $-878.12 \scriptstyle \pm 8.598$ 
& $-3458.28 \scriptstyle \pm 248.958$ 
& $0.798 \scriptstyle \pm 0.003$ 
& $\mathbf{0.994 \scriptstyle \pm 0.001}$ 
& $\mathbf{0.988 \scriptstyle \pm 0.0}$ 
\\
$\tikzcmark$
& $\tikzcmark$
& $\tikzxmark$
& $-1.51 \scriptstyle \pm 0.035$ 
& $-173.002 \scriptstyle \pm 1.548$ 
& $-825.303 \scriptstyle \pm 44.797$ 
& $0.828 \scriptstyle \pm 0.003$ 
& $\mathbf{0.993 \scriptstyle \pm 0.0}$ 
& $\mathbf{0.989 \scriptstyle \pm 0.001}$ 
\\
$\tikzxmark$
& $\tikzxmark$
& $\tikzcmark$
& $-1.621 \scriptstyle \pm 0.216$ 
& $-38.022 \scriptstyle \pm 41.035$ 
& $-43.416 \scriptstyle \pm 7.242$ 
& $0.927 \scriptstyle \pm 0.015$ 
& $0.276 \scriptstyle \pm 0.132$ 
& $0.236 \scriptstyle \pm 0.053$ 
\\
$\tikzcmark$
& $\tikzxmark$
& $\tikzcmark$
& $-1.235 \scriptstyle \pm 0.072$ 
& $-29.238 \scriptstyle \pm 20.912$ 
& $\mathbf{-33.686 \scriptstyle \pm 4.508}$ 
& $\mathbf{0.95 \scriptstyle \pm 0.005}$ 
& $0.309 \scriptstyle \pm 0.109$ 
& $0.393 \scriptstyle \pm 0.014$ 
\\
$\tikzxmark$
& $\tikzcmark$
& $\tikzcmark$
& $-1.137 \scriptstyle \pm 0.118$ 
& $\mathbf{-8.323 \scriptstyle \pm 1.718}$ 
& $-103.968 \scriptstyle \pm 68.449$ 
& $0.936 \scriptstyle \pm 0.011$ 
& $0.19 \scriptstyle \pm 0.004$ 
& $0.341 \scriptstyle \pm 0.122$ 
\\
$\tikzcmark$
& $\tikzcmark$
& $\tikzcmark$
& $\mathbf{-1.05 \scriptstyle \pm 0.099}$ 
& $-10.526 \scriptstyle \pm 2.256$ 
& $-36.254 \scriptstyle \pm 15.018$ 
& $0.913 \scriptstyle \pm 0.017$ 
& $0.381 \scriptstyle \pm 0.075$ 
& $0.435 \scriptstyle \pm 0.062$ 
\\
\bottomrule
\end{tabular}
}}
\end{sc}
\end{small}
\end{center}
\vskip -0.1in
\caption{ELBO and EMC values of MCD for learning the mean and variance of the proposal distribution $\pi_0$, the diffusion coefficient $\sigma_t$ and annealing schedule $(\beta_t)_{t=1}^T$ by maximizing the extended ELBO for varying dimensions $d$ on the MoG target.}
\label{abl_langevin}
\end{table*}

\subsection{Ablation Study: Transport Flow Type} \label{abl:flows}
\begin{figure*}[h!]
\centering
\setlength{\tabcolsep}{1pt}
\begin{tabular}{ccccccc}
  \includegraphics[width=0.3\textwidth]{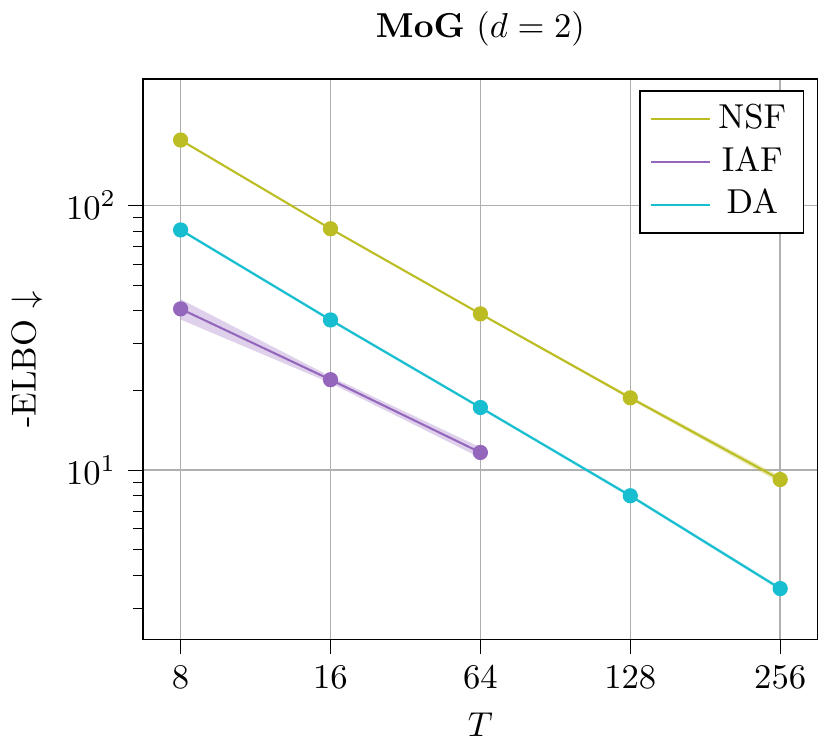} &   
  \includegraphics[width=0.3\textwidth]{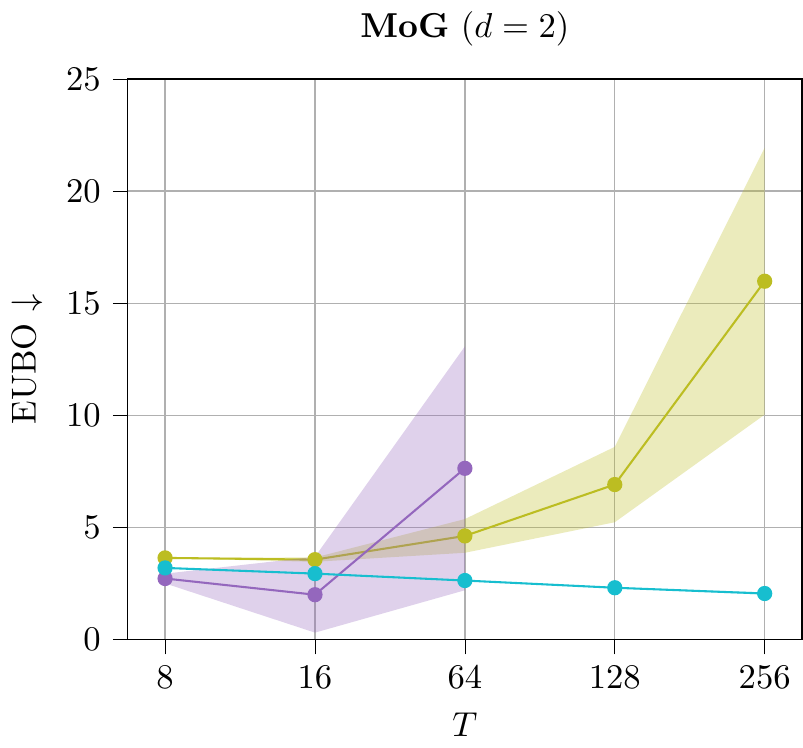} &   
  \\
\end{tabular}
\caption{ELBO and EUBO values for CRAFT for different flow types and number of temperatures T on the two-dimensional MoG target. In particular, diagonal affine flows, inverse autoregressive flows (IAF) \cite{kingma2016improved} and neural spline flows (NSF) \cite{durkan2019neural}. For larger $T$, IAF becomes numerically unstable.}
\label{abl_transport_flows}
\end{figure*}

\textbf{Experimental Setup.} We test different flow types as transport maps for CRAFT using a different number of temperatures $T$. In particular, we consider diagonal affine flows, inverse autoregressive flows \cite{kingma2016improved} and neural spline flows \cite{durkan2019neural} where we set the spline bounds to match the support of the MoG target. The results are visualized in Figure \ref{abl_transport_flows}.

\textbf{Discussion.} We found that diagonal affine paired with larger number of temperatures results in a better, more robust performance compared to using more sophisticated flow types. Moreover, the latter often result in out-of-memory problems on high dimensional problems.

\subsection{Ablation Study: Gradient Guidance} \label{abl:grad_network}
\textbf{Experimental Setup.} \cite{zhang2021path} proposed to use a network of the form $f^\theta(\x,t) = f_1^{\theta}(\x,t) + f^\theta_2(t)\nabla \log \gamma(\x)$ and initialize such that $f_1^{\theta}(\x,t)=0$. They showed that this gradient guidance helps with mode collapse and yields overall better results. \cite{vargas2023denoising, berner2022optimal, richter2023improved} adopted the approach and reported similar results. Here, we test the network architecture with and without gradient guidance $f^\theta_2(t)\nabla \log \gamma(\x)$ on the MoG target for a varying number of dimensions for the diffusion sampler.

\textbf{Discussion.} 
The results of this examination can be found in Table \ref{abl_grad_network} and indicate that both the ELBO and EMC significantly deteriorate without gradient guidance, and this degradation increases with higher dimensions. This aligns with the findings from \cite{zhang2021path,vargas2023denoising, berner2022optimal, richter2023improved}. 
\begin{table*}[h!]
\begin{center}
\begin{small}
\begin{sc} {
\resizebox{\textwidth}{!}{%
\begin{tabular}{c|ccc|ccc}
\toprule
   \textbf{Gradient} & \multicolumn{3}{c|}{\textbf{ELBO} $\uparrow$} & \multicolumn{3}{c}{\textbf{EMC} $\uparrow$}   \\ 
 \textbf{Guidance}  & $d=2$ & $d=50$  & $d=200$ & $d=2$ & $d=50$  & $d=200$   \\ 
\midrule
$\tikzxmark$
& $-3.105 \scriptstyle \pm 0.27$ 
& $-543.099 \scriptstyle \pm 13.612$ 
& $-247920.463 \scriptstyle \pm 4258.605$ 
& $0.453 \scriptstyle \pm 0.011$ 
& $0.0 \scriptstyle \pm 0.0$ 
& $0.243 \scriptstyle \pm 0.421$ 
\\
$\tikzcmark$
& $\mathbf{-0.612 \scriptstyle \pm 0.019}$ 
& $\mathbf{-15.598 \scriptstyle \pm 0.106}$ 
& $\mathbf{-63.101 \scriptstyle \pm 0.253}$ 
& $\mathbf{0.984 \scriptstyle \pm 0.001}$ 
& $\mathbf{0.992 \scriptstyle \pm 0.001}$ 
& $\mathbf{0.983 \scriptstyle \pm 0.002}$ 
\\
\bottomrule
\end{tabular}
}}
\end{sc}
\end{small}
\end{center}
\vskip -0.1in
\caption{ELBO and EMC values with and without gradient guidance $f^\theta_2(t)\nabla \log \gamma(\x)$ as part of the network architecture for the denoising diffusion sampler (DDS) on the MoG target for varying dimension $d$.}
\label{abl_grad_network}
\end{table*}

\subsection{Ablation Study: Pre-training the Proposal/Base-Distribution $\pi_0$} \label{abl:pretrain_base}
\textbf{Experimental Setup.} We test the impact of pre-training the mean and covariance matrix of the Gaussian proposal/base distribution $\pi_0$ using MFVI on the MoG target for varying dimensions. The results are shown in Table \ref{abl_pretrain_base}.

\textbf{Discussion.} Pretraining the the mean and covariance matrix of the Gaussian proposal/base distribution $\pi_0$ yields significantly higher ELBO values at the cost of EMC values close to 0. 
\begin{table*}[h!]
\begin{center}
\begin{small}
\begin{sc} {
\resizebox{\textwidth}{!}{%
\begin{tabular}{l|c|ccc|ccc}
\toprule
   &  \textbf{Pretrained} & \multicolumn{3}{c|}{\textbf{ELBO} $\uparrow$} & \multicolumn{3}{c}{\textbf{EMC} $\uparrow$}   \\ 
 \textbf{Method} & $\pi_0$  & $d=2$ & $d=50$  & $d=200$ & $d=2$ & $d=50$  & $d=200$   \\ 
\midrule
CRAFT
& $\tikzxmark$
& $\mathbf{-3.563 \scriptstyle \pm 0.057}$ 
& $-784.881 \scriptstyle \pm 14.364$ 
& $-4624.869 \scriptstyle \pm 63.618$ 
& $\mathbf{0.987 \scriptstyle \pm 0.001}$ 
& $0.0 \scriptstyle \pm 0.0$ 
& $0.0 \scriptstyle \pm 0.0$ 
\\
& $\tikzcmark$
& $-3.676 \scriptstyle \pm 0.007$ 
& $\mathbf{-3.501 \scriptstyle \pm 0.087}$ 
& $\mathbf{-3.699 \scriptstyle \pm 0.135}$ 
& $0.0 \scriptstyle \pm 0.0$ 
& $0.0 \scriptstyle \pm 0.0$ 
& $0.0 \scriptstyle \pm 0.0$ 
\\
\midrule
MCD
& $\tikzxmark$
& $\mathbf{-2.384 \scriptstyle \pm 0.059}$ 
& $-878.12 \scriptstyle \pm 8.598$ 
& $-3458.28 \scriptstyle \pm 248.958$ 
& $\mathbf{0.798 \scriptstyle \pm 0.003}$ 
& $\mathbf{0.994 \scriptstyle \pm 0.001}$ 
& $\mathbf{0.988 \scriptstyle \pm 0.0}$ 
\\
& $\tikzcmark$
& $-3.689 \scriptstyle \pm 0.0$ 
& $\mathbf{-3.746 \scriptstyle \pm 0.003}$ 
& $\mathbf{-3.938 \scriptstyle \pm 0.003}$ 
& $0.0 \scriptstyle \pm 0.0$ 
& $0.0 \scriptstyle \pm 0.0$ 
& $0.0 \scriptstyle \pm 0.0$ 
\\
\bottomrule
\end{tabular}
}}
\end{sc}
\end{small}
\end{center}
\vskip -0.1in
\caption{ELBO and EMC values for pre-trained/fixed Gaussian proposal/base distribution $\pi_0$ on the MoG target with varying dimensions $d$.}
\label{abl_pretrain_base}
\end{table*}

\end{document}